\documentclass[journal]{IEEEtran}
\usepackage{times}
\usepackage{epsfig}
\usepackage{graphicx}
\usepackage{float}
\usepackage{amsmath}
\usepackage{amssymb}
\usepackage{algorithm}
\usepackage{algpseudocode}
\usepackage{color,balance}
\usepackage{caption}
\usepackage{subcaption}
\usepackage{tabularx}
\usepackage{breqn}
\usepackage{tablefootnote}
\usepackage{multirow,comment}
\usepackage[numbers,sort&compress]{natbib}
\usepackage{lineno}
\usepackage[pagebackref=true,breaklinks=true,colorlinks,bookmarks=false]{hyperref}
\begin{document}

\title{Weight and Height Estimation from  a Single Human  Image Captured in the Wild}

\author{Hira Yaseen, Arif Mahmood,
\and
Waqas Sultani\\ 

\thanks{H. Yaseen, A. Mahmood, W. Sulatni are with the Department
of Computer Science, Information Technology University (ITU), 346-B, Ferozepur Road, Lahore, Pakistan.
E-mails:  PhDCS17002@itu.edu.pk, arif.mahmood@itu.edu.pk, waqas.sultani@itu.edu.pk}
}

\maketitle
\begin{abstract}
A person's physical characteristics such as weight and height are important indicators of his physical and mental health, daily life routines and finances. Body Mass Index (BMI) is  a well known measure that encodes the characteristics of both the weight and the height. 
BMI has been used as a self-monitoring tool,  and it  has long-term implications on one's life. For example, it  may help predicting the risk of various diseases and estimating longevity.
Automatic BMI estimation using a single person image in the wild is a challenging task due to wide variations in human pose, camera geometry, personal appearance and distracting backgrounds. In this paper, we explore the performance of deep neural networks using single and multi-task learning by employing different modalities including RGB, depth-maps, pose-affinity maps, and edge-maps to predict BMI, weight, and height from daily life images available on social networking websites.  Currently, no full  body image dataset for BMI estimation is publicly available, therefore we propose a new dataset consisting of 6105 images with ground truth labels of height, weight and BMI. Our proposed dataset is collected in the wild containing images from various ethnicity and distributed over varying age groups and gender. It consists of frontal, back, full and half body, side poses, mirror selfies with varying backgrounds and scale variations and may contain artifacts hiding partial or full face. Extensive experimentation is performed using full body, half body and face images only using different CNN backbones including VGG, Densenet and ResNet. Our experimental results demonstrate that full body images have produced better results than the other half body and facial images in the wild. Also in most of the experiments, joint learning of height, weight and BMI through multi-task learning has performed better than single task learning. The code and the dataset will soon be made publicly available. 
\end{abstract}

\begin{IEEEkeywords}
Human Weight, Human Height, Human BMI, Multitask Learning, Weight Height Data Set
\end{IEEEkeywords}


\section{Introduction}
\IEEEPARstart{A}{nalyzing} physical characteristics of a person such as weight and height are of much interest to doctors and researchers as well as to the general public in the current internet age. In order to capture various combinations of height and weight, Body Mass Index (BMI) \cite{mayer2017bmi, henderson2016perception} has been introduced as a useful measure which is defined as BMI = weight/height$^2$, when weight is measured in kilograms and height in meters. Healthcare professionals have divided BMI into four broad categories: underweight, normal, overweight, and obese. If BMI $< 18.5$, he/she is underweight, if 18.5 $\leq$ BMI $< 25$ then he/she is normal, if $25\leq$ BMI $< 30$ then he/she is overweight, and BMI $\geq 30$ is considered as obese. A healthy person must maintain a normal BMI, because an increased BMI is  correlated with many diseases such as type-2 diabetes \cite{lewis2019obesity, ling2019epigenetics, verma2017obesity}, gallbladder diseases, asthma \cite{xu2019elucidation}, sleep apnea \cite{narayanan2019asthma}, high blood pressure, angina \cite{uppunda2019association}, fatty liver,  cardiovascular diseases \cite{alpert2018obesity}, osteoarthritis, heart strokes and cancer \cite{arnold2016obesity, lin2018association}. Similarly, an underweight person may suffer from anorexia, type-1 diabetes, hyperthyroidism, amenorrhea, infertility, osteoporosis \cite{kanazawa2019overweight}, immunodeficiency \cite{nakagawa2018postnatal, ruffner2018complications}, and tuberculosis.

Instead of physically measuring a persons height and weight, various  techniques have been employed by medical practitioners to indirectly estimate a person's body composition and his BMI. 
These methods include Bioelectrical Impedance Analysis (BIA) \cite{khan2020relative, seo2018validation}, Bioimpedance Spectroscopy, Dual-energy X-ray absorptiometry (DXA) \cite{bjorkman2020associations, vermeiren2021comparison}, Air Displacement Plethysmography (ADP) \cite{shannon2019comparison, pellonpera2019body}, three-dimensional Photonic Scanner (3D-PS) \cite{ashby2020high, wells2019three}, Magnetic Resonance Spectroscopy (MRS) \cite{pasanta2018body}, and Positron Emission Tomography (PET) \cite{bini2020body}. In contrast to these techniques,  our proposed method provides an easy and efficient way to automatically estimate BMI of a person using his/her daily life pictures captured in the wild. 

On many social networking sites \footnote{Facebook, twitter, Instagram, }, people like to share pictures of their important life events. On some health and weight loss/gain related sites \footnote{https://www.reddit.com/r/progresspics/, https://myprogresspics.com/}, people share their pictures as well as corresponding weights and heights to become an inspiration for others as well as for themselves. In this work we utilize such labelled pictures for training deep neural networks to predict a person's weight, height and BMI from his/her pictures captured without a controlled environment. Once our networks are trained then we can employ the trained networks to predict these attributes of persons on social platforms at a mass scale. Our algorithm may potentially be used to monitor public health on large geographical areas \cite{bell2018detecting, kocabey2018using}. Such as system can also be used for self health assessments as well as  to warn family and friends about their health conditions. Furthermore, estimating height, and weight of a person can have significant applications in person re-identification, forensics, crime scene investigations, and surveillance and security systems.  

Due to the broad scope of BMI estimation from images, many researchers have proposed artificial intelligent and machine learning based approach for this purpose \cite{jiang2019body, kocabey2017face, dantcheva2018show}. However, most of these methods require images to be captured under control environment limiting their application to relatively narrow scope. In contrast, we propose a method to estimate BMI using images captured in the wild from different viewing conditions, camera resolutions, various scales, as well as selfies.

Many existing methods estimate BMI using frontal face images captured under controlled environment. These images exhibit  small variation and have a very constrained distribution of data. To address this issue, Jiang et al., \cite{jiang2019body}  collected paired frontal full-body images  of the same person having different BMI. They extracted hand-crafted features to train support vector regressor (SVR) and Gaussian process regressor (GPR) to predict BMI. Although the authors reported good results, however, they have not made their dataset publicly available limiting its application.
In contrast, we propose a dataset having a  large number of human body images   with the no-pair condition captured in the wild. Our collected dataset has significant variations and contains frontal/side poses, faces/full body/upper body images. We will soon make this dataset publicly available.  We train deep neural networks including VGG-16, DensNet-121 and ResNet-50 in multi-task fashion as well as single task learning for the estimation of height, weight, and BMI. We also pose the problem as classification of under-weight, normal, over-weight and obese. In addition to RGB and gray scale image representations, we have thoroughly experimented with pose affinity maps, depth maps and  human segmentation to improve BMI estimation. Our experimental results have demonstrated significantly improved performance as compared to the existing state of the art methods.

The main contributions of this work include:
\begin{itemize}\setlength\itemsep{0em}
    \item We propose a large scale publicly available data set of labeled 6105 images
    \item We propose the use of multiple modalities such as pose affinity maps, depth maps and  human segmentation, in addition to RGB and gray scale image representations for BMI estimation.
    \item We propose joint learning of weight, height and BMI through multi-task learning using deep neural networks.
    \end{itemize}

The rest of the paper is organized as follows: Section 2 gives a brief overview of the existing methods to estimate  weight, height, and/or BMI from faces or body images. Section 3 covers our proposed methodology including multi-task learning, single task learning and different modalities to improve weight, height and BMI estimations. Section 4 contains experiments and results while Section 5 concludes the paper.

\section{Review of Existing BMI Estimation Methods}
Due to a large scale applications of BMI for the prediction of health conditions, several automatic and efficient methods have been proposed. These methods  include BMI estimation from facial and full body images using hand crafted features and deep neural networks. We broadly categorize these methods as follows:

\subsection{BMI Estimation Using Facial Images}
Due to the availability of face images in social media websites such as Facebook, Instagram and Twitter, recently, there has been a growing interest in predicting a person's BMI from his face image.

\subsubsection{Face to BMI estimation using hand-crafted features}
Wen et al. \cite{wen2013computational} used hand-crafted features to estimate BMI of a person using least squares regression, support vector regression (SVR), and gaussian process regression (GPR) from only frontal facial images in MORPH-II dataset \cite{ricanek2006morph}. The authors used seven hand-crafted facial features including cheekbone to jaw width ratio, width to upper facial height ratio, perimeter to area ratio, eye size, lower face to face height ratio, face width to lower face height ratio and mean of eyebrow height, which were detected automatically using active shape model. They have reported good results, however,  their dataset/code is not publicly available. Barr et al. \cite{barr2018detecting} extracted hand-crafted features using facial landmarks and employed SVR for BMI estimations. Their dataset was captured in a controlled environment and consisted of frontal facial images in neutral emotion. They used the same seven facial features as proposed by \cite{wen2013computational}. Andersson et al. \cite{andersson2015gender} used anthropometric features and gait information to classify gender and BMI estimation on 106 persons whose images were taken using Microsoft Kinect sensor. Different combinations of attributes from the extracted anthropometric features were used and several classifiers such as SVM, K-nearest neighbour (KNN), and multi-layer perceptron (MLP) were employed. Pascali et al. \cite{pascali2016face} proposed a method to automatically extract geometric features from 3D facial images (using depth scanners) and predicted body weight. Affuso et al. \cite{affuso2018method} collected data from a selected set of 343 participants under controlled environment. For each participant, three full body images were captured including frontal, back and side poses. They extracted four features from the captured images including body volume, front curve, side curve, and body shape which are then fed to SVR to predict body fatness. Farina et al. \cite{farina2016smartphone} introduced the use of smartphones camera images to estimate body fat mass.  They collected a data set (digital images, and DXA) of 117 subjects. Estimating body fat mass can help in weight management and BMI estimation.

\subsubsection{Face to BMI estimation using Deep-Nets}
Kocabey et al. \cite{kocabey2017face} used VGG-Face and VGG-Net to extract facial features and epsilon SVR for BMI estimation over a dataset of facial images captured in the wild. They achieved excellent results compared to hand-crafted features. In their dataset, Americans and Africans were mostly obese, therefore their algorithm training is considered biased due to high prior probability. Jahandideh et al. \cite{jahandideh2018physical} trained ResNet-50 on the VGG-Face dataset and then fine-tuned on Faces in the Real World (FIRW) dataset to estimate body type, ethnicity, gender, height, and weight. A separate ResNet-50  was trained for height and weight classification. Jiang et al. \cite{jiang2019visual} predicted BMI from face images using deep features with SVR. They compared results of VGG-Face, Light CNN, center loss \cite{wen2016discriminative}, and Arcface \cite{deng2019arcface}. Dantcheva et al. \cite{dantcheva2018show} used ResNet-50 to predict height, weight, and BMI from the VIP attribute face images dataset.  Three separate networks were trained for height, weight, and BMI estimation. One limitation of their work is clear frontal face images of 1026 celebrities cropped using the Viola-Jones algorithm \cite{viola2001robust}. Another limitation is a biased dataset that contains 22 $\leq$ BMI $\leq$ 28 for around 80\% samples. Similarly, Jiang et al. \cite{jiang2020visual} estimated BMI from facial images using a label distribution. BMI was estimated as a discrete probability distribution and estimation results were demonstrated on FIW-BMI
\cite{jiang2019visual}, Morph-II, and VIP-attribute data sets. 

\subsection{BMI Estimation using Full Body Images}
Some of the recent studies have demonstrated the estimation of BMI from 2D and 3D images. Below, we discuss them briefly.

\subsubsection{Body to BMI estimation using Hand-crafted Features}
Jiang and Guo \cite{jiang2019body} work computes five anthropometric features (waist width to thigh width, waist width to hip width, waist width to head width, hip-width to head width, and the area between waist and hip) obtained from Body contour and skeleton joints (CSJ) detection network on frontal full-body images. After extracting anthropometric features, the authors apply support vector regression (SVR) and gaussian process regression (GPR) methods to predict BMI directly from images. Velardo et al. \cite{velardo2010weight} have collected anthropometric data from the National Health and Nutrition Examination Survey (NHANES) to estimate human body weight. However, the authors did not consider samples that exhibit weights below 35kg. and beyond 130 kilograms. The anthropometric features obtained from \cite{NHANES1999} are fed directly to regression models for weight estimation. The anthropometric features used by this method include height, upper leg length, calf circumference, upper arm length, upper arm circumference, waist circumference, and upper leg circumference. The authors also collected another dataset of 40 pictures (a frontal shot and a profile one) taken at a fixed distance from the camera and provided results as real-case analysis. Pfitzner et al. \cite{pfitzner2017evaluation} provides weight estimation on a dataset of 233 people. 
The authors computed anthropometric features from the frontal body RGB-D sensor data, and then employed artificial neural network (ANN) for weight estimation. Temperature features from a thermal camera are also included to better estimate the weight of a human body. In the subsequent work \cite{pfitzner2018body}, Pfitzner et al. proposed an approach to predict the weight of a person using RGBD camera sensor photos in three different poses: lying, walking, and standing.

\subsubsection{Body to BMI estimation using Deep-Nets}
Nahavandi et al. \cite{nahavandi2017skeleton} proposed to use ResNet-18 \& ResNet-34 to extract deep features and present a skeleton-free Kinect system to estimate the human body mass index. Specifically,  the authors generated depth images using a Kinect camera and encoded them into RGB for ResNet fine-tuning. The generated depth images are translated into a 3D point cloud to calculate the body surface area (BSA). BSA and other anthropometric features are then used to estimate the weight. Vakli et al. \cite{vakli2020predicting}, for the first time, investigated structural brain images/MRI to predict body mass index of a person. The authors used a CNN on MRI along with age and sex information to estimate BMI. Gunel et al. \cite{gunel2019face} have collected a new 2D images data set with unknown camera parameters, and scene geometry. The authors estimate the height of a person using pose, bone-lengths ratios, and facial features using different linear, shallow, and deep machine learning algorithms.

In contrast to the above mentioned works, we propose to \textit{publicly} release a large scale full body dataset for BMI estimation. To the best of our knowledge, no such dataset is \textit{publicly} available yet. We performed extensive experiments on this dataset and demonstrated that, in general, joint estimation of height, weight and BMI performs better than that of their estimation in the separate frameworks. 
Our thorough analysis using different modalities including RGB, depth-maps, pose-affinity maps, and masks employing various deep networks validate the proposed ideas.
\section{Review of Existing Image to BMI estimation Datasets}
In this section, we briefly review image to BMI estimation datasets which are discussed in the literature.  Table \ref{tab:datasetcompare} shows a comparison of our proposed \textit{Body2BMI-ITU} dataset with the other datasets which were proposed for the similar problem.

\subsection{Face2BMI Dataset}
Kocabey et al. \cite{kocabey2017face}  proposed a publicly available Face2BMI dataset for the estimation of BMI from face images. This dataset contains annotated faces of 2103 pairs from reddit.com, with  a total number of 4206 images with past and current weights. This dataset contains 2438 males and 1768 females. The authors define seven BMI categories where 7 subjects have underweight BMI, 680 have normal BMI, 1151 have overweight BMI, 941 are moderately obese, 681 are severely obese, and 746 are very severely obese.

\subsection{FIW-BMI Dataset}
Jiang et al. \cite{jiang2019visual} proposed this dataset which contains 7930 wild face images of 4881 individuals along with the labels of gender, height, and weight. In this dataset, there exist 3192 males (5197 images) and 1689 females (2733 images). Among these face images, 43 are underweight, 1662 are normal, 2455 are overweight, and 3770 are obese. This dataset is not publicly available.

\subsection{Morph-II Data set}
The Morph II database \cite{ricanek2006morph} contains 55608 mugshot-style frontal view face images of 13673 subjects of the age between 16 and 99 years where 47057 images correspond to male persons and 8551 to female persons.  
The dataset has the age, gender, ethnicity, weight and height labels and  55\% of the data lies under 'normal' BMI range.  
\subsection{VIP-Attribute Dataset} This publicly available dataset contains 1026 celebrities' frontal face images. It contain 523 male  and equal number of  female  high resolution cropped images with no background. 
 \begin{table}[H]
    \scriptsize
    \centering
    \caption{A comparison of existing datasets with proposed datasets: ITU-BMI, ITU-BMI\_F (only face images), ITU-BMI\_U (all images with upper-body), ITU-BMI\_B (only images with full-body) }
    \setlength\tabcolsep{1pt}
    \begin{tabular}{c| c c c c c c}
        \hline
         {\bf Data set}& {\bf \# Images} & {\bf Faces} & {\bf Upper-body} & {\bf Full-body} & {\bf Availability}\\
         \hline
         Face2BMI \cite{kocabey2017face}& 4206 &\checkmark&-&-&\checkmark \\[0.5ex]
         FIW-BMI \cite{jiang2019visual}& 7930 & \checkmark&-&-&-\\[0.5ex]
         Morph-II \cite{ricanek2006morph} & 29033 & \checkmark&-&-&-\\[0.5ex]
         Dantcheva et al. \cite{dantcheva2018show} & 1026 &\checkmark&-&-&\checkmark\\[0.5ex]
         Jiang et al. \cite{jiang2019body}& 5900 &-&-&\checkmark&- \\[0.5ex]
         \hline
         ITU-BMI\_F & 3372 & \checkmark & - & -& \checkmark\\[0.5ex]
         ITU-BMI\_U & 5300 & -&\checkmark&-&\checkmark\\[0.5ex]
         ITU-BMI\_B & 537 &-&-&\checkmark&\checkmark\\[0.5ex]
         ITU-BMI & 6105 & \checkmark & \checkmark & \checkmark & \checkmark\\[0.5ex]
         \hline
    \end{tabular}
    \label{tab:datasetcompare}
\end{table}
 
\subsection{Visual-body-to-BMI}
Jian et al. \cite{jiang2019body}  collected this dataset from `reddit.com'  consisting of frontal full-body images. The dataset contains 5900 images of 2950 subjects  along with gender, height, and weight labels. Each subject contains a pair of images showing past and present height, and weight. In terms of BMI, the under-weight category has 46 images, 1416 are normal, 1863 are overweight, and 2575 are in  obese category. This data set lacks diversity of poses and partial body images because it contains only frontal  full-body images. This dataset is not publicly available.
\section{Proposed Body2BMI-ITU Dataset}
\begin{figure}[!t]
\begin{subfigure}[b]{\linewidth}  
\subfloat a.
\includegraphics[width=1.1cm, height=1.7cm]{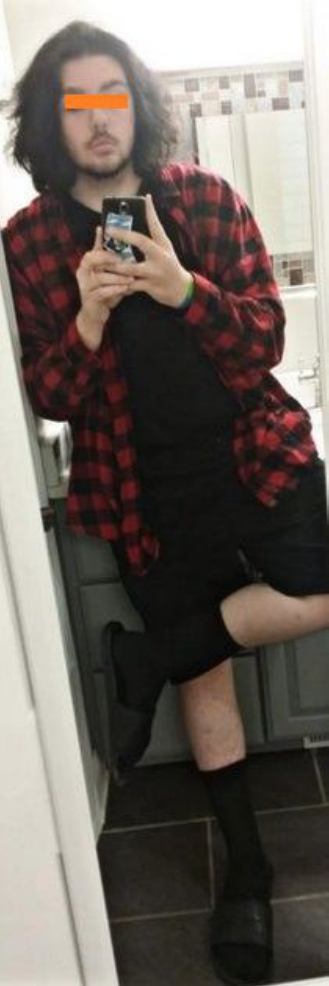}{}
\includegraphics[width=1.1cm, height=1.7cm]{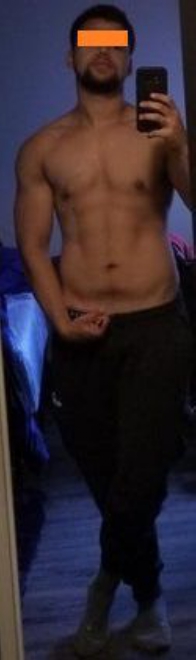}{}
\includegraphics[width=1.1cm, height=1.7cm]{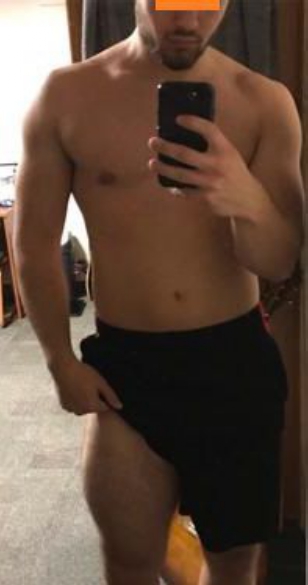}{}
\includegraphics[width=1.1cm, height=1.7cm]{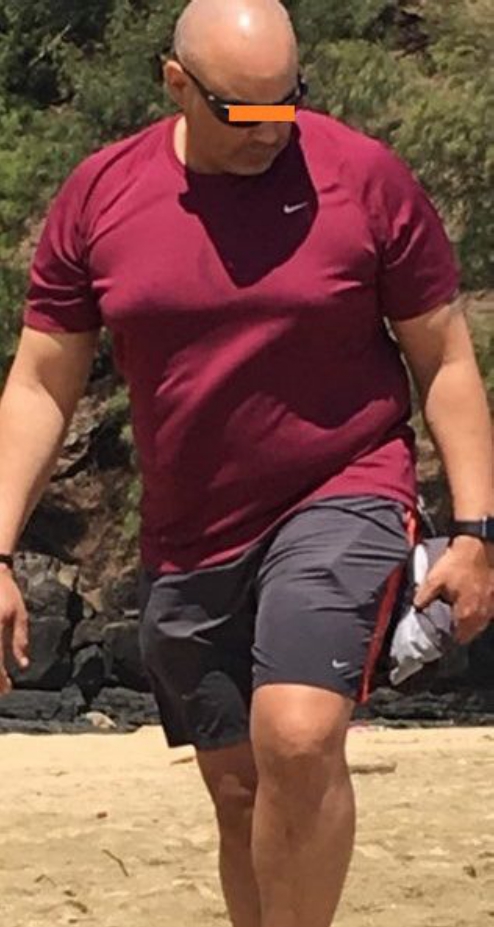}{}
\includegraphics[width=1.1cm, height=1.7cm]{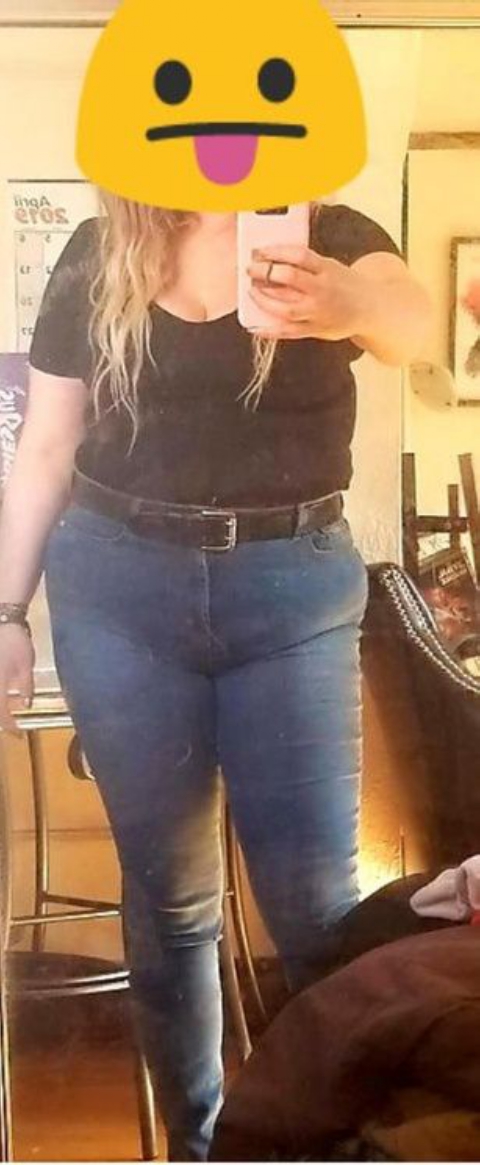}{}
\includegraphics[width=1.1cm, height=1.7cm]{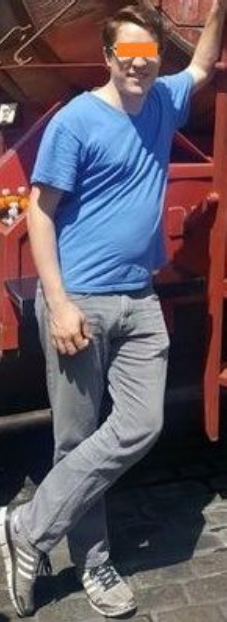}{}
\includegraphics[width=1.1cm, height=1.7cm]{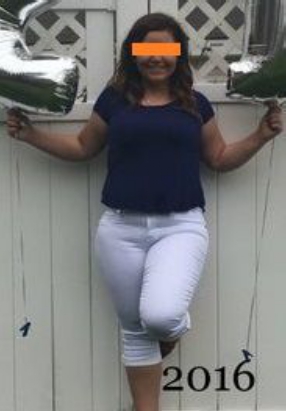}{}
\end{subfigure}
\newline
\begin{subfigure}[b]{\linewidth} 
\subfloat b.
\includegraphics[width=1.1cm, height=1.7cm]{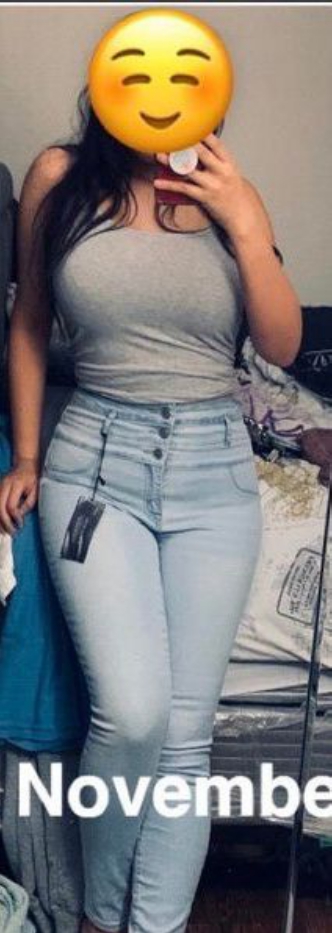}{}
\includegraphics[width=1.1cm, height=1.7cm]{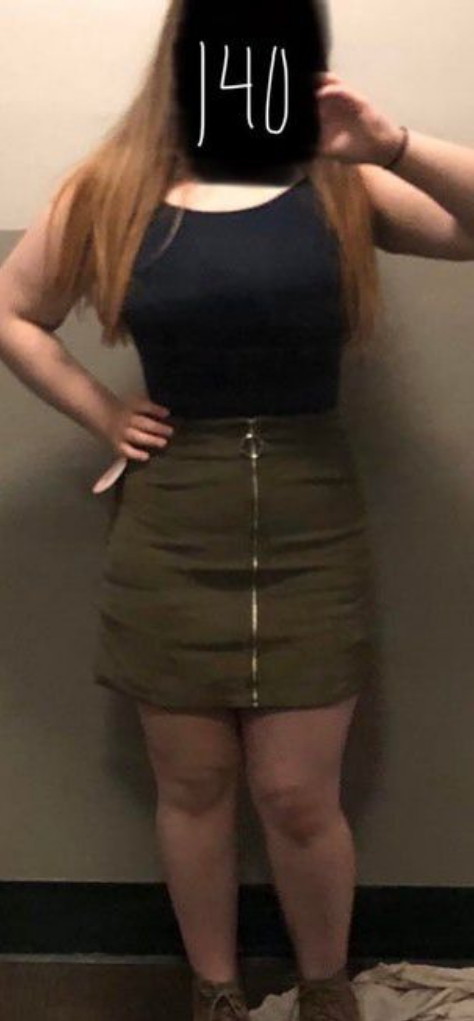}{}
\includegraphics[width=1.1cm, height=1.7cm]{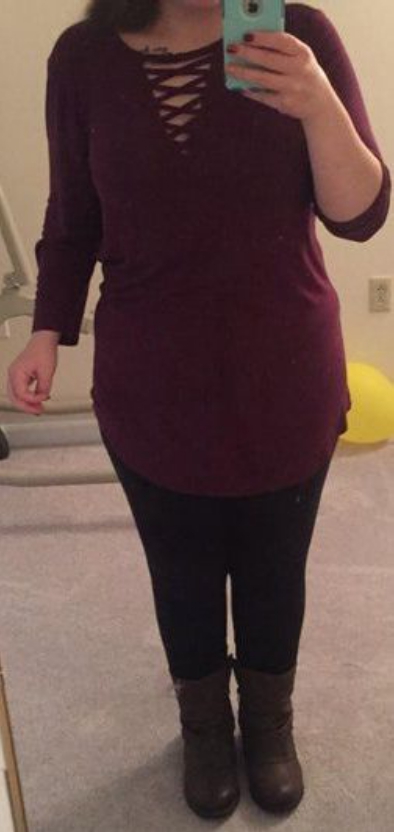}{}
\includegraphics[width=1.1cm, height=1.7cm]{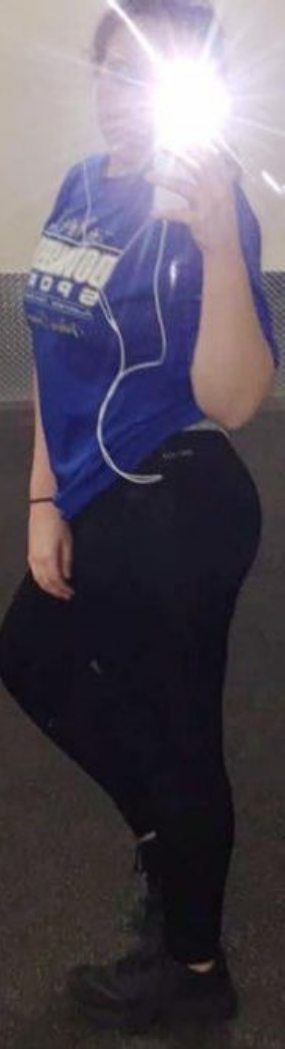}{}
\includegraphics[width=1.1cm, height=1.7cm]{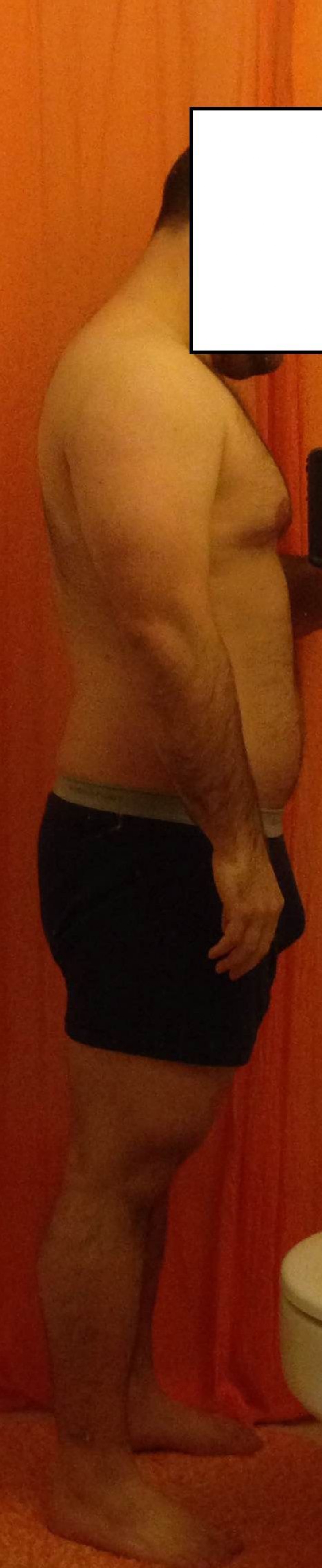}{}
\includegraphics[width=1.1cm, height=1.7cm]{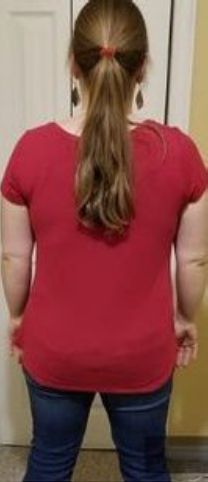}{}
\includegraphics[width=1.1cm, height=1.7cm]{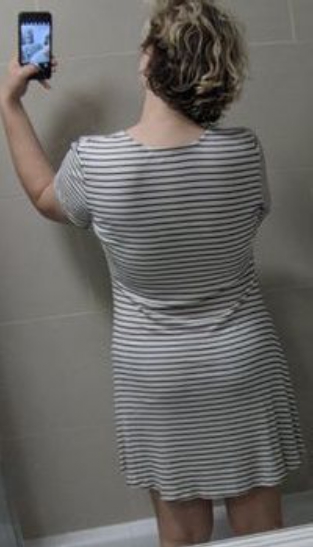}{}
\end{subfigure}
\newline
\begin{subfigure}[b]{\linewidth}
\subfloat c.
\includegraphics[width=1.1cm, height=1.7cm]{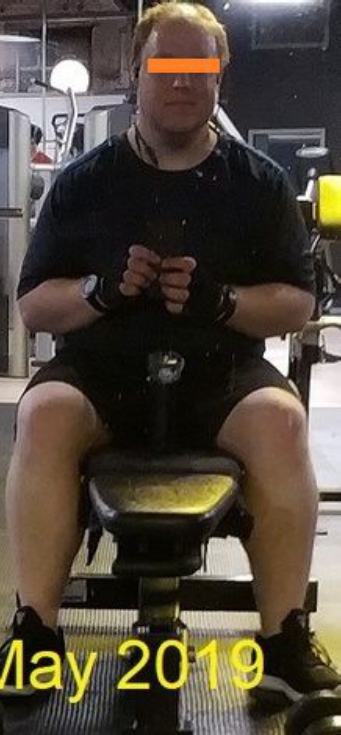}{}
\includegraphics[width=1.1cm, height=1.7cm]{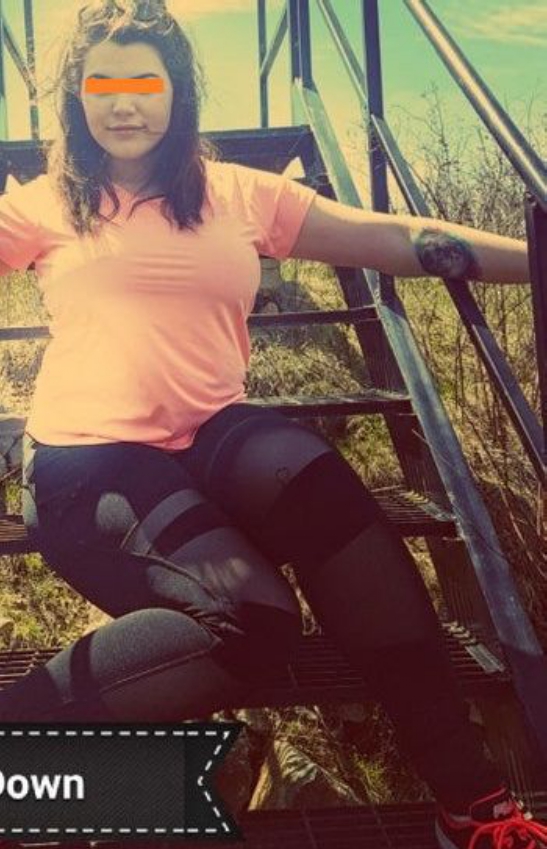}{}
\includegraphics[width=1.1cm, height=1.7cm]{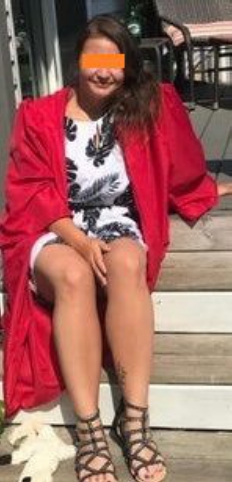}{}
\includegraphics[width=1.1cm, height=1.7cm]{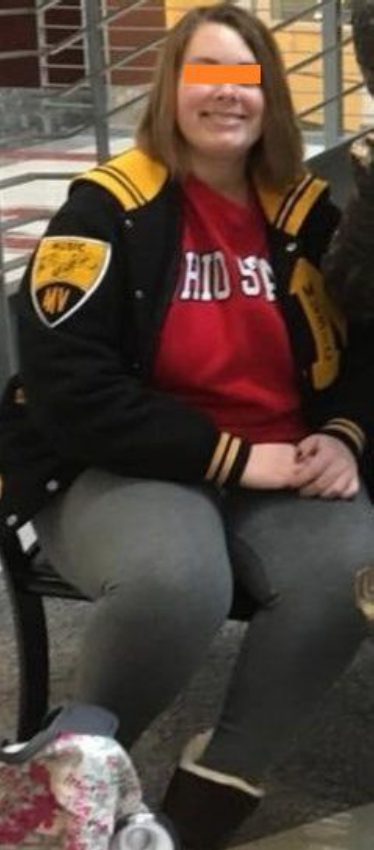}{}
\includegraphics[width=1.1cm, height=1.7cm]{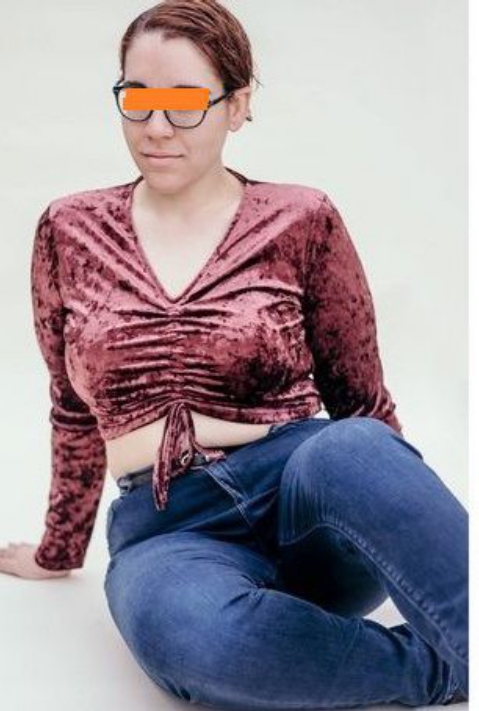}{}
\includegraphics[width=1.1cm, height=1.7cm]{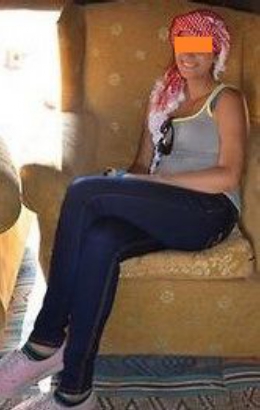}{}
\includegraphics[width=1.1cm, height=1.7cm]{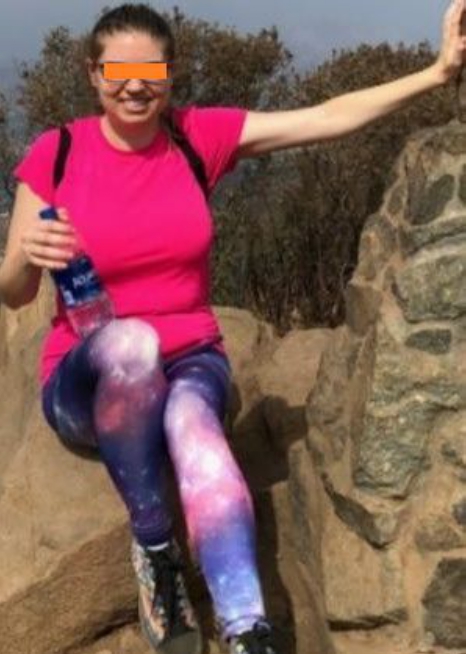}{}
\end{subfigure}
\newline
\begin{subfigure}[b]{\linewidth}
\subfloat d.
\includegraphics[width=1.1cm, height=1.7cm]{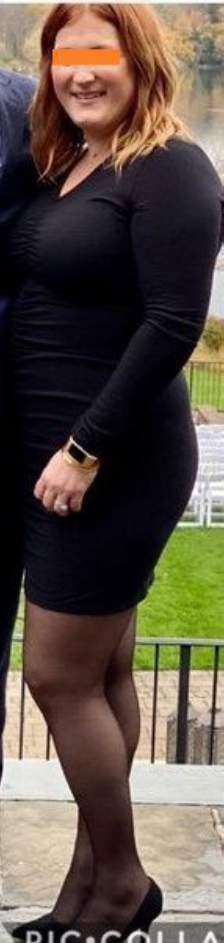}{}
\includegraphics[width=1.1cm, height=1.7cm]{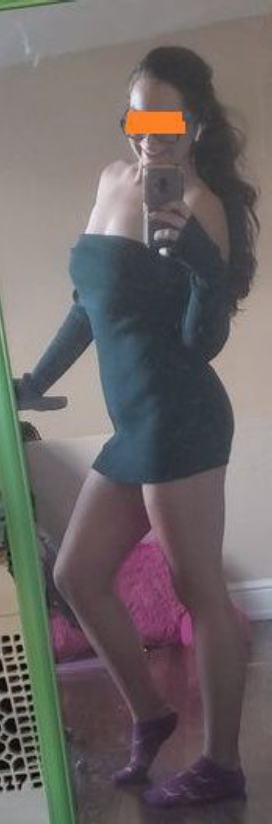}{}
\includegraphics[width=1.1cm, height=1.7cm]{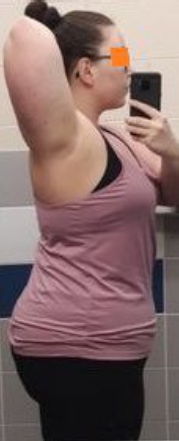}{}
\includegraphics[width=1.1cm, height=1.7cm]{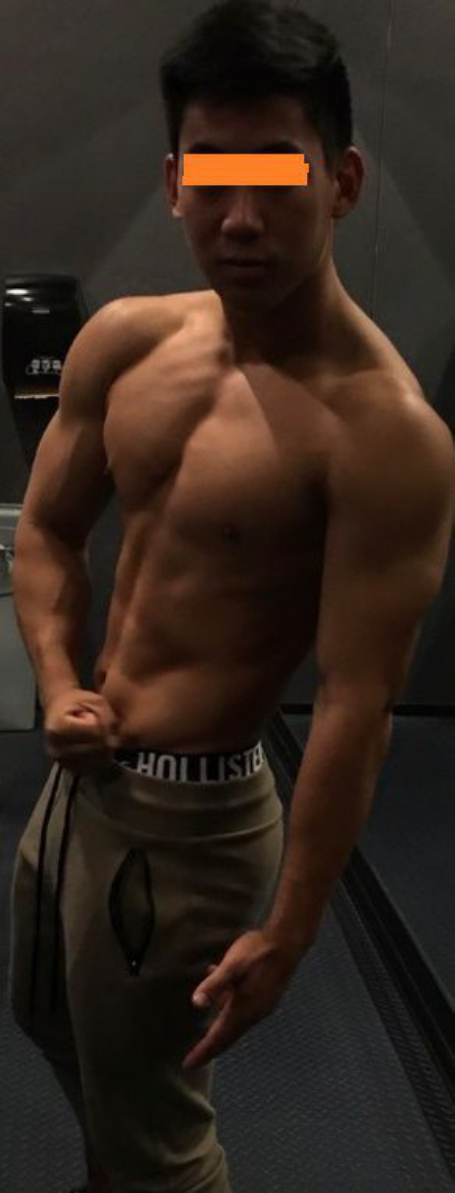}{}
\includegraphics[width=1.1cm, height=1.7cm]{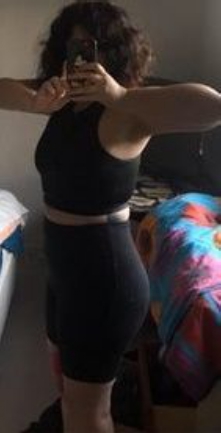}{}
\includegraphics[width=1.1cm, height=1.7cm]{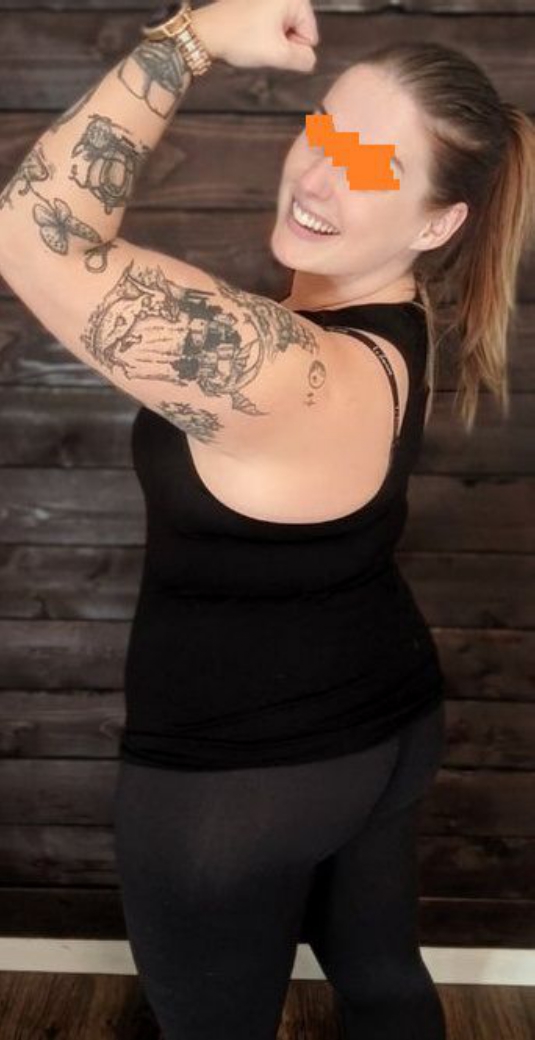}{}
\includegraphics[width=1.1cm, height=1.7cm]{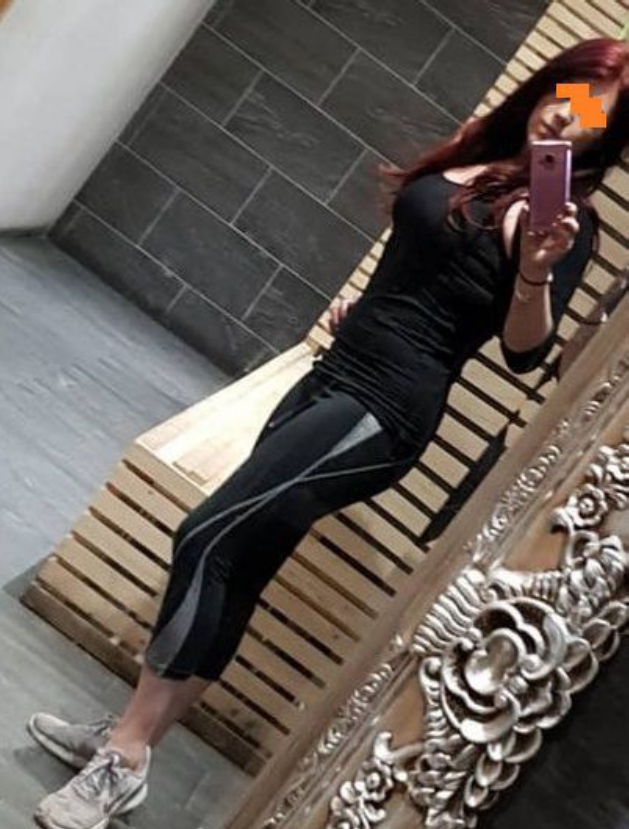}{}
\end{subfigure}
\newline
\begin{subfigure}[b]{\linewidth}
\subfloat e.
\includegraphics[width=1.1cm, height=1.7cm]{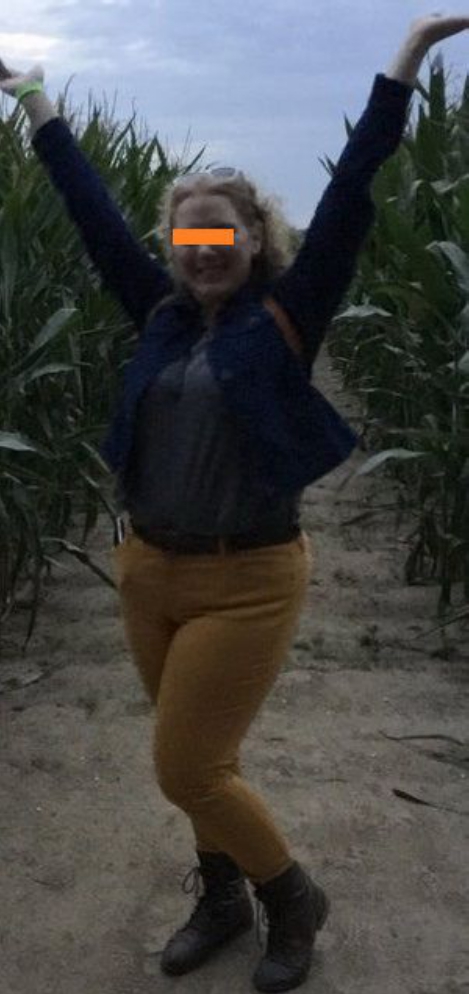}{}
\includegraphics[width=1.1cm, height=1.7cm]{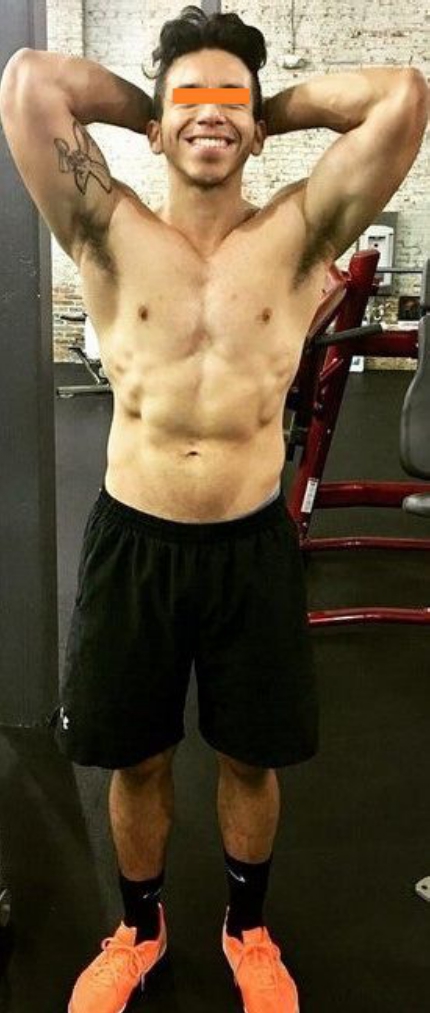}{}
\includegraphics[width=1.1cm, height=1.7cm]{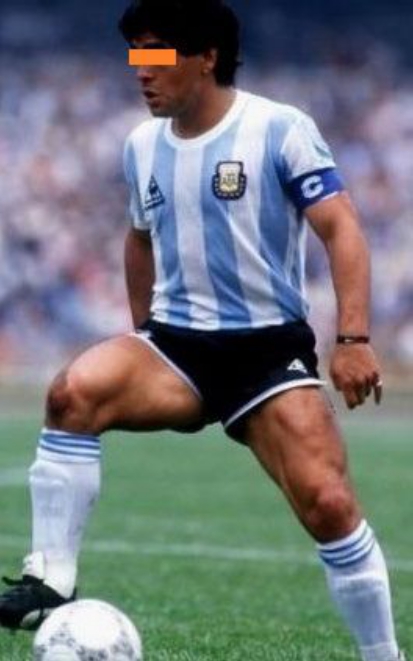}{}
\includegraphics[width=1.1cm, height=1.7cm]{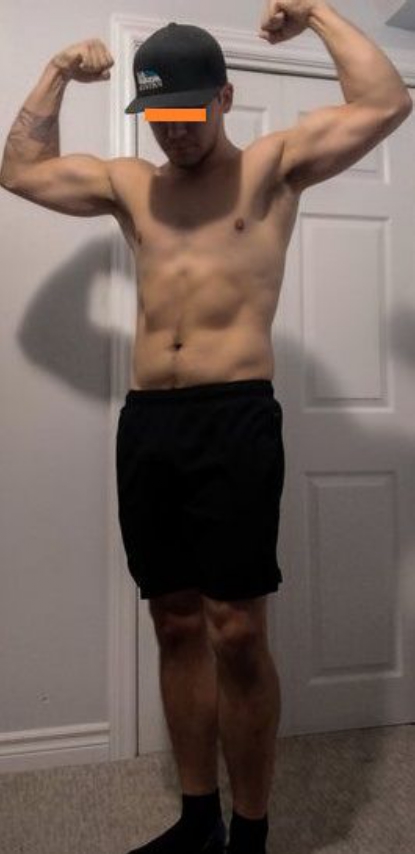}{}
\includegraphics[width=1.1cm, height=1.7cm]{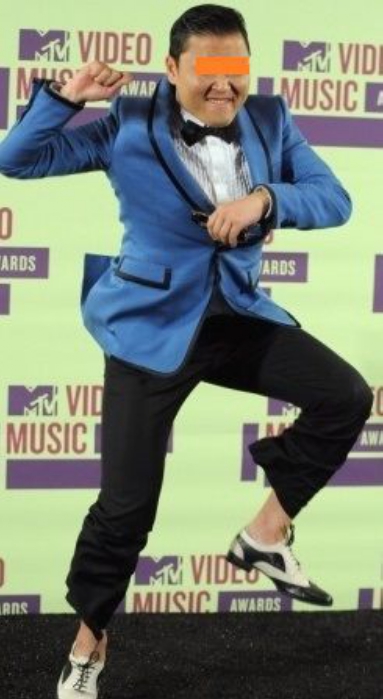}{}
\includegraphics[width=1.1cm, height=1.7cm]{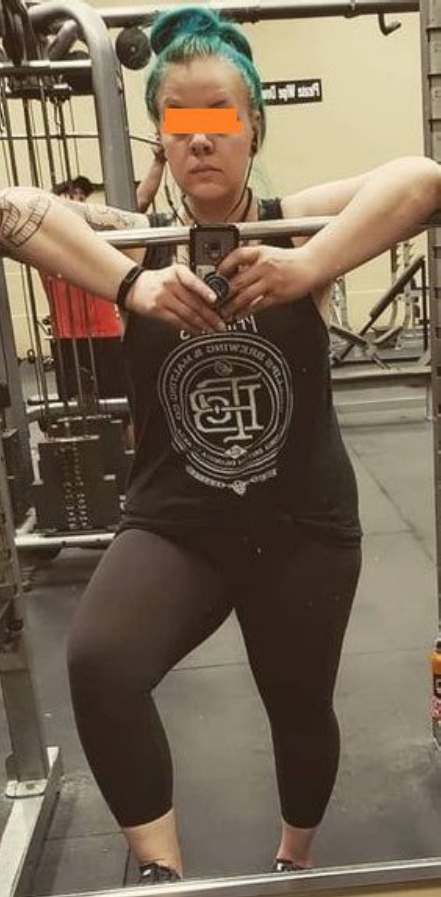}{}
\includegraphics[width=1.1cm, height=1.7cm]{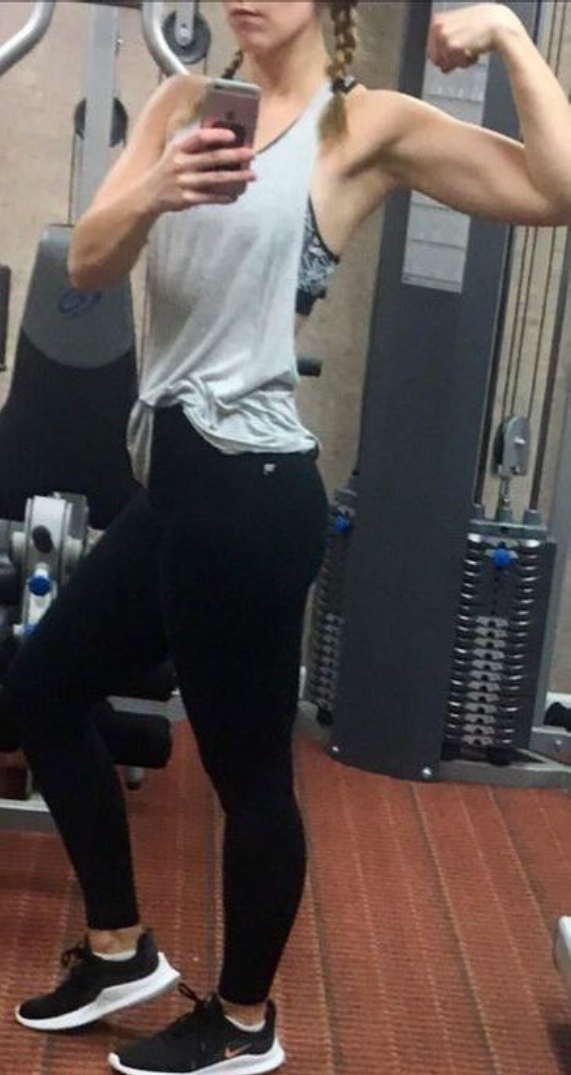}{}
\end{subfigure}
\newline
\begin{subfigure}[b]{\linewidth}
\subfloat f.
\includegraphics[width=1.1cm, height=1.7cm]{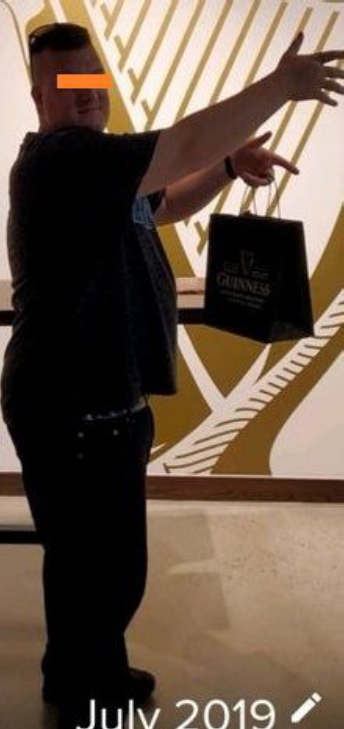}{}
\includegraphics[width=1.1cm, height=1.7cm]{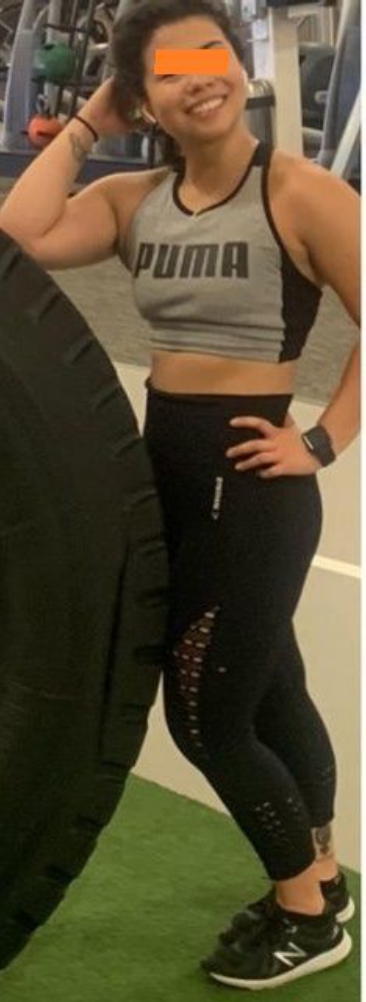}{}
\includegraphics[width=1.1cm, height=1.7cm]{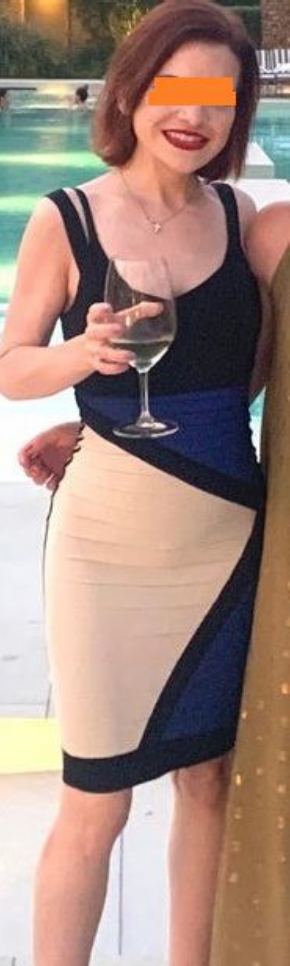}{}
\includegraphics[width=1.1cm, height=1.7cm]{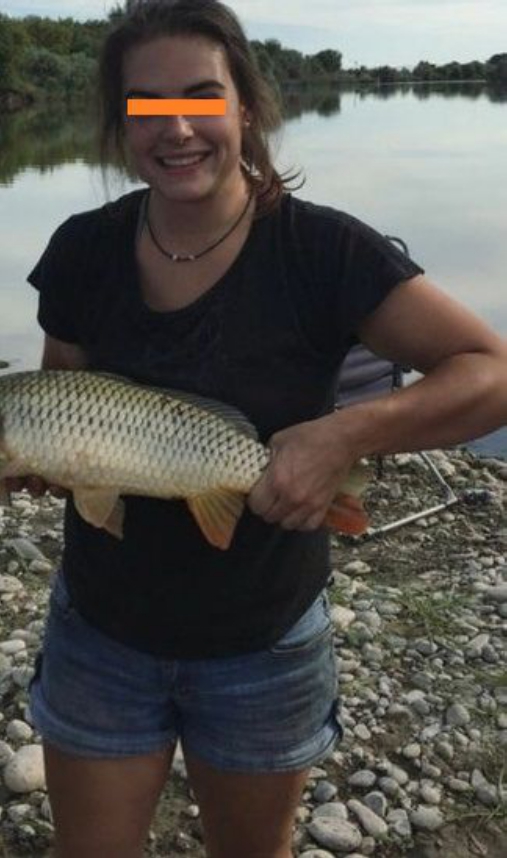}{}
\includegraphics[width=1.1cm, height=1.7cm]{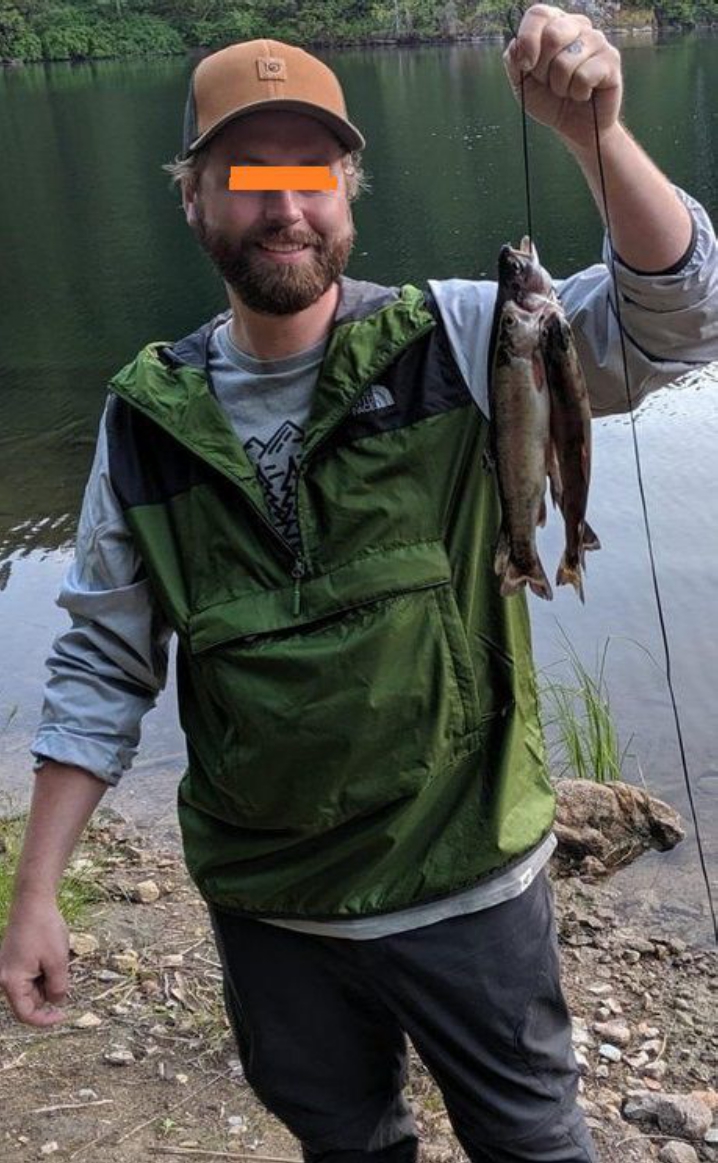}{}
\includegraphics[width=1.1cm, height=1.7cm]{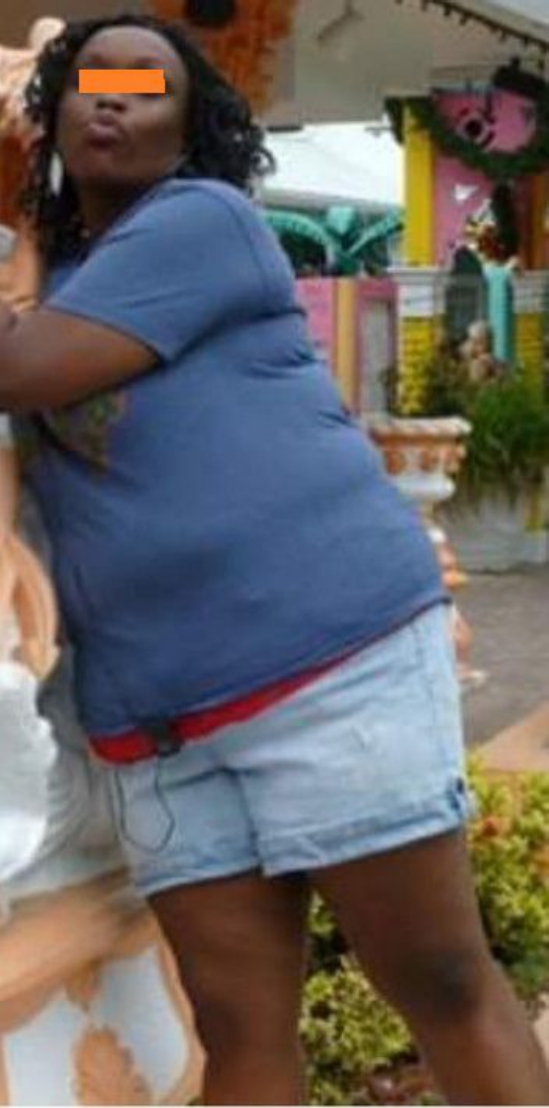}{}
\includegraphics[width=1.1cm, height=1.7cm]{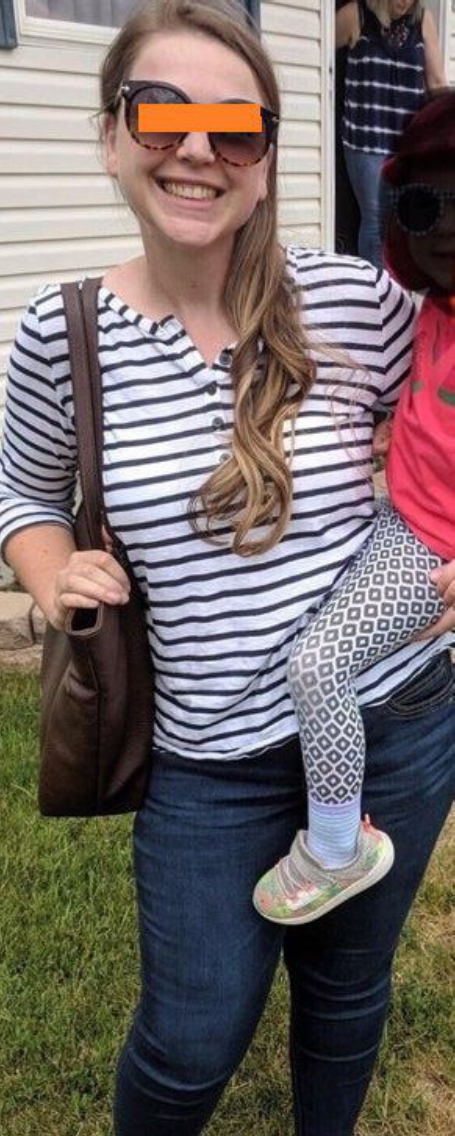}{}
\end{subfigure}
\newline
\begin{subfigure}[b]{\linewidth}
\subfloat g.
\includegraphics[width=1.1cm, height=1.7cm]{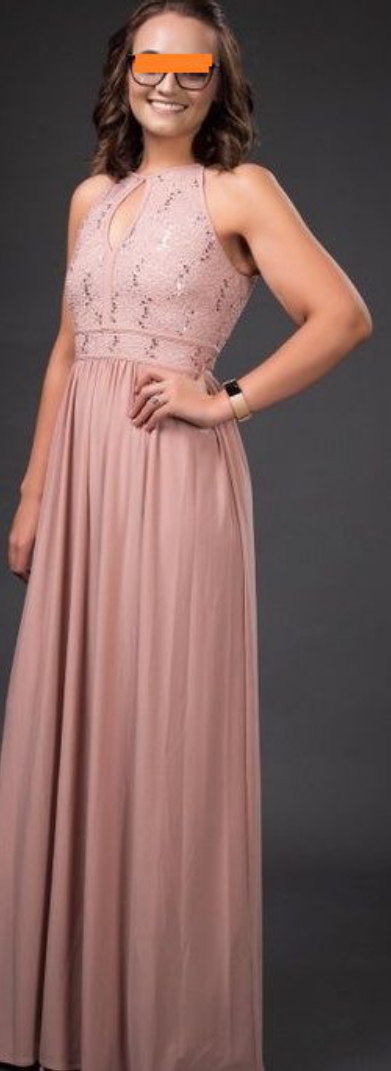}{}
 \includegraphics[width=1.1cm, height=1.7cm]{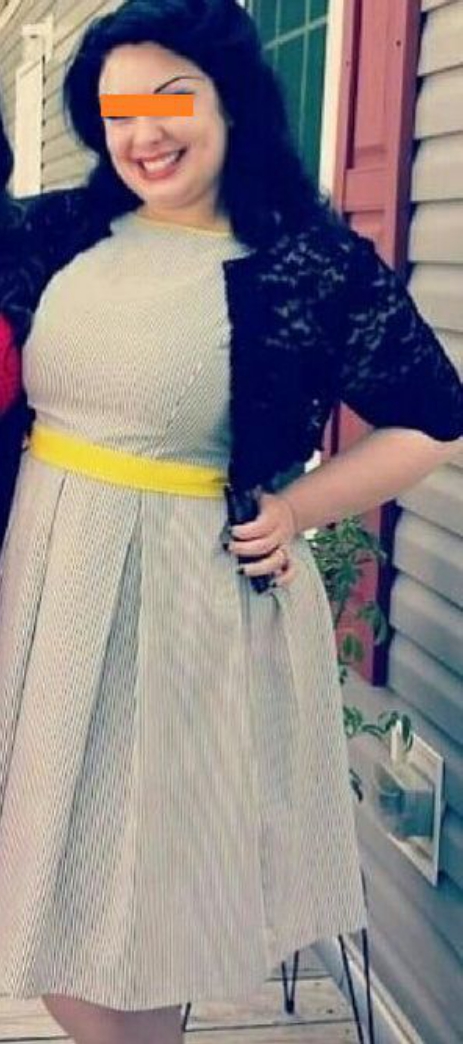}{}
\includegraphics[width=1.1cm, height=1.7cm]{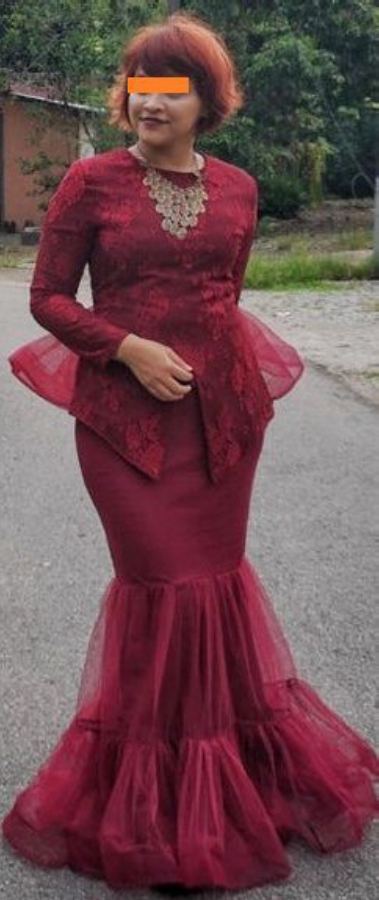}{}
\includegraphics[width=1.1cm, height=1.7cm]{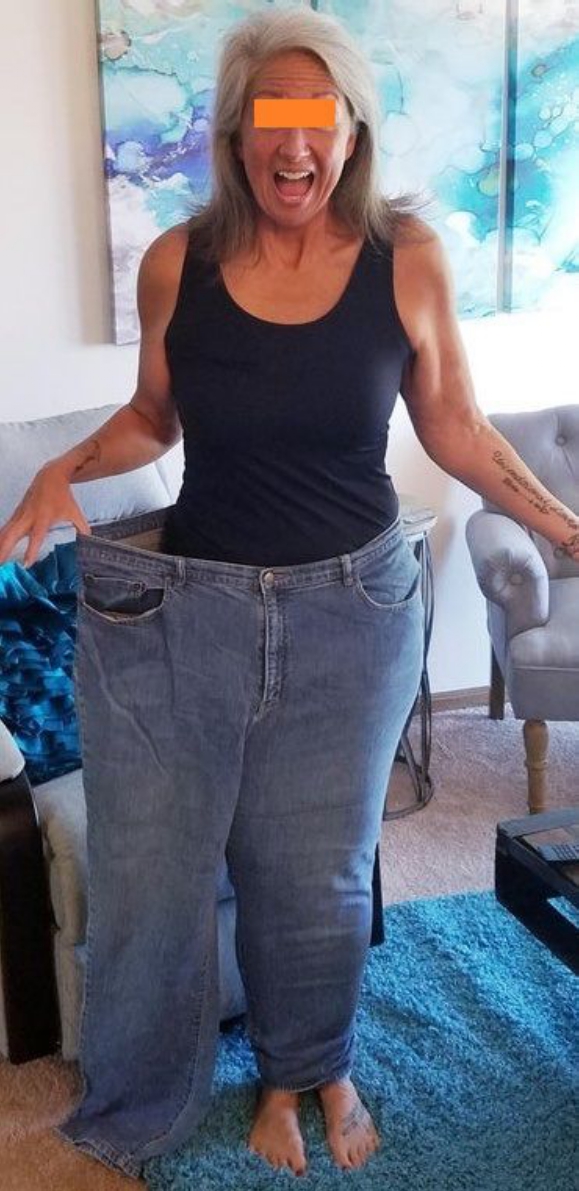}{}
 \includegraphics[width=1.1cm, height=1.7cm]{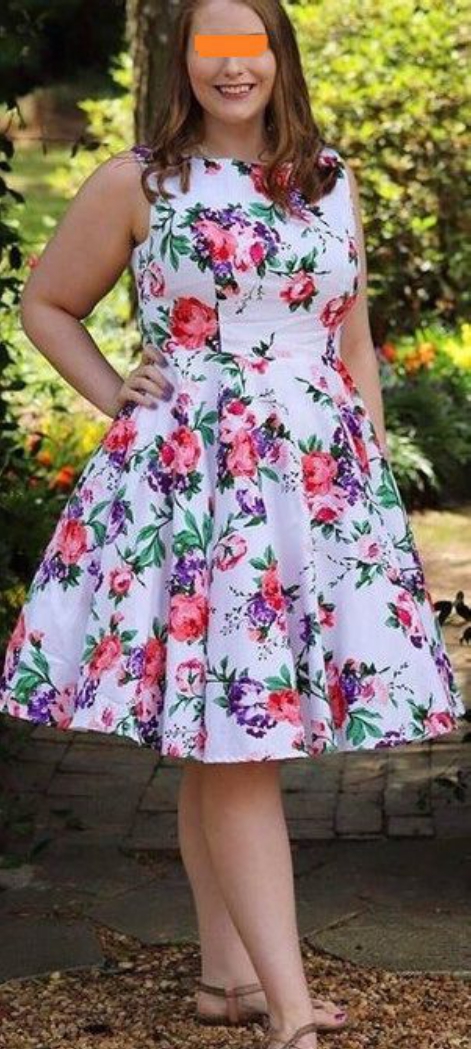}{}
 \includegraphics[width=1.1cm, height=1.7cm]{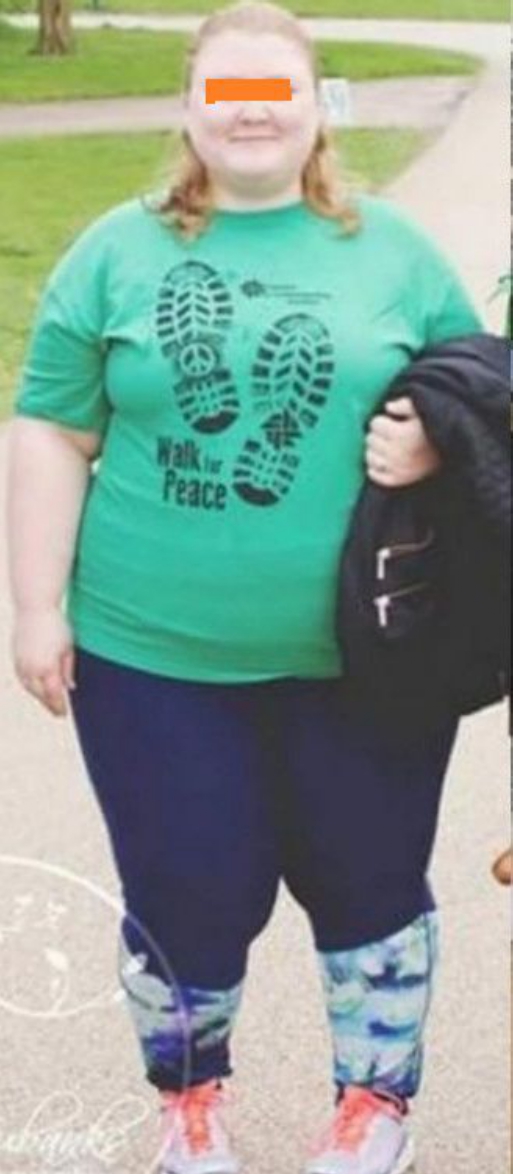}{}
\includegraphics[width=1.1cm, height=1.7cm]{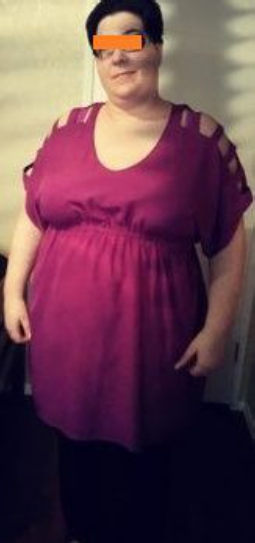}{}
\end{subfigure}
\newline
\begin{subfigure}[b]{\linewidth}
\subfloat h.
\includegraphics[width=1.1cm, height=1.7cm]{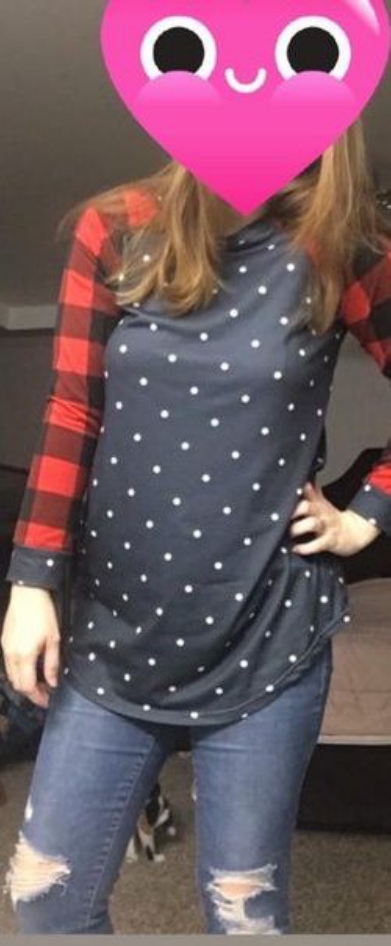}{}
\includegraphics[width=1.1cm, height=1.7cm]{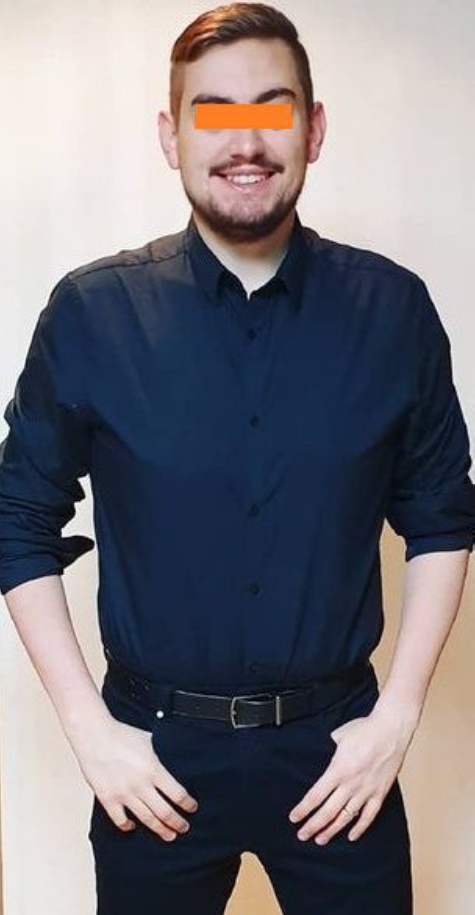}{}
\includegraphics[width=1.1cm, height=1.7cm]{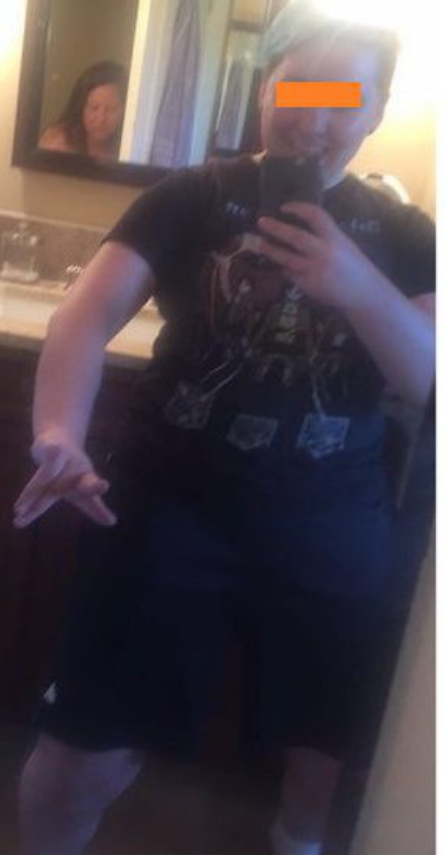}{}
\includegraphics[width=1.1cm, height=1.7cm]{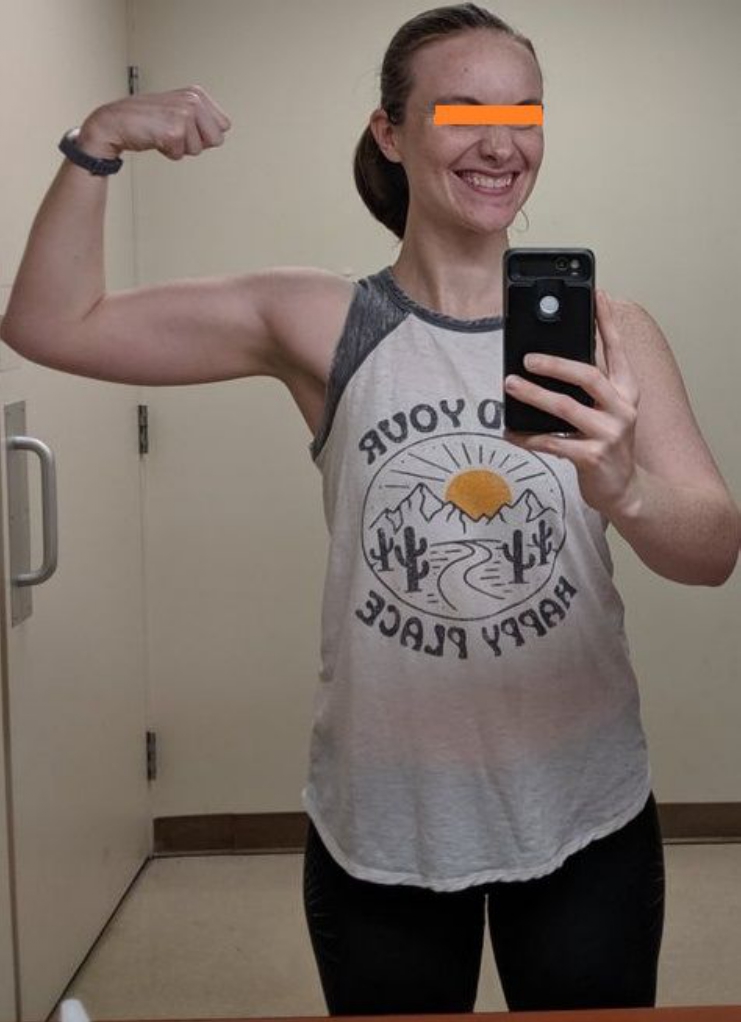}{}
\includegraphics[width=1.1cm, height=1.7cm]{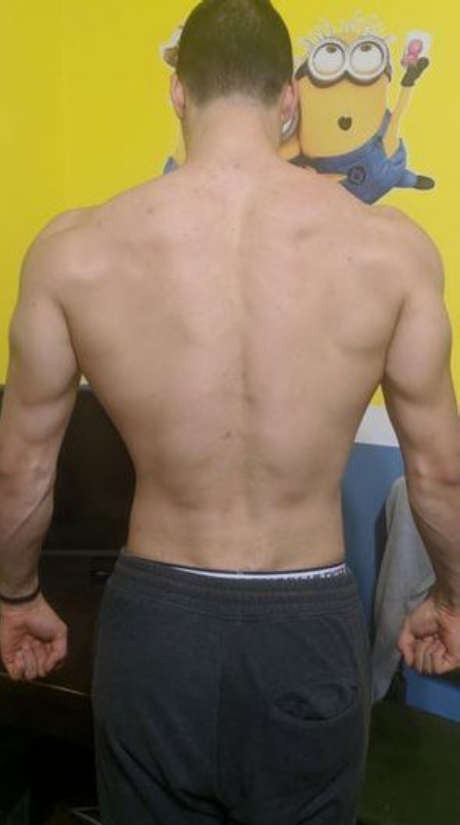}{}
\includegraphics[width=1.1cm, height=1.7cm]{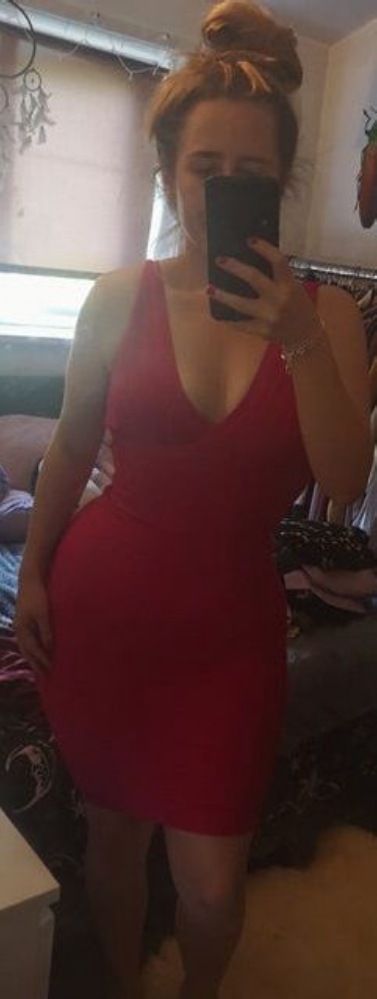}{}
\includegraphics[width=1.1cm, height=1.7cm]{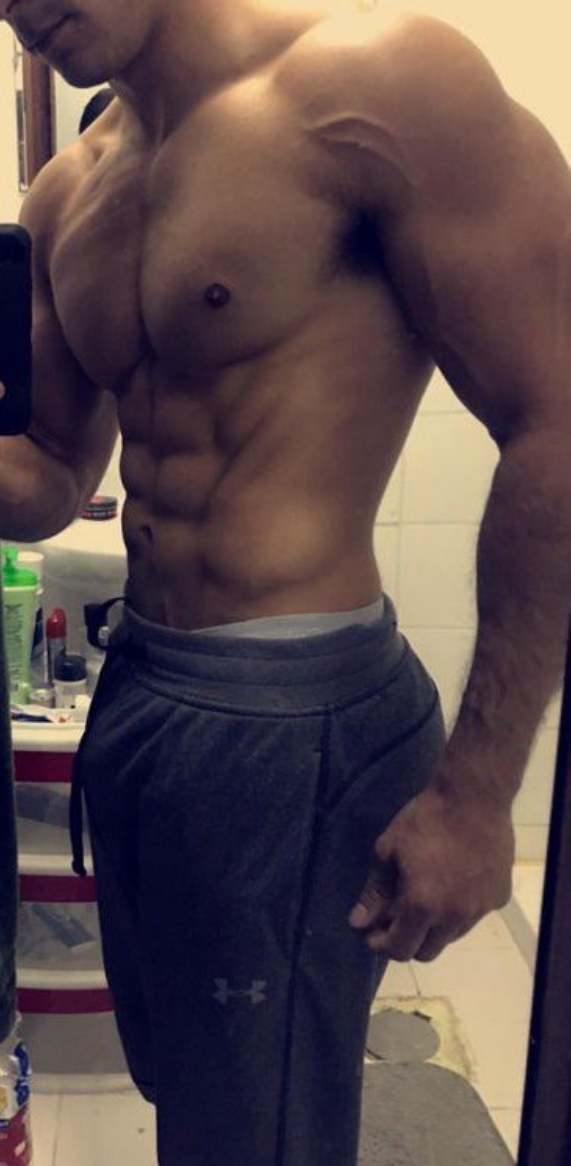}{}
\end{subfigure}
\newline
\begin{subfigure}[b]{\linewidth}
\subfloat i.
\includegraphics[width=1.1cm, height=1.7cm]{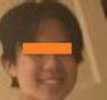}{}
\includegraphics[width=1.1cm, height=1.7cm]{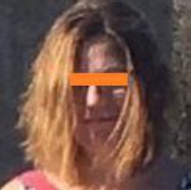}{}
\includegraphics[width=1.1cm, height=1.7cm]{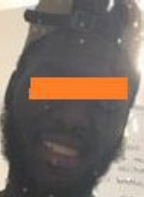}{}
\includegraphics[width=1.1cm, height=1.7cm]{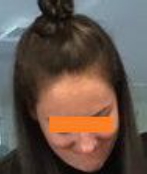}{}
\includegraphics[width=1.1cm, height=1.7cm]{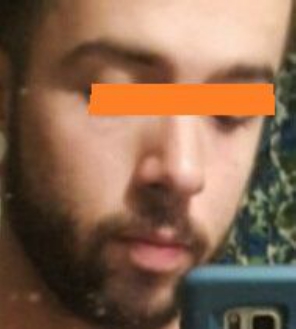}{}
\includegraphics[width=1.1cm, height=1.7cm]{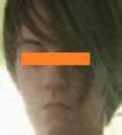}{}
\includegraphics[width=1.1cm, height=1.7cm]{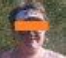}{}
\end{subfigure}
\newline
\begin{subfigure}[b]{\linewidth}
\subfloat j.
\includegraphics[width=1.1cm, height=1.7cm]{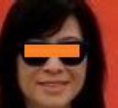}{}
\includegraphics[width=1.1cm, height=1.7cm]{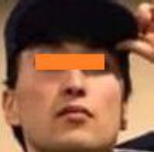}{}
\includegraphics[width=1.1cm, height=1.7cm]{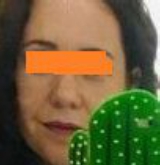}{}
\includegraphics[width=1.1cm, height=1.7cm]{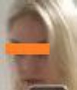}{}
\includegraphics[width=1.1cm, height=1.7cm]{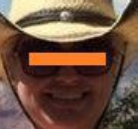}{}
\includegraphics[width=1.1cm, height=1.7cm]{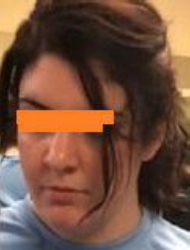}{}
\includegraphics[width=1.1cm, height=1.7cm]{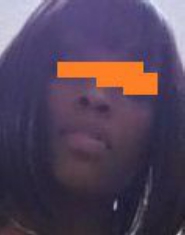}{}
\end{subfigure}
\newline
\caption{Sample images from ITU-BMI dataset: (a) examples with leg's occlusion. (b)  missing faces, (c)  sitting  position makes height estimation challenging, (d) side-poses, (e) full body images with wide pose variations, (f) holding objects,  (g) some outfits make weight estimation more challenging, (h) frontal and side-pose upper-body images (correspond to ITU-BMI\_U data set), (i \& j) face images only (correspond to ITU-BMI\_F data set) }
\label{fig:ann}
\end{figure}

\begin{figure*}[ht]
    \begin{subfigure}[b]{0.3\linewidth}
        \includegraphics[width=1\linewidth]{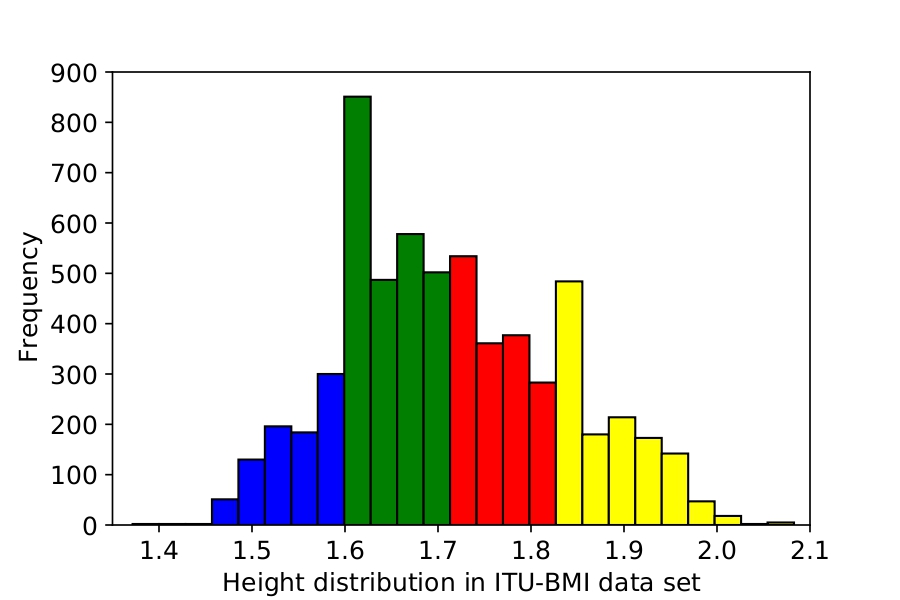}
        \subcaption{}
    \end{subfigure}
    ~ 
     \begin{subfigure}[b]{0.3\linewidth}
        \includegraphics[width=1\linewidth]{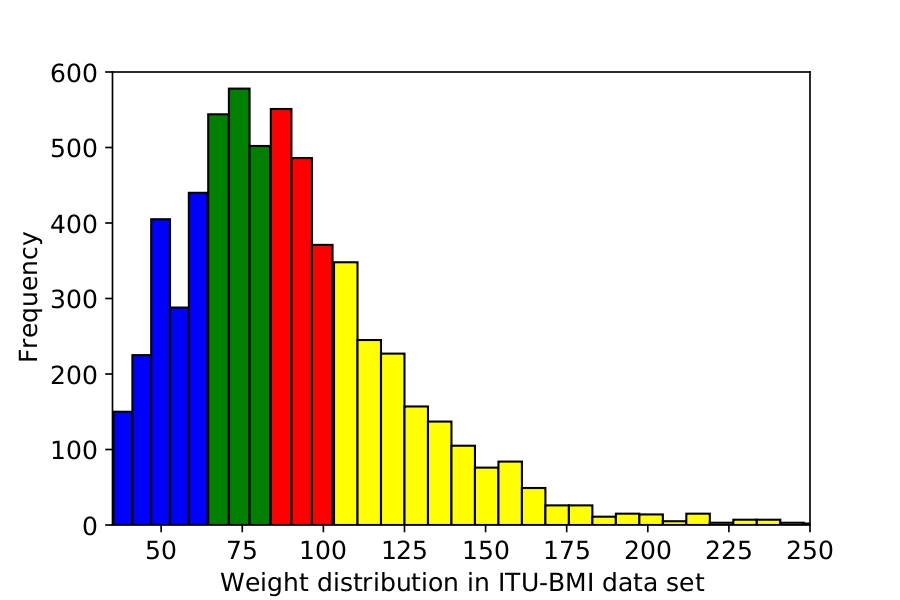}
        \subcaption{}
    \end{subfigure}
    ~ 
     \begin{subfigure}[b]{0.3\linewidth}
        \includegraphics[width=1\linewidth]{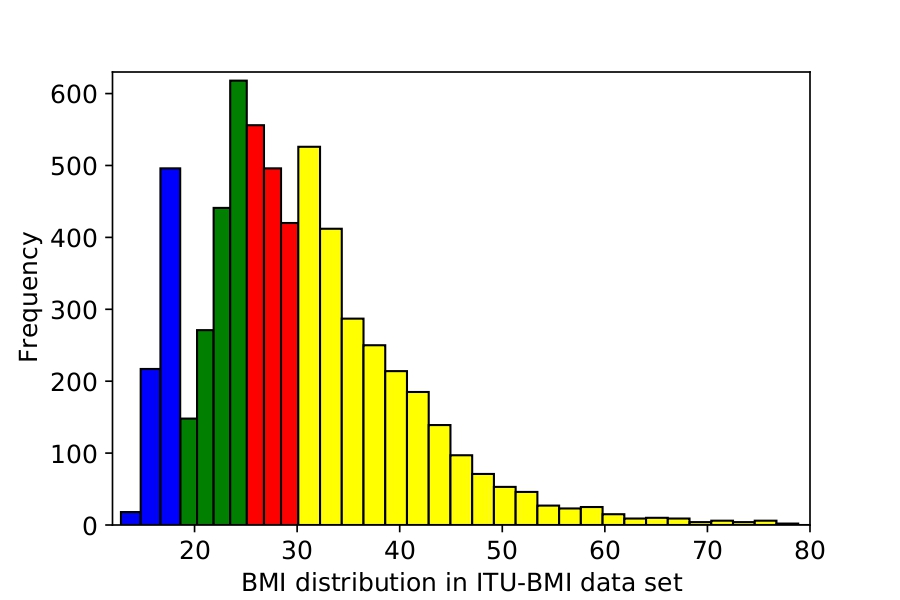}
        \subcaption{}
    \end{subfigure}
\caption{Statistics of the proposed ITU-BMI dataset: (a) distribution of person heights in meters,  (b) distribution of person weights in kilograms,  (c) distribution of BMI.}
\label{heightbarchart}
\end{figure*}

We have collected 6105 daily-life images from two famous websites `www.reddit.com' and `imgur.com'. At these websites,  people share their images to show weight loss or weight gain progress to inspire others. In our dataset, each image is annotated  with gender, weight, height and BMI information.
As show in Figure \ref{fig:ann}, our dataset contains diverse attributes consisting of selfies, camera pictures, frontal body images, back body images, side poses, only faces, full-body images, half body images (include hip joints). The diversities in poses include people sitting, lying, wearing heavy clothes and caps, may have occluded their faces through cartoons or mobile phones, blackened whole faces or eyes only, may carry food items, pets, and other unavoidable objects in the image.
To further illustrate the diversity in our dataset, we have separately shown height, weight and BMI distribution of our collected dataset in Figure \ref{heightbarchart}. In figure 2(a) height of subjects ranges between 1.40 meters to 2.20 meters, showing peak values from 1.63 m to 1.73 m. In figure 2(b), four classes of weights are shown ranging from 34 kgs to 250kgs. In figure 2(c) colored bars show four BMI classes including underweight (BMI $\leq$ 18.5), normal (18.5 $<$ BMI $\leq$ 25), overweight (25 $<$ BMI $\leq$ 30), and obese (BMI $>$ 30). 

The dataset collection becomes challenging if the user uploads one image containing both weight loss or weight gain pictures as shown in Fig. \ref{fig:crop}. In such cases, the target person had to be found manually employing human detection. One such instance is shown Fig. \ref{fig:crop}a. \& \ref{fig:crop}b. in which uploaded picture contained two or more images of the same person. We use Masked RCNN \cite{he2017mask} to separate the humans and then manually corrected the annotation. As shown in Fig. \ref{fig:crop}b, additional persons also appear in the uploaded pictures and common person is found manually. Other challenges include manual correction of data format and correcting the incorrect correspondence between image and labels. Finally, since different people prefer different units of measurement e.g., kilograms or pounds for weight and centimeters or feet for height, we make all ground consistent: kilograms for weight and meters for height.{Because the collected data is posted at a health forum, the uploaded images are with exact measurements to keep track of a person's success. A person has to register himself to make an account at that forum, and then post their success journey. This forum also includes other suggestions, like keto diet , and everyone is open to share his story, receive comments and chat with others. People measure and upload the exact BMI at the start of their journey to a healthy body with a collage of another taken at some progress point. 
\subsection{Data for Deep Learning}
We collected a variety of RGB images, to get deep features on them and train a model. However, the data was split into different categories such as only faces, half-body images, and full-body images to apply suitable deep learning models for each category. since, previously there are no trained networks for human body images captured in the wild, we used ResNet-50, and DenseNet for the estimation. For face images, we have used the tained weights of VGG-face to initialize the model and then learned on the wild face images. 
\begin{figure}[H]
\begin{subfigure}[b]{\linewidth}
    \includegraphics[width=9cm, height=2.5cm]{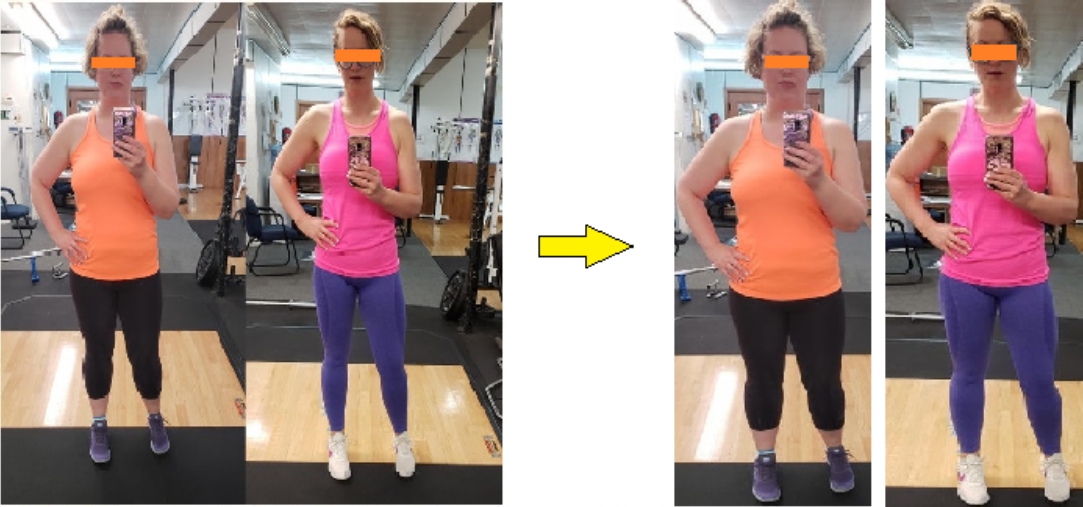}{}
\subcaption{}
  \end{subfigure}
  \newline
\begin{subfigure}[b]{\linewidth}
    \includegraphics[width=9cm, height=2.5cm]{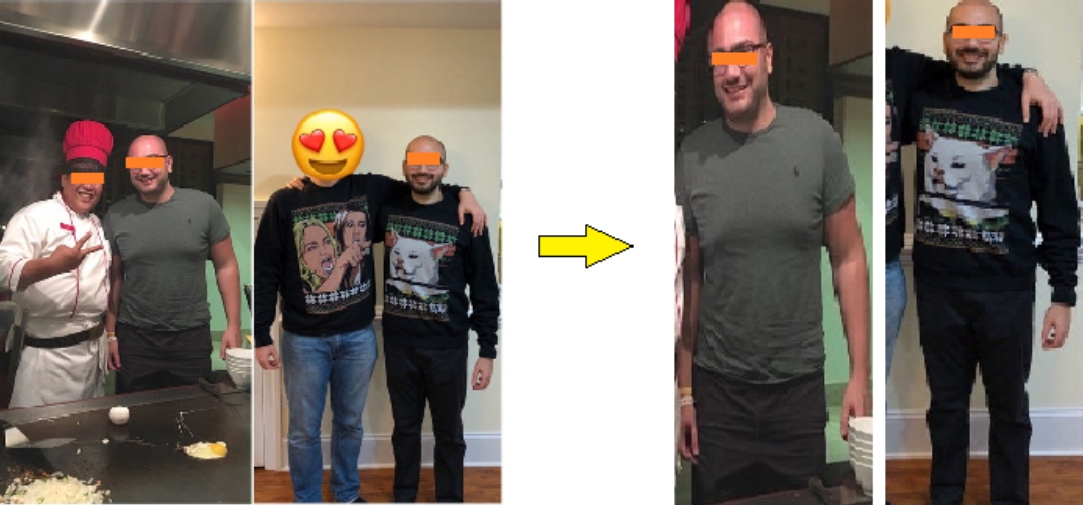}{}
  \subcaption{}
  \end{subfigure}
\newline
\caption{ITU-BMI dataset cleaning: Mask-RCNN \cite{he2017mask} is used to crop the persons however, the cropped persons may still hold other objects with varying backgrounds. (a): a collage of two images of the same person is provided with `before' and `after' weights.  (b): a collage of two persons is shown while corresponding data is for only one person. The required common person is manually selected.}
\label{fig:crop}
\end{figure}

\begin{figure}[H]
\begin{center}
\begin{subfigure}[b]{2cm}
    \includegraphics[width=2cm, height=2cm]{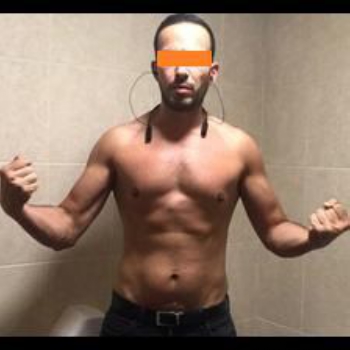}{}
  \end{subfigure}
\begin{subfigure}[b]{2cm}
    \includegraphics[width=2cm, height=2cm]{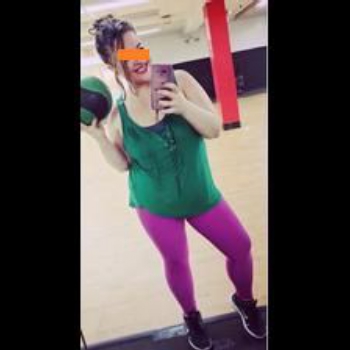}{}
  \end{subfigure}
\begin{subfigure}[b]{2cm}
    \includegraphics[width=2cm, height=2cm]{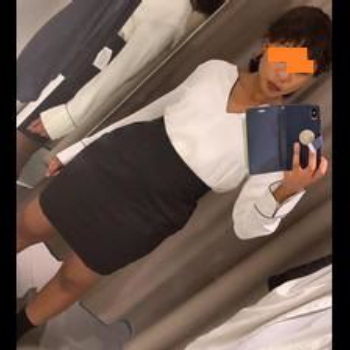}{}
  \end{subfigure}
\begin{subfigure}[b]{2cm}
    \includegraphics[width=2cm, height=2cm]{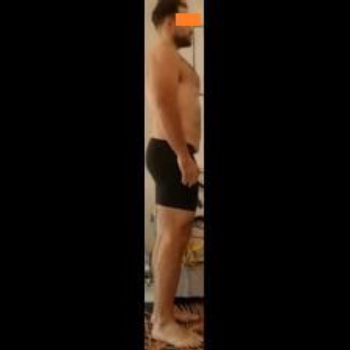}{}
  \end{subfigure}
\newline
 \caption{ Image resizing should not alter the ratio of height to width of a person in the image. Therefore, the larger image dimension is re-scaled to the required size while preserving aspect-ratio. The  smaller image dimension is then padded with zeros to make it compatible with input layer of deep neural networks. }
\label{fig:aspratio}
\end{center}
\end{figure}
\section {Proposed Approach}
We propose the use of deep neural network for the purpose of human body height, weight and BMI estimation using 2D images captured in the wild. We explore different modalities including RGB, gray scale, pose-affinity, depth maps and foreground/background masks. We also explore different architectures including single task and multi-task learning.
\begin{figure*}[ht]
\begin{center}
 \includegraphics[width=1\linewidth]{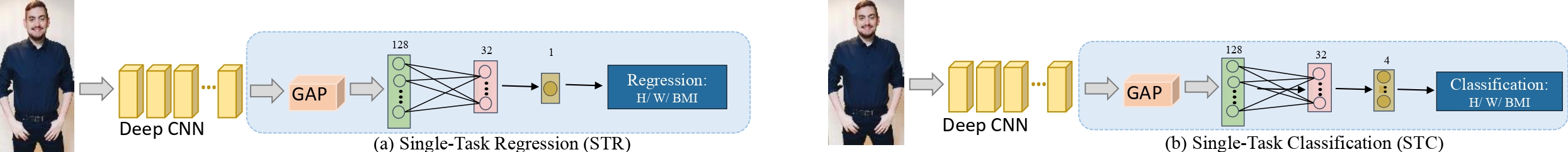}
\caption{Proposed deep neural network architectures for estimation of person weight, height and BMI using single-task learning: (a) Single Task Regression (STR), (b) Single Task Classification (STC).}
\label{tab:modelsingle}
\end{center}
\end{figure*}

\begin{figure*}[t]
\begin{center}
 \includegraphics[width=1\linewidth]{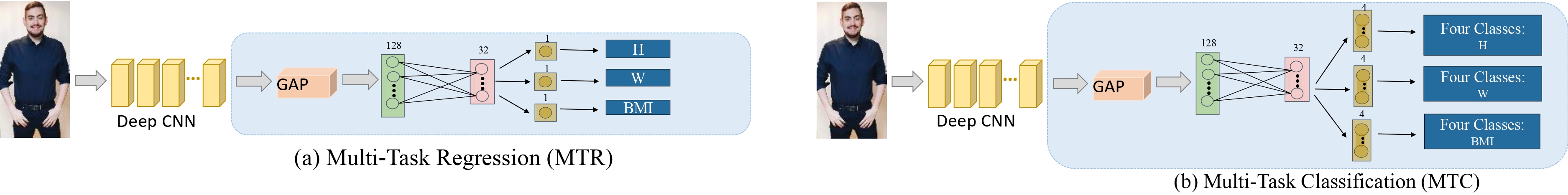}
\caption{Proposed deep neural network architectures for estimation of person weight, height and BMI using multi-task learning: (a) Multi-Task Regression (MTR), (b) Multi-Task Classification (MTC).}
\label{tab:modelmulti}
\end{center}
\end{figure*}

\begin{figure}[H]
  \centering
    \includegraphics[width=1.2cm, height=2cm]{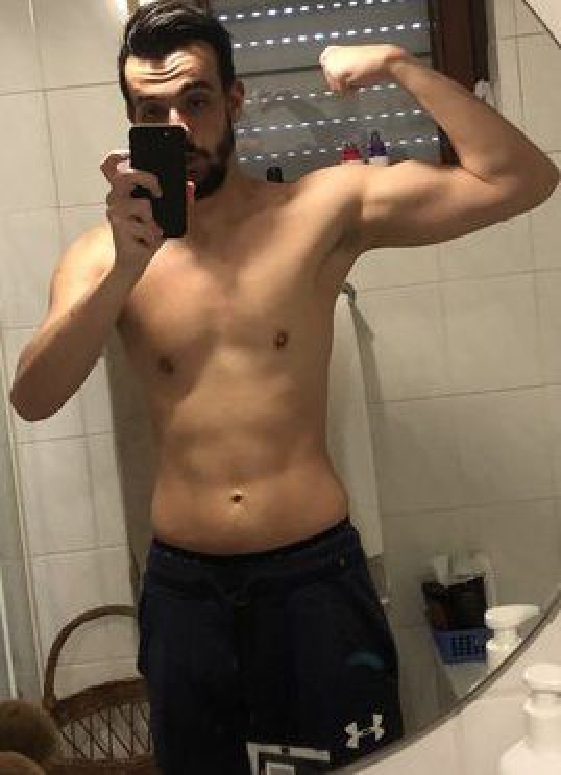}{}
    \includegraphics[width=1.2cm, height=2cm]{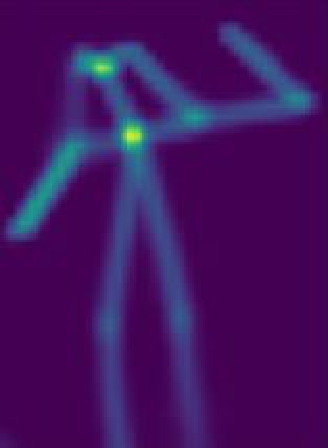}{}
    \includegraphics[width=1.2cm, height=2cm]{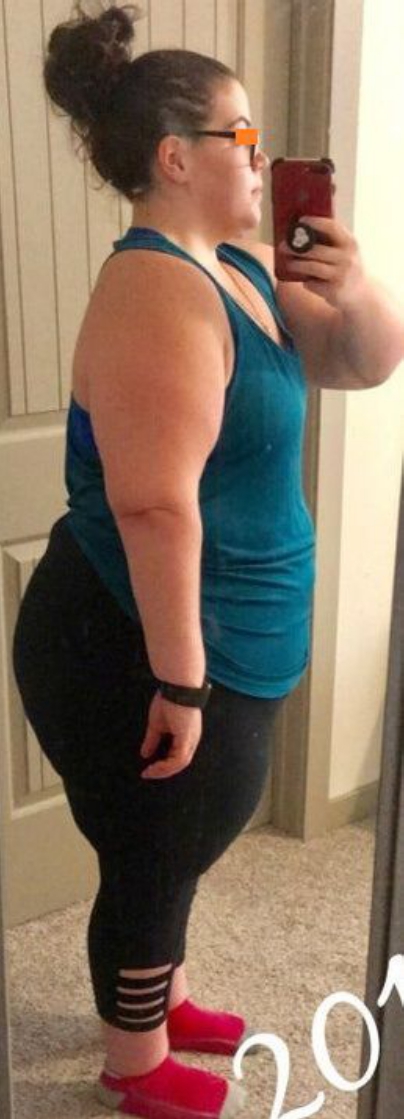}{}
    \includegraphics[width=1.2cm, height=2cm]{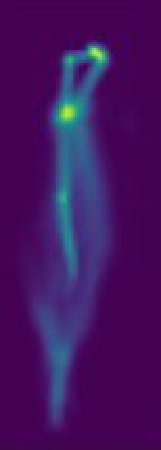}{}
    \includegraphics[width=1.2cm, height=2cm]{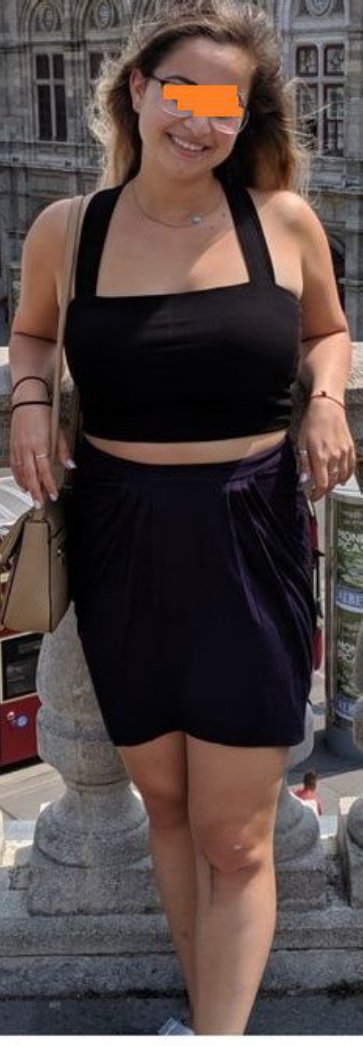}{}
    \includegraphics[width=1.2cm, height=2cm]{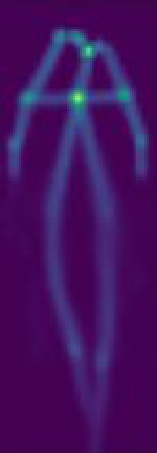}{}
  \\[2mm]
    \includegraphics[width=1.2cm, height=2cm]{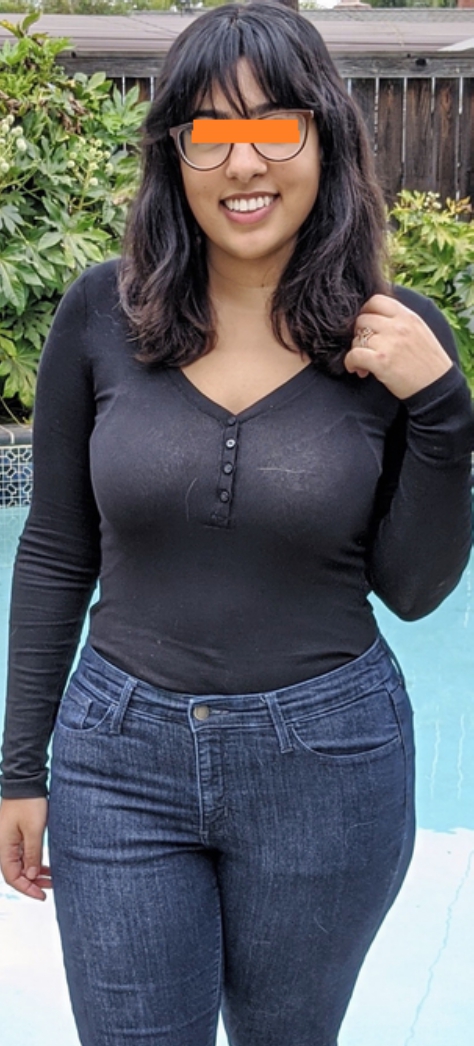}{}
    \includegraphics[width=1.2cm, height=2cm]{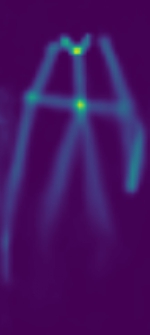}{}
    \includegraphics[width=1.2cm, height=2cm]{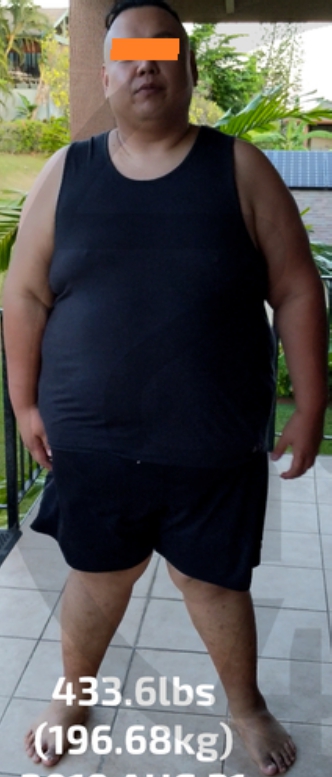}{}
    \includegraphics[width=1.2cm, height=2cm]{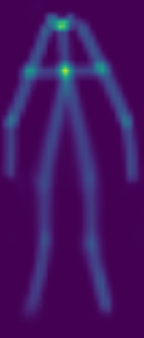}{}
    \includegraphics[width=1.2cm, height=2cm]{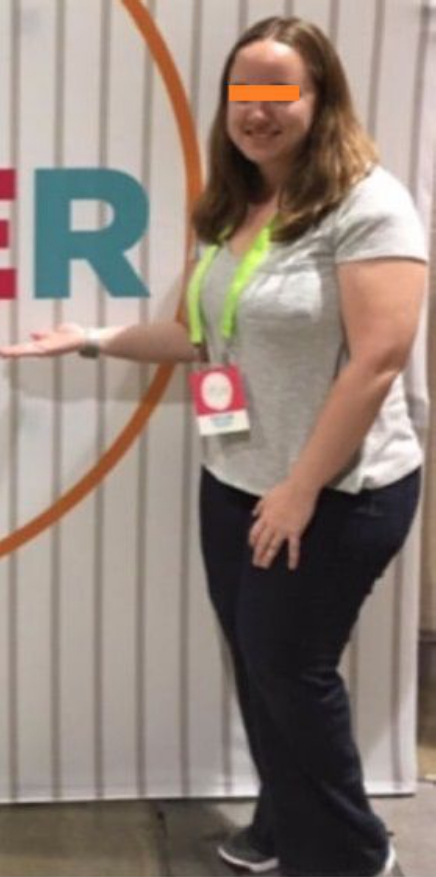}{}
    \includegraphics[width=1.2cm, height=2cm]{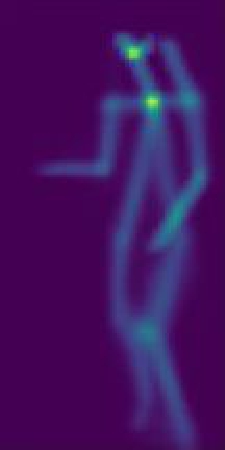}{}
\caption{Sample RGB images from ITU-BMI dataset and corresponding affinity maps computed by Openpose \cite{cao2018openpose}. Pose estimation errors can be seen in some cases.  
  }
  \label{fig:affmaps}
\end{figure}
 
\subsection{Single-task learning}
We performed an end-to-end training of ResNet-50 with pre-trained image-net weights and channel-wise normalization.
ResNet-50 uses residual links and skip connections to deal with vanishing and exploding gradients problems. Our data set includes large diversity, therefore we  choose to fine tune all layer's weights using low learning rate.  The target variables are normalized with their corresponding mean and variance because the units and ranges of height and weight are quite different. As shown in Figure 5a,  in the single task learning with regression network (STR), we used mean squared error loss between ground-truth values $v_i^t$ and predicted estimations $\hat{v}_i^t$ both normalized to unit variance and zero mean: $v_i^t=(\bar{v}_i^t-\mu_t)/\sigma_t$, where $\bar{v}_i^t$ is the observed value, $\mu_t$ and $\sigma_t$ are the mean and variance computed over the training data, and $t \in \{ \text{height}, \text{weight}, \text{BMI}\}$.    
\begin{equation}
   {r}_t=\frac{1}{m}\sum_{i=1}^{m}||v_i^t-\hat{v}_i^t||_2^2,
    \label{equ:regress}
\end{equation}
where  $m$ are the number of images in a batch.
For single task classification network (STC), as shown in Figure 5.b, we used softmax activation with categorical cross entropy loss function:
\begin{equation}
   f_t= -\frac{1}{m}\sum_{i=1}^{m} \sum_{c=1}^{C}\ell_{i,c}^t \log(\hat{\ell}_{i,c}^t),
    \label{equ:class}
\end{equation}
where $\ell_{i,c}$ and $\hat{\ell}_{i,c}$ are the ground truth and predicted labels for the $i$-th instance of $c$-th class, $t \in \{ \text{height}, \text{weight}, \text{BMI}\}$, and $C$ are the total number of classes in our dataset.  
 
\subsection{Multi-task learning}
The weight, height and BMI of a person are highly correlated to each other, which gives the intuition of learning these parameters simultaneously. A multi-task learning arrangement will facilitate common information sharing across these parameters resulting in improved network performance.
Multitask learning shares representations between related tasks by optimizing more than one loss functions at the same time. Since different tasks have different noise patterns, a multitask learning model effectively optimizes all tasks simultaneously and learns a more general representation. Multitask learning improves generalization as it applies the knowledge acquired by learning other correlated tasks. 

In the current work, we propose a hard parameter sharing configuration of multitask learning, which  reduces the risk of over-fitting. In such configuration, we share the initial hidden layers between all tasks, while keep several later layers as task-specific. The shared layers learn the common information across all tasks while the specific layers learn the individual behaviour of each task.

We have  implemented different  multitask learning configurations including multiple regressions and classifications as shown in Figure 6. In case of multi-task regression (MTR), as shown in Figure 6.a, different regression losses are combined:
\begin{equation}
\mathcal{L}_r= \lambda_{1}r_h+\lambda_{2}r_w +\lambda_{3}r_{bmi},
\label{equ:losseq}
\end{equation}
where $r_h$, $r_w$, and $r_{bmi}$ are the regression losses of height, weight and BMI respectively as defined by Eq. \eqref{equ:regress}, and $\lambda_1$, $\lambda_2,$ and $\lambda_3,$ are hyper-parameters.  
For multi-task classification (MTC) as shown in Figure 6.b, the classification losses are combined together.
\begin{equation}
\mathcal{L}_c= \lambda_{4}f_h+\lambda_{5}f_w +\lambda_{6}f_{bmi},
\label{losseq1}
\end{equation}
where $f_h$, $f_w$, and $f_{bmi}$ are the classification losses of height, weight and BMI respectively as defined by Eq. \eqref{equ:class}, and $\lambda_4$, $\lambda_5$ and $\lambda_6,$ are hyper-parameters.
 
\subsection{Learning Using Different Modalities}
In addition to the RGB images, we extend our approach to include human body and depth information to facilitate the network in estimating body weight, height and BMI. Body pose estimates the location of body joints which help to better capture the relationship between different body parts of the same person.

\subsubsection{Human Body Parts Affinity Maps} Part affinity maps learn association between different human body parts in a given image. To compute these maps we used 'open pose' proposed by Cao et al.~\cite{cao2018openpose}. 
Open-pose is a multi-stage CNN that produces two different outputs, including  confidence maps of different body part locations such as left and right ears, wrists, knees, ankles, etc. The second output is affinity maps showing a degree of association between these body parts.  
Openpose detects 18 different body joints and by joining nearby joints 19 limbs are obtained. In the current work, we employed  convolutional neural networks trained on COCO dataset \cite{lin2014microsoft}  to obtain parts affinity fields  in the proposed dataset, as shown in Fig. \ref{fig:affmaps}. 
The obtained affinity maps are not always accurate, especially on the side pose images and persons holding objects as well as partial occlusions. Therefore, affinity maps alone are not enough for accurate weight, height and BMI estimation.

\subsubsection{Human Body Depth Maps:}  Depth maps provide information about the body shape of a person.  Since there is no human body depth estimation network present to date that is trained on in the wild human body images, so we used NYU-v2 depth pre-trained network \cite{alhashim2018high} for our experimentation. 
 
To compute the depth maps we employ encoder-decoder network by AlHashim et al., \cite{alhashim2018high}. In their architecture, DenseNet-169 is used as an encoder, and a series of up-sampling layers are used as a decoder network.   The input to the network is RGB image and output is the final depth map at half the input image resolution. The loss function for depth estimation is the summation of $\ell_1$ error between ground-truth and the network prediction, and also between gradients of the computed and predicted depth maps. In addition, the loss of structural similarity between these depth maps is also included in the objective function.
\begin{equation}
    \mathcal{L}_d =f_{d} +\beta_1 f_{g} +\beta_2 f_{s},
    \label{equ:depth}
\end{equation}
where $\beta_1$ and $\beta_2$ are hyper parameters, $f_{d}$, $f_{g}$, and $f_{s}$ are different types of losses defined as: 
\begin{equation}
    f_{d}=\frac{1}{n}\sum_{i=1}^{n}|d_i-\hat{d}_i|,
    \label{equ:ldepth}
\end{equation}
\begin{equation}
    f_{g}=\frac{1}{n}\sum_{i=1}^{n}|g_x(d_i)-g_x(\hat{d}_i)|+|g_y(d_i)-g_y(\hat{d}_i)|,
    \label{equ:gdepth}
\end{equation}
\begin{equation}
    f_{s}=\frac{1-SSIM(d,\hat{d})}{2},
    \label{equ:sdepth}
\end{equation}
where $g_x(\cdot)$, $g_y(\cdot)$ are the depth image gradients along X-axis and Y-axis, $SSIM(\cdot)$ is the structural  similarity \cite{wang2004image} used for image reconstruction tasks. The network is trained on the NYU v2 dataset comprising of various indoor scenes' videos recorded by both  RGB and Depth cameras at a resolution of $640 \times 480$. The dataset contains 120K training samples, and the network is trained on a 50K subset. This dataset also contains some human body images. Obtained depth maps on our proposed ITU-BMI dataset are  shown in Fig. \ref{fig:depth}. Because NYU dataset contains fewer human images,  the obtained depth map results are not very accurate. Fig. \ref{fig:depth} shows that body parts closer to the camera appear darker and the distant body parts appear lighter in colour.

\begin{figure}[H]

  \centering
    \includegraphics[width=1.2cm, height=2cm]{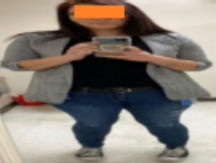}{}
    \includegraphics[width=1.2cm, height=2cm]{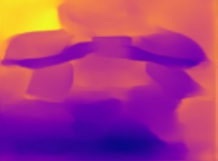}{}
    \includegraphics[width=1.2cm, height=2cm]{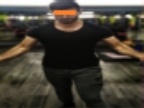}{}
    \includegraphics[width=1.2cm, height=2cm]{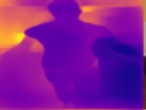}{}
    \includegraphics[width=1.2cm, height=2cm]{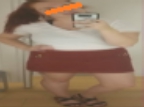}{}
    \includegraphics[width=1.2cm, height=2cm]{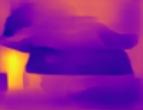}{}
    \\[2mm]
    \includegraphics[width=1.2cm, height=2cm]{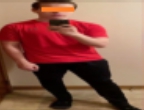}{}
    \includegraphics[width=1.2cm, height=2cm]{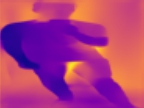}{}
    \includegraphics[width=1.2cm, height=2cm]{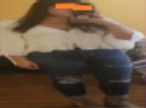}{}
    \includegraphics[width=1.2cm, height=2cm]{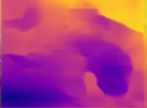}{}
    \includegraphics[width=1.2cm, height=2cm]{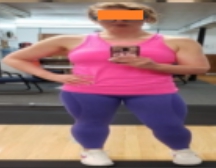}{}
    \includegraphics[width=1.2cm, height=2cm]{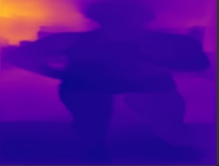}{}
  \caption{Sample RGB images and the corresponding depth maps computed from NYU depth pre-trained network \cite{alhashim2018high}. Despite some errors, body parts closer to the camera appear dark and body parts distant from camera appear light showing depth variation across the human body.}
  \label{fig:depth}
\end{figure}

\subsubsection{Human Body Foreground Masks}
To damper the effect of background noise and make our network focused on the human body, we have employed human foreground-background segmentation. For this purpose, we used Mask-RCNN \cite{he2017mask} which is an instance segmentation deep neural network that employs a region proposal network (RPN), which uses anchors to generate object proposals. Anchors are  sets of bounding boxes with predefined locations and scales relative to an input image. All ground-truth classes and bounding boxes are assigned to individual anchors according to a criterion based on  Intersection over Union (IoU). In the next stage, another neural network takes proposals from RPN and locates relevant areas on feature map using RoIAlign. This stage outputs object classes, bounding boxes, and a network branch also generates masks for each object on  pixel level. Few output masks for our proposed dataset by MaskRCNN are shown in Fig. \ref{fig:masks}. Our dataset is quite diverse and it contains low-resolution images, therefore the produced masks are not very accurate in some cases. 

\begin{figure}[H]
\begin{center}
    \includegraphics[width=1.2cm, height=2cm]{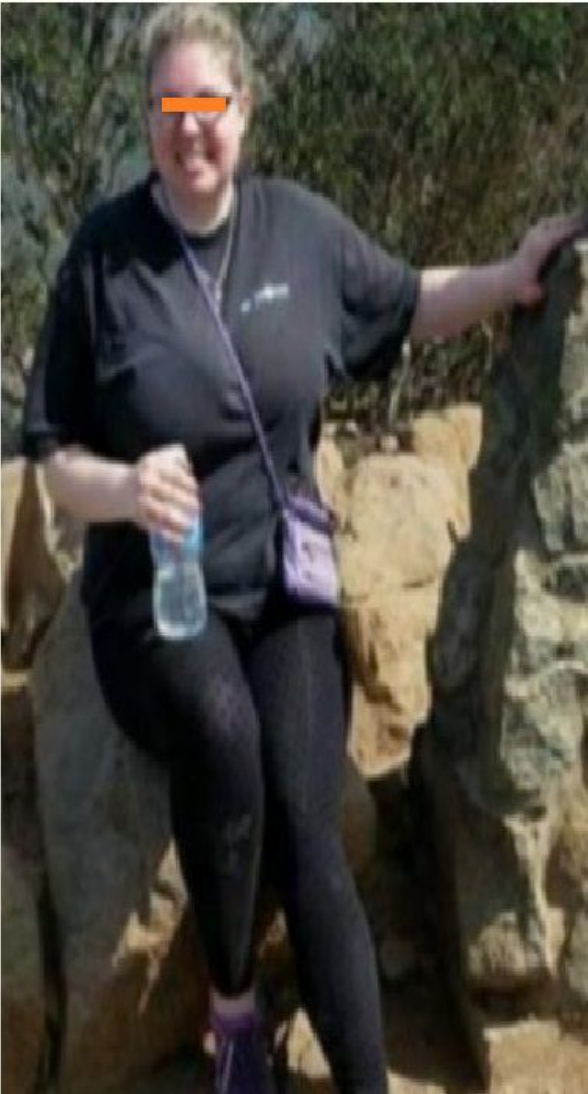}{}
    \includegraphics[width=1.2cm, height=2cm]{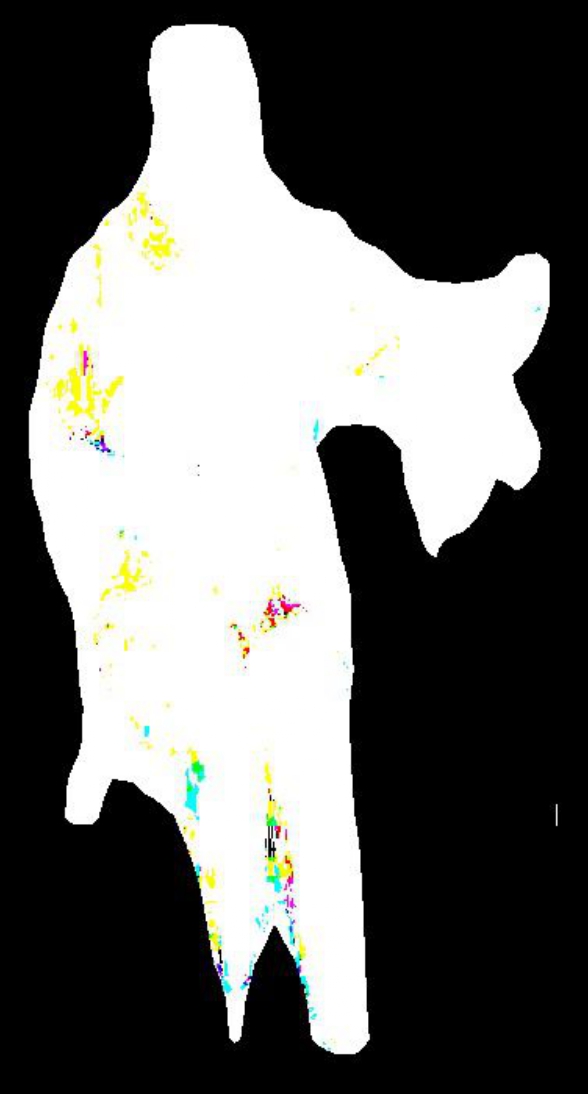}{}
    \includegraphics[width=1.2cm, height=2cm]{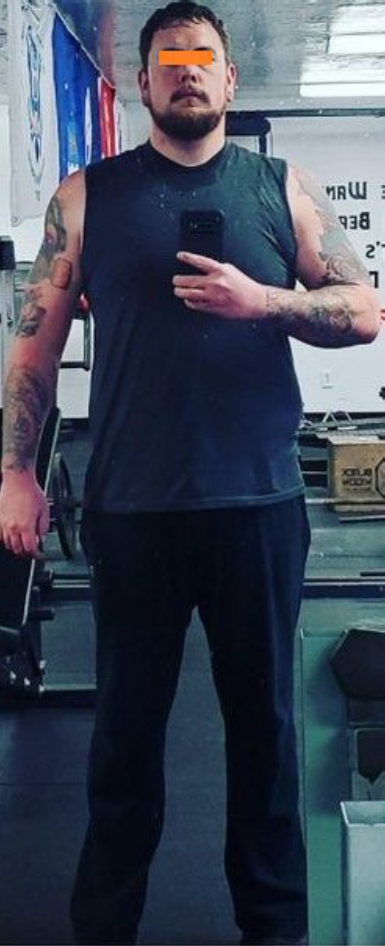}{}
    \includegraphics[width=1.2cm, height=2cm]{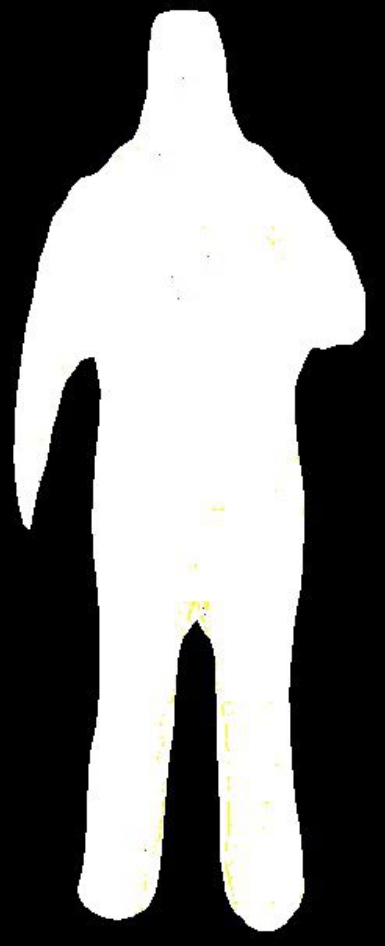}{}
   \includegraphics[width=1.2cm, height=2cm]{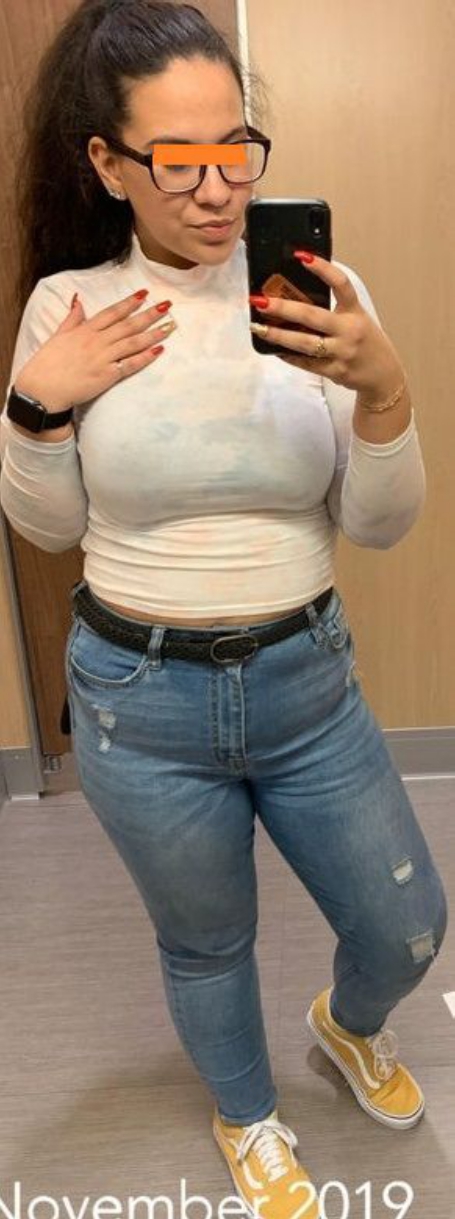}{}
    \includegraphics[width=1.2cm, height=2cm]{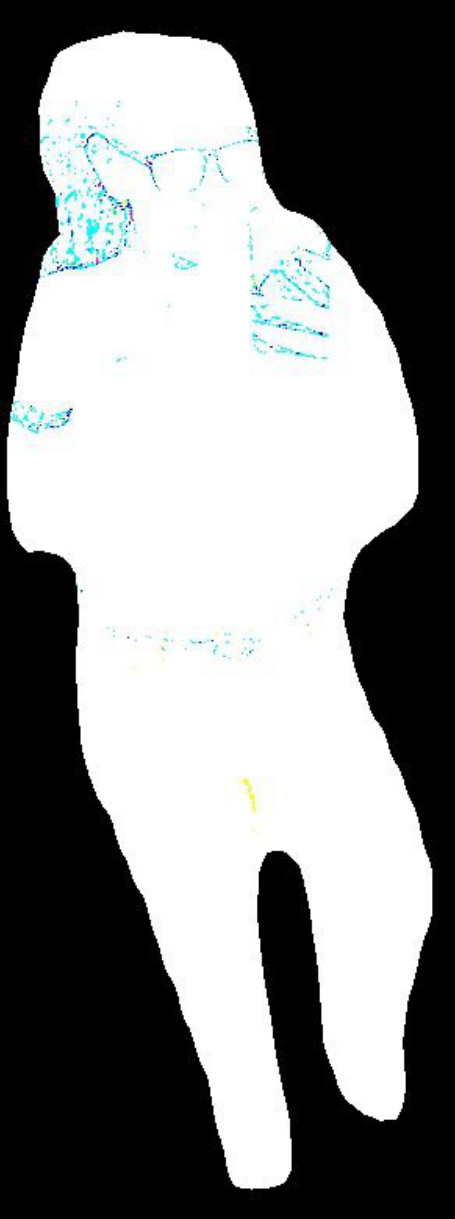}{}
 \\[2mm]
    \includegraphics[width=1.2cm, height=2cm]{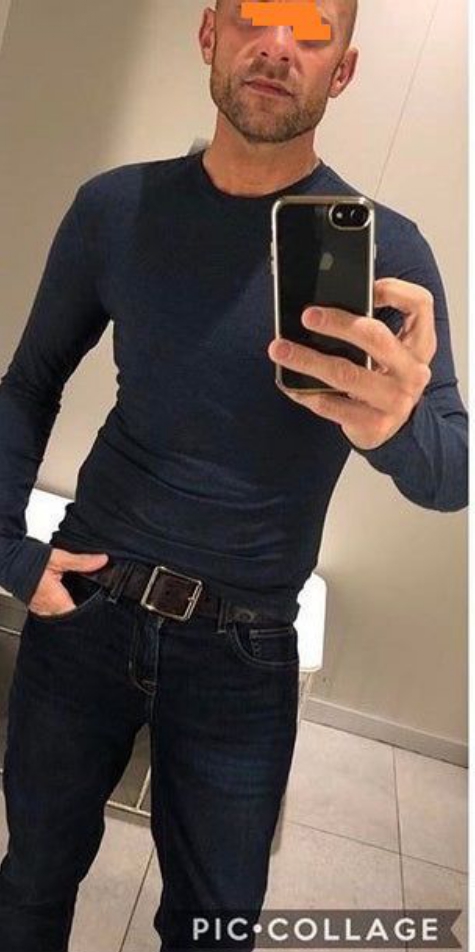}{}
    \includegraphics[width=1.2cm, height=2cm]{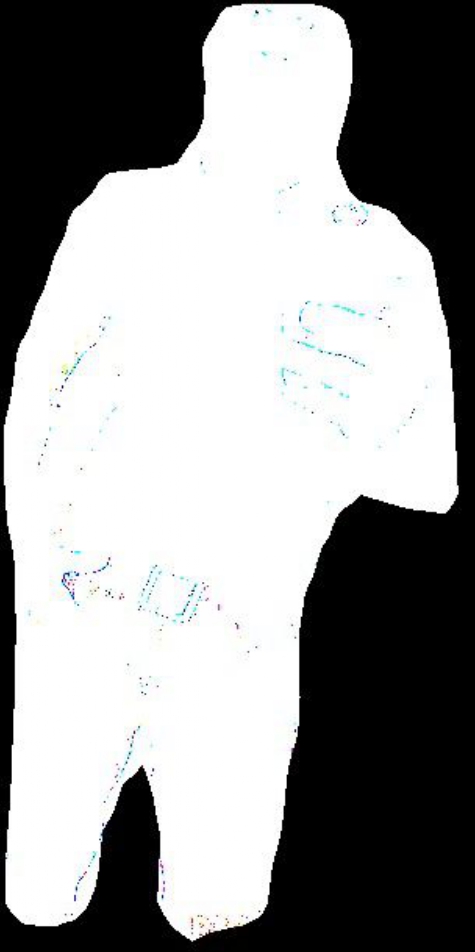}{}
    \includegraphics[width=1.2cm, height=2cm]{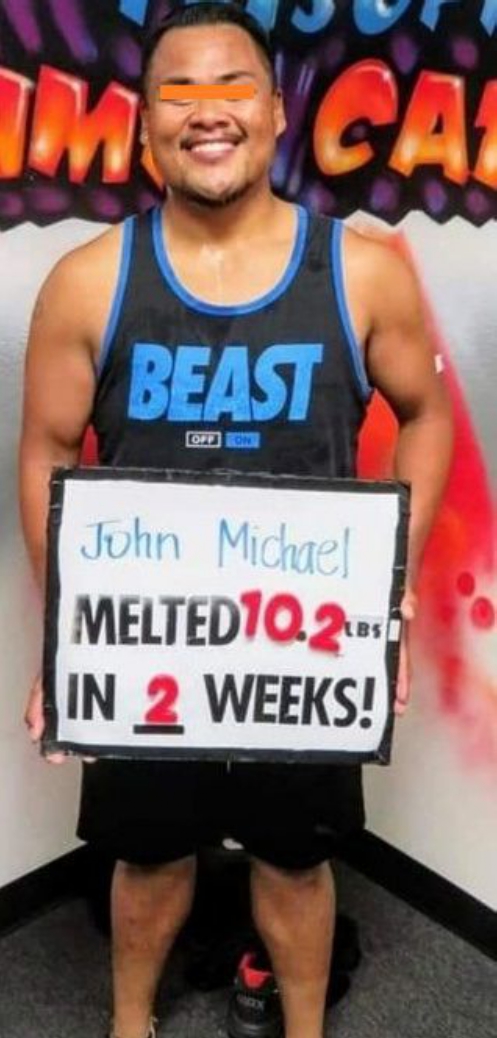}{}
    \includegraphics[width=1.2cm, height=2cm]{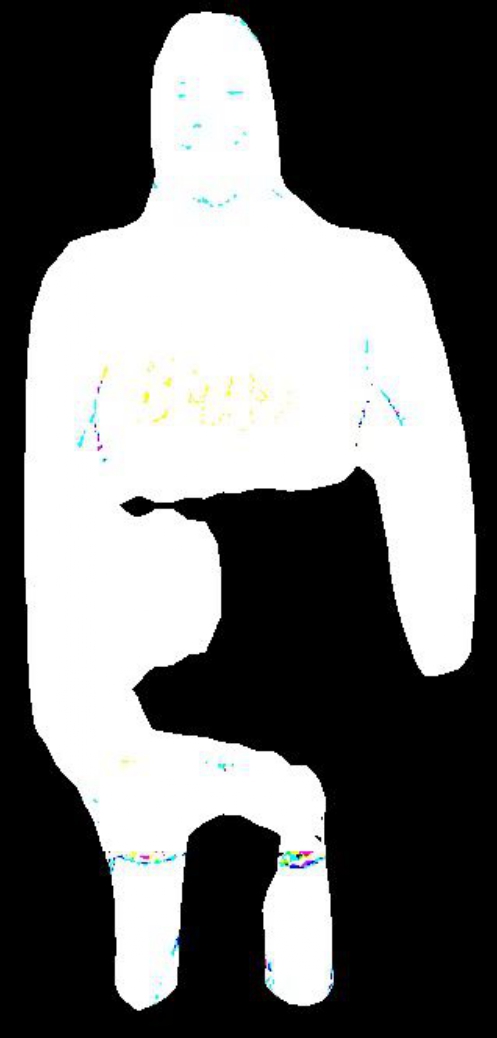}{}
    \includegraphics[width=1.2cm, height=2cm]{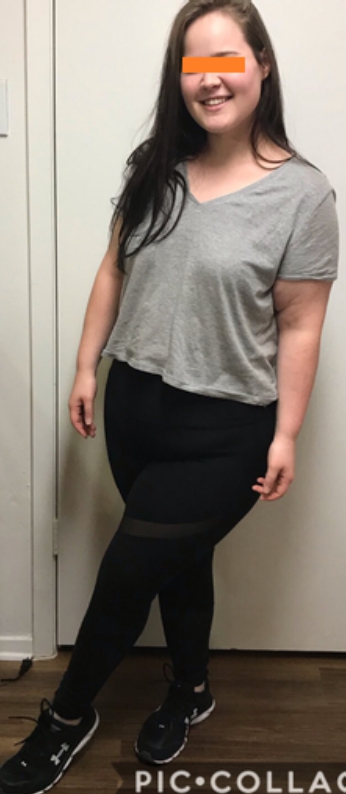}{}
    \includegraphics[width=1.2cm, height=2cm]{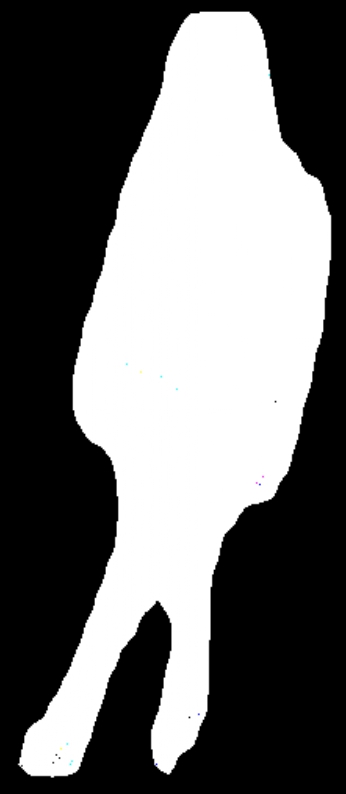}{}
\caption{Sample RGB images with corresponding segmented person maps using He et al. \cite{he2017mask} on ITU-BMI dataset. In some cases, mask estimation errors can be seen.  
}
\label{fig:masks}
\end{center}
\end{figure}
\subsubsection{Combining different Modalities: GAD and GAM}
We have performed experiments with two different combinations of the above discussed modalities. In the first combination, we consider fusion of Gray-scale images with Affinity maps and Depth maps which we dub as `GAD'. In the second combination, we fuse Gray-scale, Affinity maps and body Masks dubbed as `GAM'.  In GAD and GAM, different modalities are concatenated as three channels and the resulting data-structure is input to the proposed network. 

\subsection{Transformers for BMI Classification and Regression}
\subsection{Transformers for MTC and MTR of BMI, Weight and Height}
\section{Experiments and Results}
\subsection{Experimental Setup}
All the images in our proposed  ITU-BMI dataset  are resized to $224\times 224$ while maintaining the aspect ratio to ensure compatibility with the input layers of the feature extractor networks which are VGG16, ResNet50, and Denset121.  For the purpose of resize, the larger image dimension is first re-scaled to 224 and the smaller dimension is then zero-padded to obtain the required size (Fig. \ref{fig:aspratio}). Our dataset contains 6105 images which are randomly divided into 80/20 train/test split. The  ITU-BMI dataset contains  537 full body images (taken from test set only) referred to as ITU-BMI\_B dataset and 5300 images contain only upper body referred as ITU-BMI\_U dataset. Openpose face detector is used to crop faces in the dataset resulting in 3372 facial images referred as ITU-BMI\_F dataset. More details of these categories are shown in Table \ref{tab:datasetcompare}.

We used ResNet-50 as baseline architecture with our proposed STR, STC, MTR, MTC networks (Fig. 5, 6) to predict height, weight, and BMI of a person from his given image. In each network, we added 2 dense fully connected layers of 128 and 32 neurons with leakyrelu activation function (alpha=0.3) to the ResNet50 baseline model. After that three separate dense layers are used in case of multitask regression/ classification, and one dense layer is used for  single-task regression/classification. For  initialization,  Imagenet pre-trained weights were used, followed by complete end-to-end training. We used stochastic gradient descent (SGD) optimizer with learning rate = 0.001, decay rate= 0.0008, batch size =16, and number of epochs $\geq$ 100. A weight and bias decay of $5e^{-4}$ was also applied to all layers of the model.  In the current work, the hyper-parameters used in the equation 3, and 4 are empirically found. More details will be discussed in Ablation study section.

\subsection{Regression Experiments}
We provide a complete analysis of different learning techniques to predict height, weight, and BMI of a human from 2D body image.  Mean absolute error (MAE) and accuracy (\%Acc.) metrices are being used to evaluate regression and classification results. We have compared the proposed approach with existing state of the art (SOTA) methods on the proposed dataset. \newline
\noindent\textbf{Comparison of Single-task Vs. Multi-task Regression:} Table \ref{tab:reg_analysis} shows the comparison of MTR with STR and current state of the art including Dantcheva et al. \cite{dantcheva2018show} and Jiang et al. \cite{jiang2019body}.  The proposed MTR has consistently shown lower MAE over all compared methods on all four proposed datasets. The STR has produced the second best results which are also significantly better than the existing state of the art methods. For the case of MTR, the BMI regression over face dataset (ITU-BMI\_F) has obtained an error of 4.87 which is significantly higher than the BMI estimated from the full body (ITU-BMI\_B) which is 3.32. ITU-BMI dataset which includes upper body, and fully body has obtained 3.73 MAE which is lesser than the upper body BMI due to more training images and full body images. For the weight estimation, minimum MAE of 12.60 Kg is obtained by MTR on ITU-BMI\_B which is significantly lower than the weight estimation from face images in ITU-BMI\_f, which is 17.11 Kg. For height estimation, minimum MAE of 0.08m is observed on ITU-BMI\_B dataset which is better compared to ITU-BMI\_F and ITU-BMI\_U images. We observe that full human body images provide more information, and produce more accurate estimations of H, W and BMI as compared to the upper-body or face images only. 
 
\begin{table}[H]
    \scriptsize
    \centering
    \caption{Performance comparison of the proposed Single-task Regression (STR), Multi-task Regression (MTR) for BMI, weight (kilograms), and height (meters) estimation, in terms of Means Absolute Error 
    (MAE) on the proposed data sets using RGB images}
    \setlength\tabcolsep{2pt}
    \begin{tabular}{|c| c |c| c| c| c| c|}
        \hline
        
         \multirow{2}{*}{}
         {\bf Datasets}& {\bf Task} & {\bf STR} & {\bf MTR}&{\bf{Dantcheva et al.}}& \multicolumn{2}{c|}{\bf{Jiang et al. \cite{jiang2019body}}}\\
         &&&&\bf{ \cite{dantcheva2018show}}& SVR & GPR\\
         \hline
         \multirow{3}{*}{ITU-BMI\_F} 
        &BMI& 5.21 & \bf {4.87} & 7.13 &-&-\\
        \cline{2-7}
        &W& 17.62 & \bf {17.11} & 22.59 &-&-\\
        \cline{2-7}
        &H& 0.11 & \bf {0.09} & 0.44 &-&-\\
         \hline
       \multirow{3}{*}{ITU-BMI\_U} 
        &BMI& 4.17 & \bf {4.11} & 5.94 &-&-\\
        \cline{2-7}
        &W& 15.11 & \bf {14.89} & 19.58 &-&-\\
        \cline{2-7}
        &H& 0.11 & \bf {0.09} & 0.38 &-&-\\
         \hline
        \multirow{3}{*}{ITU-BMI\_B} 
        &BMI& 4.39 & \bf {3.32} & 5.33 & 12.49 & 12.66\\
        \cline{2-7}
        &W& 14.92 & \bf {12.60} & 17.63 & 23.77 & 24.04\\
        \cline{2-7}
        &H& 0.09 & \bf {0.08} & 0.36 & 0.10 & 0.09\\
         \hline
        \multirow{3}{*}{ ITU-BMI} 
        &BMI& 4.29 & \bf{3.73} & 5.86 &-&-\\
        \cline{2-7}
        &W& 14.82 & \bf{13.69} & 19.11 &-&-\\
        \cline{2-7}
        &H& 0.11 & \bf{0.08} & 0.37 &-&-\\
         \hline
    \end{tabular}
    \label{tab:reg_analysis}
\end{table} 

 Among the existing BMI prediction methods, Jiang et al. \cite{jiang2019body}, similar to our approach estimates BMI using only fully body images. However, the code and the dataset used by them is not publicaly available. In order to report their performance on ITU-BMI\_B dataset, we replicated their method by closely following the approach mentioned in their paper.  The results of Jiang et al. using SVR and GPR were compared with our approaches in Table \ref{tab:reg_analysis}. Both versions of Jiang et al. has produced significantly larger error for the case of BMI and weight prediction. For the case of height prediction where GPR version has obtained the same MAE as the proposed STR approach. Their degraded performance may be attributed to the lack of representation power of anthropometric features which require frontal full body images.  In our dataset, the pose varies widely therefore, the anthropometric features have performed poor. 
 
 The comparison has also been made with the work of Dantcheva et al. \cite{dantcheva2018show} which was originally proposed for BMI, weight and height prediction using only frontal face images.  In our proposed ITU-BMI\_F dataset, facial pose varies significantly which has resulted in the degraded performance of their algorithm. In addition to the face images, their algorithm is also trained and evaluated for the uppper body, fully body and the overall ITU-BMI dataset. On full body, Dantcheva et al. has performed significantly better than Jiang et al. and remian a close competitor to the STR network. However, the performance of the proposed MTR network has remained the best across all the experiments. 
 
\begin{figure}[H]
\begin{center}
    \includegraphics[width=1.0cm, height=2cm]{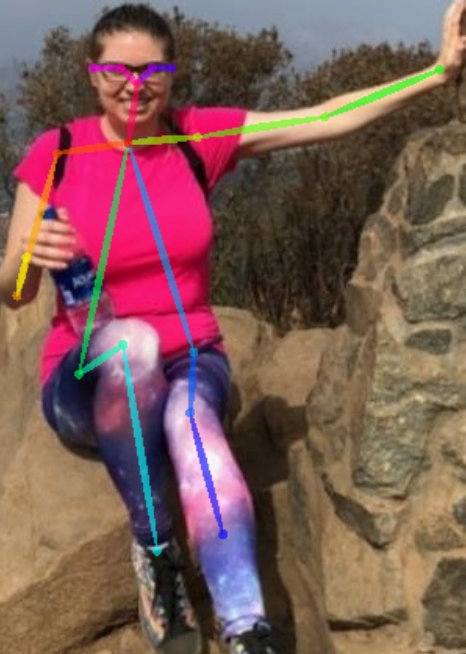}{}
    \includegraphics[width=1.0cm, height=2cm]{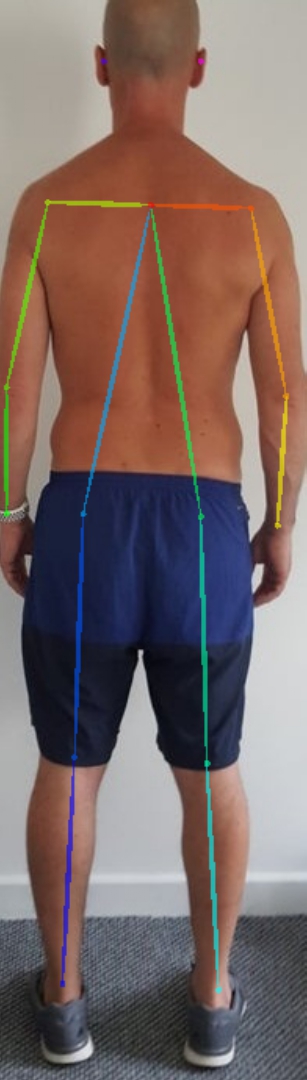}{}
    \includegraphics[width=1.0cm, height=2cm]{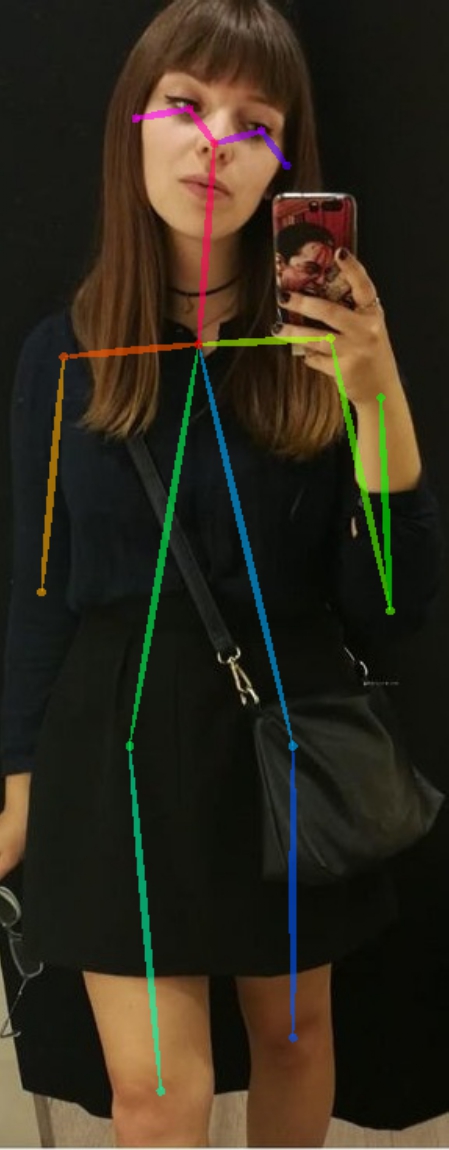}{}
    \includegraphics[width=1.0cm, height=2cm]{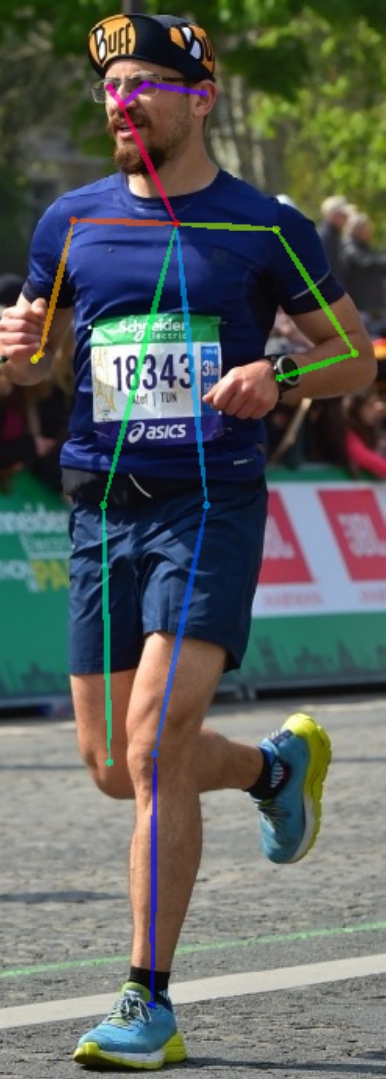}{}
    \includegraphics[width=1.0cm, height=2cm]{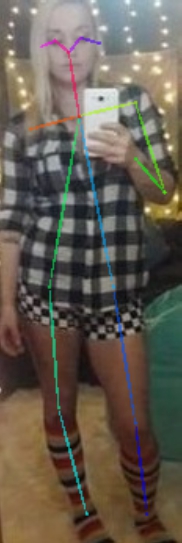}{}
    \includegraphics[width=1.0cm, height=2cm]{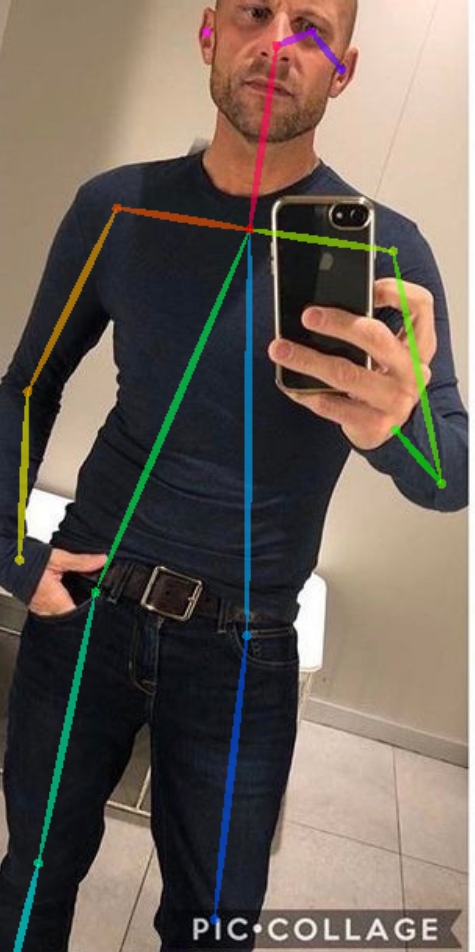}{}
    \includegraphics[width=1.0cm, height=2cm]{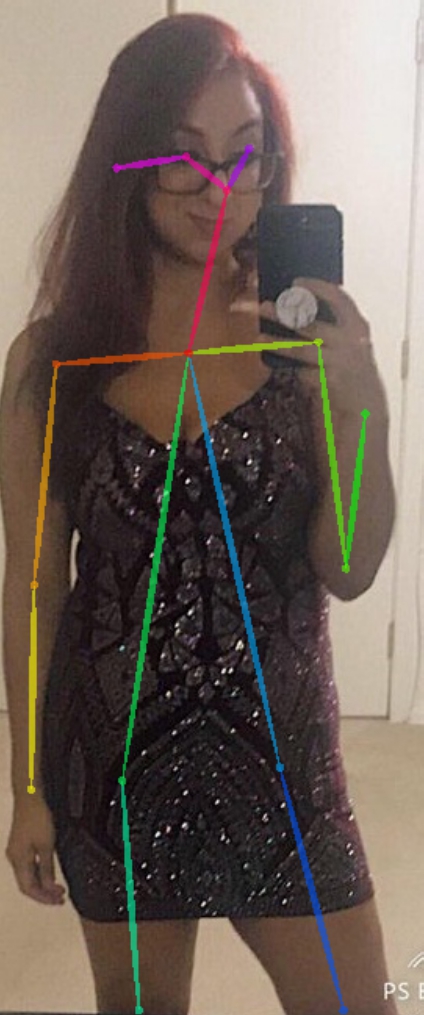}{}
    \includegraphics[width=1.0cm, height=2cm]{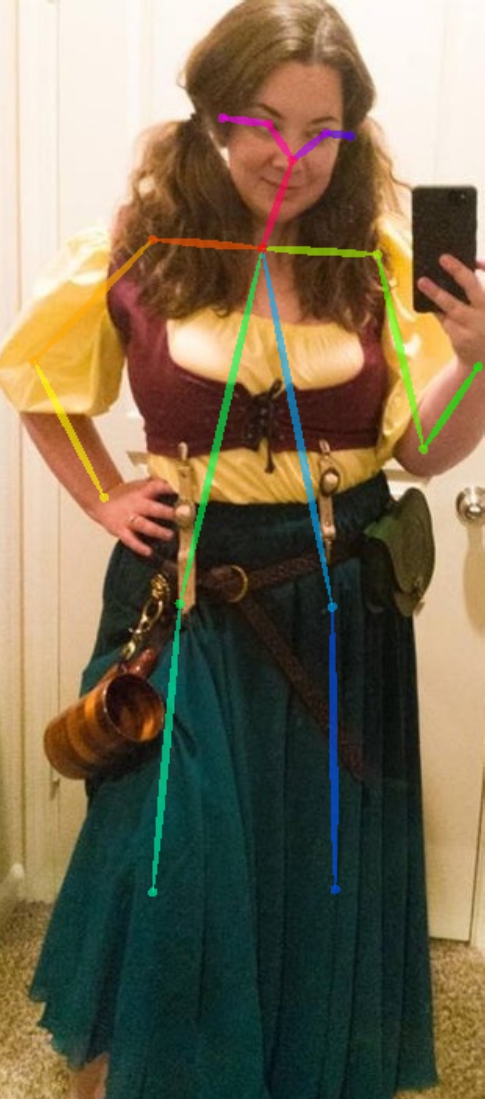}{}
\caption{Sample RGB images with overlaid body joint positions as detected by Openpose \cite{cao2018openpose}. Missing joints can be observed in some cases resulting in height, weight and BMI estimation errors.  
}
\label{fig:joints}
\end{center}
\end{figure}
 
\noindent\textbf{Comparison of GAD and GAM Modalities.}
In addition to the RGB images, the proposed STR and MTR methods have also been evaluated on Gray scale-Affinity-Depth (GAD) and Gray scale-Affinity-Mask (GAM) multi-modality images as discussed in Section V.C.  The experiments are performed on full ITU-BMI dataset using proposed STR and MTR networks and results are shown in Table \ref{tab:gad}. We observe that in all the experiments GAM has resulted better performance than GAD which is mainly due to the reduce quality of depth images. A better estimation method may improve the GAD results. We also observe that MTR is able to obtain performance than STR which may be attributed to the overlap information between parameters to be estimated. The proposed approach is also compared with Dantcheva et al. \cite{dantcheva2018show} method. However, the performance of MTR remains the best in all experiments. These findings are consistent with the observations in the Table \ref{tab:reg_analysis}.

\begin{table}[H]
    \scriptsize
    \centering
    \caption{ MAE results on proposed ITU-BMI data set using GAD\& GAM images}
    \setlength\tabcolsep{1pt}
        \begin{tabular}{|c| c| c| c| c|}
        \hline
        \bf{Modalities} & \bf {Task} & \bf {STR} & \bf {MTR} & \bf {Method \cite{dantcheva2018show}}\\
        \hline
        \multirow{3}{*}{GAD} 
        &BMI & 4.70 & \bf{4.39} & 6.24 \\
        \cline{2-5}
        &W & 17.21 & \bf{15.91} & 20.53 \\
        \cline{2-5}
        &H & 0.11 & \bf{0.10} & 0.48  \\
        \hline
        \multirow{3}{*}{GAM} 
        &BMI & \bf{4.01} & 4.11 & 6.46  \\
        \cline{2-5}
        &W & 15.56 & \bf {15.11} & 21.46 \\
        \cline{2-5}
        &H & 0.11 & \bf {0.09} & 0.51 \\
        \hline
        \end{tabular}
    \label{tab:gad}
\end{table}

\noindent\textbf{Analysis of MTR across different BMI classes:} The MTR architecture is found to be the best performer in a wide range of experiments. Therefore, we further analyze only MTR results across different BMI classes as defined in  Section IV. Table \ref{tab:bmiclasserror} demonstrates that full body images dataset ITU-BMI\_B has resulted in the minimum MAE across all the classes.  The performance comparison of different BMI classes show maximum MAE in obese and underweight classes which is due to the minimum number of training examples in these classes. For the case of normal and overweight classes, the error is reduced due to more training examples in these classes. 

\begin{table}[H]
    \scriptsize
    \centering
    \caption{A comparison of MAE amongst different BMI classes using MTR.}
    \setlength\tabcolsep{1pt}
        \begin{tabular}{|c| c| c| c| c| c|}
        \hline
         {\bf Datasets} & {\bf All} & {\bf Underweight} & {\bf Normal}& {\bf Overweight} & {\bf Obese} \\[1ex]
         \hline
         ITU-BMI & 3.73 & 3.82 & 2.95 & 2.76 & 4.93 \\[1ex]
         \hline
         ITU-BMI\_U & 5.94 & 4.38 & 3.09 & 2.80 & 5.47 \\[1ex]
         \hline
         ITU-BMI\_F & 4.87 & 7.97 & 3.99 & 3.09 & 5.97 \\[1ex]
         \hline
         ITU-BMI\_B & \bf 3.32 & \bf 3.34 & \bf 2.43 & \bf 2.36 & \bf 4.74 \\[1ex]
        \hline
            \end{tabular}
    \label{tab:bmiclasserror}
\end{table}
\noindent\textbf{Analysis of MTR across different backbone networks:}
In addition to the ResNet-50 network, which is used in the all previous experiments, we have also done some experimentation using VGG-16 \cite{simonyan2014very} and DenseNet-121 \cite{iandola2014densenet} as backbone architecture in the proposed MTR approach. The experimental results in the Table \ref{tab:compnet} show that  ResNet-50 has resulted in the better estimation of BMI, H, and W as compared to the VGG-16 and DenseNet-121.
  
\begin{table}[H]
    \scriptsize
    \centering
    \caption{Multitask Regression MAE results using different baseline architectures on ITU-BMI RGB images}
    \setlength\tabcolsep{1pt}
        \begin{tabular}{|c| c| c| c|}
        \hline
         {\bf Network} & {\bf BMI} & {\bf Weight[kg]} & {\bf Height[m]}\\[1ex]
         \hline
         VGG16 & 4.23 & 14.73 & 0.08 \\[1ex]
         \hline
         DenseNet121 & 4.11 & 14.56 & 0.09 \\[1ex]
         \hline
         ResNet50 & \bf 3.73 & \bf 13.69 & \bf 0.08 \\[1ex]
        \hline
            \end{tabular}
    \label{tab:compnet}
\end{table}
 
\subsection{Classification Experiments}
In the previous subsection, the values of BMI, weight and height were estimated using deep regression networks. Based on different BMI values, underweight, normal, overweight and obese categories are defined in the literature as shown in the Table \ref{tab:classHW}.  We have also defined four classes for height and weight as well.


\noindent\textbf{Comparison of Single-task Vs. Multi-task Classification for Different Modalities:}
The images in the proposed ITU-BMI dataset are grouped into different classes based on  height and weight ranges.  For each task, images are divided into four different classes, and the ranges are selected such that  approximately balanced number of samples are obtained in each class as shown in Table \ref{tab:classHW}. For the case of BMI, the four classes are defined as per World Health Organisation (WHO) categorization. We performed single-task and multi-task classification on RGB, GAD, and GAM images and compare the classification accuracy (\% Acc.), and Area Under the Curve (\%AUC.) for different modalities.
\begin{table}[H]
    \scriptsize
    \centering
    \caption{Summary of Height (meters), Weight (kilograms) and BMI classes}
    \setlength\tabcolsep{1pt}
  \begin{tabular}{|c| c| c| c| c|}
         \hline
         {}& {\bf Class 0} & {\bf Class 1} & {\bf Class 2} & {\bf Class 3}\\[1ex]
         \hline
         Height [m]& 1.30-1.64 & 1.65-1.71 & 1.72-1.81 & $\geq$ 1.82\\
         \hline
         \# Samples & 1718 & 1567 & 1555 & 1265 \\  
         \hline
         Weight [kg]& 34.0-64.3 & 64.4-83.7 & 83.8-103.1 & $\geq$ 103.2 \\[1ex]
         \hline
         \# Samples & 1508 & 1624 & 1408 & 1565 \\
         \hline
         BMI & 10-18.5 & 18.6-25.0 & 25.1- 30.0 & $\geq$ 30.1\\[1ex]
         \hline
         \# Samples & 731 & 1478 & 1472 & 2424 \\
         \hline
            \end{tabular}
       \label{tab:classHW}
\end{table}
For BMI classification, we define four classes, where class 0 is underweight, class 1 is  normal, class 2 is overweight and class 3 is obese. All classification results are reported on a randomly selected hold-out 20\% test and 80\% train  dataset.

Figure. 5b. and 6b. show the STC and MTC network architectures, respectively. Table \ref{tab:classh} shows the classification results and Fig. \ref{fig:cmrgb}, \ref{fig:cmgad}, and \ref{fig:cmgam} show the corresponding confusion matrices of height, weight, and BMI based classification using STC and MTC employing different modalities.

From Table \ref{tab:classh}, we can see that multi-task classification (MTC) has performed better than single-task classification (STC) in all experiments. In case of BMI, the best performance is obtained on GAM modaility which is 64.61\% classification accuracy and 80.9\% AUC. by MTC. 
AUC measure tells sensitivity which means how well predictions are ranked, rather than their absolute values. AUC provides an aggregate measure of performance across all possible classification thresholds. Classification-threshold invariance is useful in this optimization problem, because it is not critical to minimize one type of classification error. It means there does not exist much disparity between the cost of false negatives vs. false positives.
The heighest correlation is found between height and BMI which is 0.80, and between weight and BMI is 0.58, which is less and does not suggest a linear machine learning model to be fitted. This observation led us to use deep learning feature extraction to be used for estimations. Using deep features multitask predictions of height, weight, and BMI as dependent variables provide more accuracy compared to single-task predictions.
For the case of Weight and Height, RGB modalities has produced the best results. However, the GAD has remained the second best for the case of weight based classification and GAM has remained second best for the case of height based classification. 
  
\begin{table}[H]
    \scriptsize
    \centering
    \caption{Classification accuracy (\%Acc.) and Area under Curve (\%AUC) of H, W \& BMI on proposed ITU-BMI data set using ResNet50}
    \setlength\tabcolsep{1pt}
  \begin{tabular}{|c| c| c| c| c| c| c| c| }
        \hline
        {\bf Modalities} & &\multicolumn{2}{c|}
         {\bf{BMI}}&\multicolumn{2}{c|}
         {\bf{W}}&\multicolumn{2}{c|}
         {\bf{H}}\\
         \cline{3-8}
         & &{\bf Acc.} & {\bf AUC}
         &{\bf Acc.} & {\bf AUC}
         &{\bf Acc.} & {\bf AUC}\\
         \hline
         \multirow{2}{*}{RGB} 
        &STC & 62.32 & 80.
        6 & 57.82 & 80.9 & 47.82 & 71.5\\
        \cline{2-8}
        & MTC & 62.08 & 80.3 & \bf{58.39} & \bf{81.1} & \bf{50.36} & \bf{73.8}\\
        \hline
         \multirow{2}{*}{GAD} 
        &STC & 60.27 & 78.4 & 55.03 & 77.9 & 46.68 & 72.3\\
        \cline{2-8}
        & MTC& 64.12 & 80.7 & 58.06 & 79.9 & 46.19 & 73.3\\
        \hline
         \multirow{2}{*}{GAM} 
        &STC & 61.26 & 77.5 & 54.95 & 78.5 & 47.50 & 73.1\\
        \cline{2-8}
        & MTC & \bf{64.61} & \bf{80.9} & 57.00 & 78.0 & 47.82 & 73.2 \\
        \hline
    \end{tabular}
       \label{tab:classh}
\end{table}

\begin{figure}[H]
\centering
\begin{subfigure}[b]{0.30\linewidth}
\includegraphics[width=2.8cm, height=3cm]{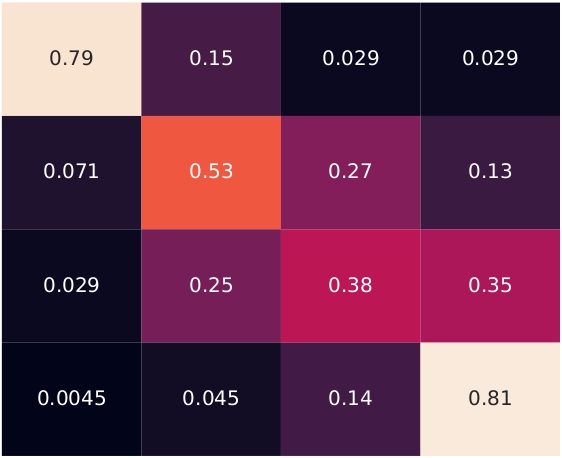}
   \subcaption{} 
\end{subfigure}
~
\begin{subfigure}[b]{0.30\linewidth}
\includegraphics[width=2.8cm, height=3cm]{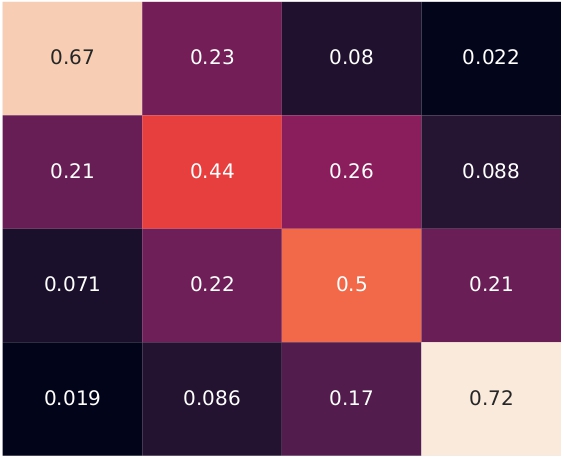}
   \subcaption{} 
\end{subfigure}
~
\begin{subfigure}[b]{0.30\linewidth}
\includegraphics[width=2.8cm, height=3cm]{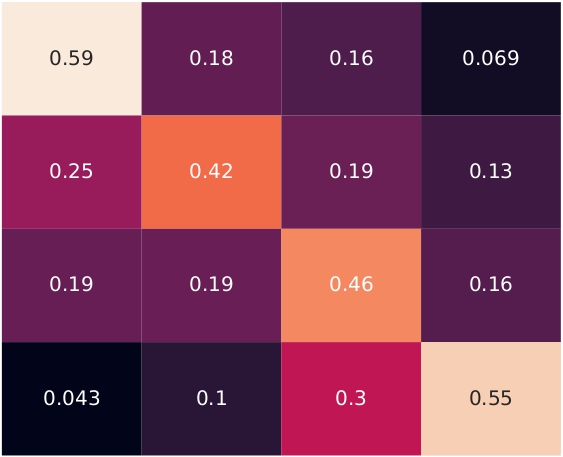}
\subcaption{}
\end{subfigure}
~
\begin{subfigure}[b]{0.30\linewidth}
\includegraphics[width=2.8cm, height=3cm]{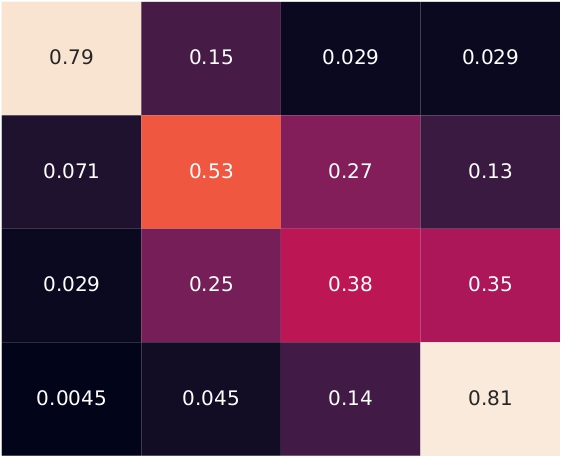}
   \subcaption{} 
\end{subfigure}
~
\begin{subfigure}[b]{0.30\linewidth}
\includegraphics[width=2.8cm, height=3cm]{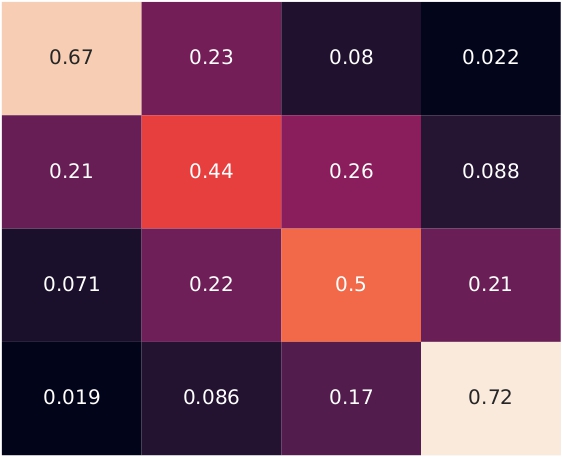}
   \subcaption{} 
\end{subfigure}
~
\begin{subfigure}[b]{0.30\linewidth}
\includegraphics[width=2.8cm, height=3cm]{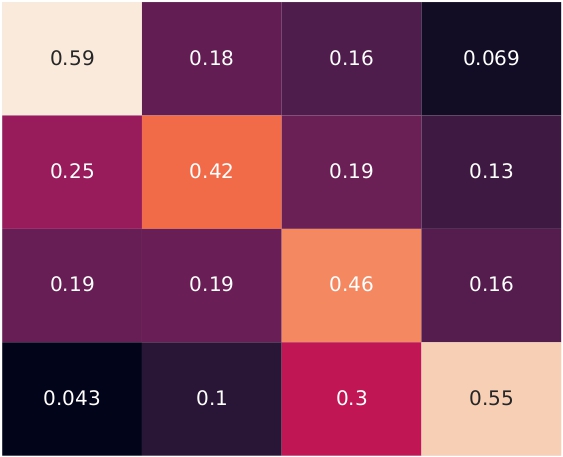}
\subcaption{}
\end{subfigure}
\caption{Confusion matrices of STC and MTC on RGB images. Subfigures (a), (b) and (c)  correspond to BMI, W and H based classification using STC. Subfigures (d), (e) and (f)  correspond to BMI, W and H based classification using MTC.}
\label{fig:cmrgb}
\end{figure}
\begin{figure}[H]
\centering
\begin{subfigure}[b]{0.30\linewidth}
\includegraphics[width=2.8cm, height=3cm]{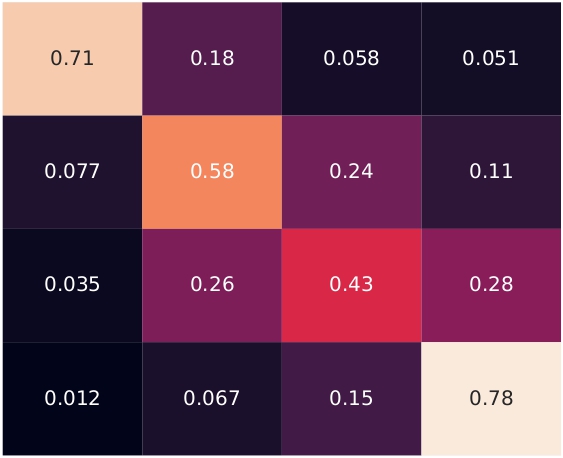}
   \subcaption{ } 
\end{subfigure}
~
\begin{subfigure}[b]{0.30\linewidth}
\includegraphics[width=2.8cm, height=3cm]{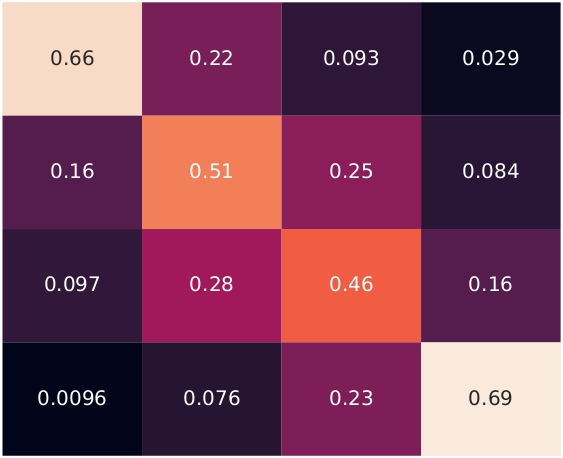}
   \subcaption{ } 
\end{subfigure}
~
\begin{subfigure}[b]{0.30\linewidth}
\includegraphics[width=2.8cm, height=3cm]{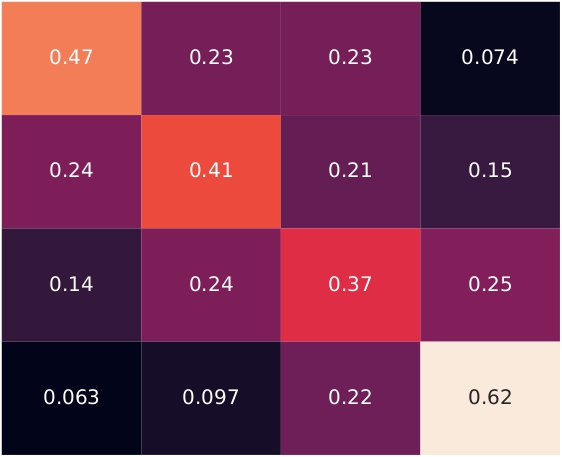}
\subcaption{ }
\end{subfigure}
~  
\begin{subfigure}[b]{0.30\linewidth}
\includegraphics[width=2.8cm, height=3cm]{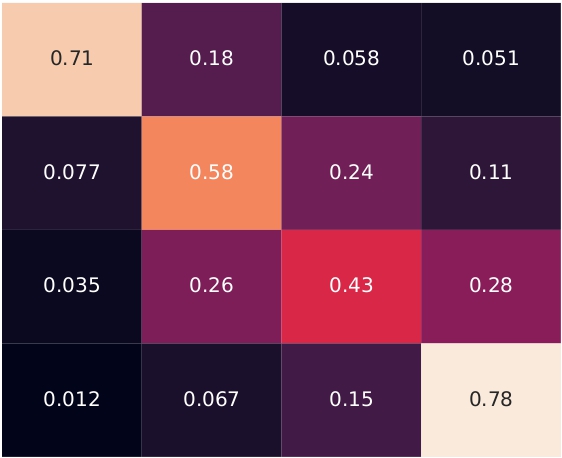}
   \subcaption{ } 
\end{subfigure}
~
\begin{subfigure}[b]{0.30\linewidth}
\includegraphics[width=2.8cm, height=3cm]{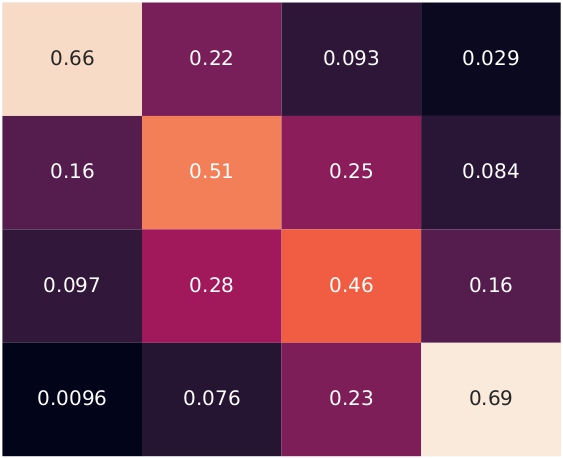}
   \subcaption{ } 
\end{subfigure}
~
\begin{subfigure}[b]{0.30\linewidth}
    \includegraphics[width=2.8cm, height=3cm]{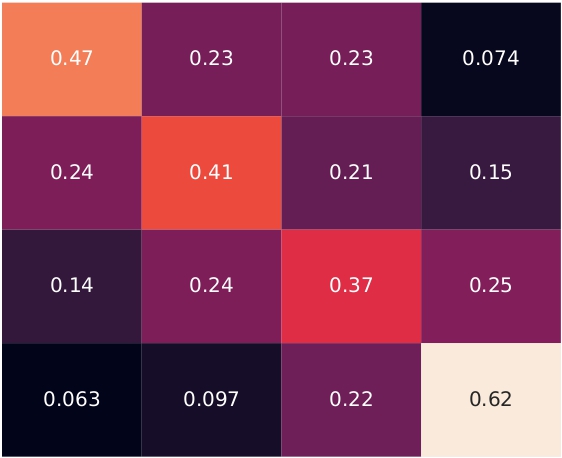}
    \subcaption{ }
    \end{subfigure}
\caption{Confusion matrices of STC and MTC on GAD images. Subfigures (a), (b) and (c)  correspond to BMI, W and H based classification using STC. Subfigures (d), (e) and (f)  correspond to BMI, W and H based classification using MTC.}
\label{fig:cmgad}
\end{figure}
\begin{figure}[H]
\centering
\begin{subfigure}[b]{0.30\linewidth}
\includegraphics[width=2.8cm, height=3cm]{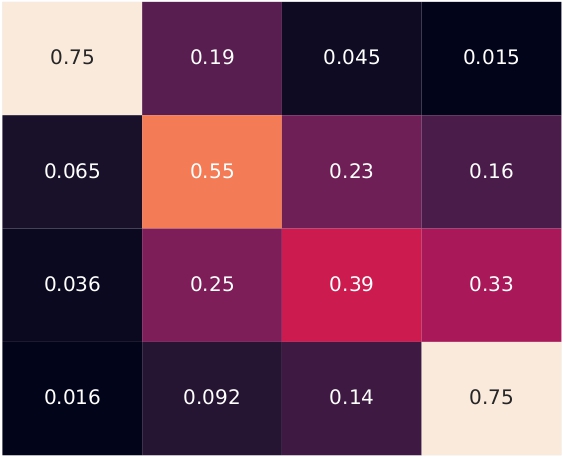}
   \caption{ } 
\end{subfigure}
~
\begin{subfigure}[b]{0.30\linewidth}
\includegraphics[width=2.8cm, height=3cm]{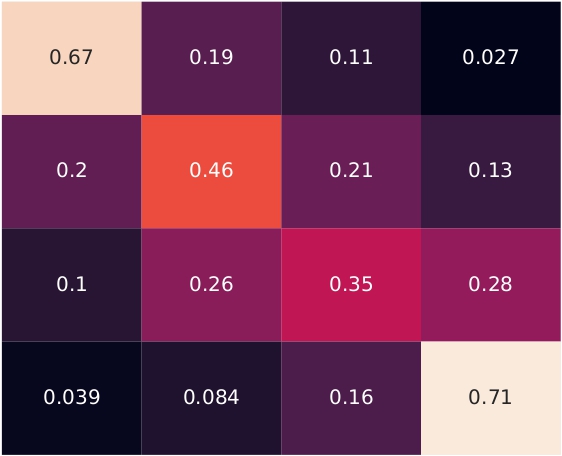}
   \subcaption{ }
\end{subfigure}
~
\begin{subfigure}[b]{0.30\linewidth}
    \includegraphics[width=2.8cm, height=3cm]{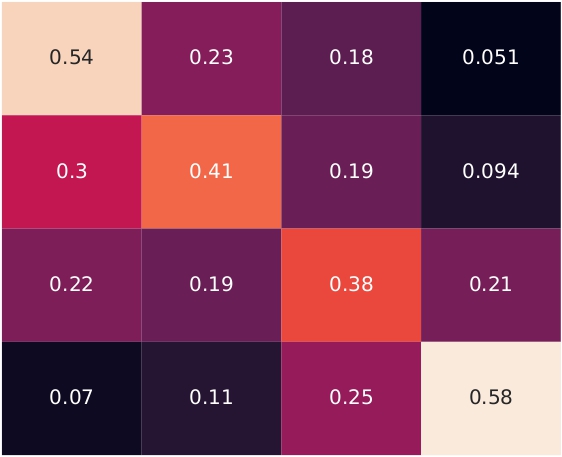}
   \subcaption{ }
    \end{subfigure}
 ~
\begin{subfigure}[b]{0.30\linewidth}
\includegraphics[width=2.8cm, height=3cm]{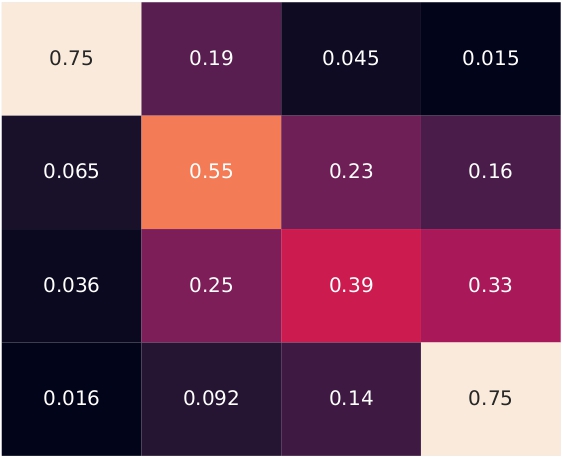}
   \subcaption{ } 
\end{subfigure}
~
\begin{subfigure}[b]{0.30\linewidth}
\includegraphics[width=2.8cm, height=3cm]{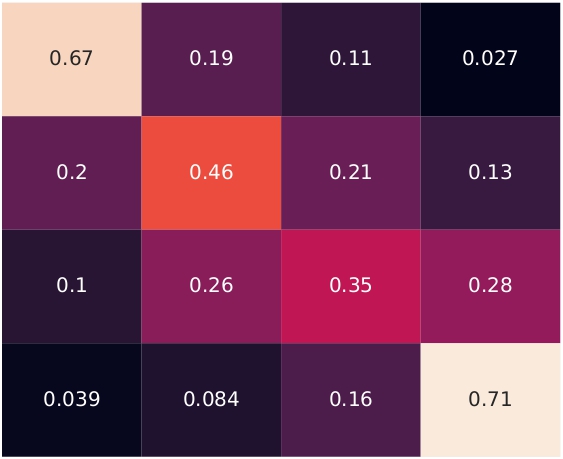}
   \subcaption{ } 
\end{subfigure}
~  
\begin{subfigure}[b]{0.30\linewidth}  
\includegraphics[width=2.8cm, height=3cm]{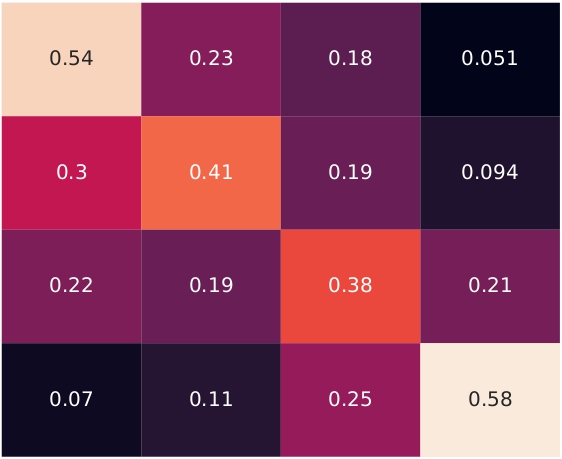}
   \subcaption{ } 
    \end{subfigure}
\caption{Confusion matrices of STC and MTC on GAM images. Subfigures (a), (b) and (c)  correspond to BMI, W and H based classification using STC. Subfigures (d), (e) and (f)  correspond to BMI, W and H based classification using MTC.}
\label{fig:cmgam}
\end{figure}
\noindent\textbf{Analysis of MTC across different backbone networks:}
We have also experimented MTC on RGB images using different deep-nets as baseline architectures to compare the performance as shown in Table \ref{tab:classhnets}. The DenseNet-121 has relatively performed better than other two backbone networks, however, the performance difference is not significant. Motivated by our experiments on MTR, where ResNet-50 was the best performer, most of the experiments reported in this work use ResNet-50 as a backbone.

\begin{table}[H]
    \scriptsize
    \centering
    \caption{Multitask classification Acc. and AUC (\%) of H, W, \& BMI on RGB images using different deep nets.}
    \setlength\tabcolsep{1pt}
  \begin{tabular}{|c| c| c| c|c|}
         \hline
         {\bf Task} & {\bf Networks}& {\bf Acc.}& {\bf AUC}\\
         \hline
        \multirow{3}{*}{BMI} 
        & ResNet-50&62.40&80.5\\
        &DenseNet-121&62.08&{\bf 80.8}\\
        &VGG-16&{\bf 62.81}&80.4\\
         \hline
        \multirow{3}{*}{W} 
        & ResNet-50&57.16&{\bf 80.0}\\
        &DenseNet-121&{\bf 59.54}&79.7\\
        &VGG-16&{\bf 59.54}&77.7\\
        \hline
        \multirow{3}{*}{H} 
        & ResNet-50&48.48&72.2\\
        &DenseNet-121&{\bf 50.20}&{\bf 72.6}\\
        &VGG-16&48.07&68.7\\
        \hline
    \end{tabular}
       \label{tab:classhnets}
\end{table}
\section{Discussion}
We have used mean absolute error (MAE) to estimate regression error. MAE works for same scale of data, and measures the absolute difference between two continuous variables. Our MAE values show the average difference between predicted and target values of H, W, and BMI of ITU-BMI dataset. In general, mean squared error (MSE) is also used to measure the effectiveness of a regression model.

\section{Conclusion}
 In this work, a new dataset and deep learning based framework is proposed for robust estimation of weight, height, and Body Mass Index (BMI) using images captured in uncontrolled environments. For this purpose, deep neural networks are used for regression over weight, height and BMI. In addition to that, based on weight, height and BMI, the dataset is labelled into five categories including under-weight, normal, over-weight, and obese using the WHO recommendations. Deep neural networks are also trained for the purpose of classification using visual images as input. For both regression and classification,  single task as well as multitask approaches are compared and multitask approaches are found to be the more accurate. In addition to RGB mode, other modalities such as depth, edge masks, and affinity maps are also proposed to get improved performance. Extensive experiments are performed using various backbone CNNs including ResNet-50, DesNet-121, and  VGG-16. In most of the experiments, ResNet-50 has yielded the best performance. A new dataset consisting of 6105 images with weight and height labels, extracted features, and the trained networks will soon be made publicly available for research purposes.


\begin{thebibliography}{10}

\bibitem{dantcheva2015else}
Dantcheva, Antitza and Elia, Petros and Ross, Arun.
\newblock What else does your biometric data reveal? A survey on soft biometrics.
\newblock {\em IEEE Transactions on Information Forensics and Security}, volume 11, number 3, pages 441--467. IEEE, 2015.

\bibitem{gonzalez2018facial}
Gonzalez-Sosa, Ester and Fierrez, Julian and Vera-Rodriguez, Ruben and Alonso-Fernandez, Fernando.
\newblock Facial soft biometrics for recognition in the wild: Recent works, annotation, and COTS evaluation.
\newblock {\em IEEE Transactions on Information Forensics and Security}, volume 13, number 8, pages 2001--2014. IEEE, 2018.

\bibitem{neal2019you}
Neal, Tempestt J and Woodard, Damon L.
\newblock You are not acting like yourself: A study on soft biometric classification, person identification, and mobile device use.
\newblock {\em IEEE Transactions on Biometrics, Behavior, and Identity Science}, volume 1, number 2, pages 109--122. IEEE, 2019.

\bibitem{dantcheva2018show}
Dantcheva, Antitza and Bremond, Francois and Bilinski, Piotr.
\newblock Show me your face and I will tell you your height, weight and body mass index.
\newblock {\em 2018 24th International Conference on Pattern Recognition (ICPR)}, pages 3555--3560. IEEE, 2018.

\bibitem{kocabey2017face}
Kocabey, Enes and Camurcu, Mustafa and Ofli, Ferda and Aytar, Yusuf and Marin, Javier and Torralba, Antonio and Weber, Ingmar.
\newblock Face-to-bmi: Using computer vision to infer body mass index on social media.
\newblock {\em Eleventh International AAAI Conference on Web and Social Media}, 2017.

\bibitem{jiang2019body}
Jiang, Min and Guo, Guodong.
\newblock Body Weight Analysis from Human Body Images.
\newblock {\em IEEE Transactions on Information Forensics and Security}. IEEE, 2019.

\bibitem{bell2018detecting}
Bell, Dane and Laparra, Egoitz and Kousik, Aditya and Ishihara, Terron and Surdeanu, Mihai and Kobourov, Stephen.
\newblock Detecting Diabetes Risk from Social Media Activity.
\newblock {\em Proceedings of the Ninth International Workshop on Health Text Mining and Information Analysis}, pages 1--11, 2018.

\bibitem{arnold2016obesity}
Arnold, Melina and Leitzmann, Michael and Freisling, Heinz and Bray, Freddie and Romieu, Isabelle and Renehan, Andrew and Soerjomataram, Isabelle.
\newblock Obesity and cancer: an update of the global impact.
\newblock {\em Cancer epidemiology}, volume 41, pages 8--15. Elsevier, 2016.

\bibitem{lewis2019obesity}
Lewis, Matthew T and Lujan, Heidi L and Tonson, Anne and Wiseman, Robert W and DiCarlo, Stephen E.
\newblock Obesity and inactivity, not hyperglycemia, cause exercise intolerance in individuals with type 2 diabetes: Solving the obesity and inactivity versus hyperglycemia causality dilemma.
\newblock {\em Medical hypotheses}, volume 123, pages 110--114. Elsevier, 2019.

\bibitem{ling2019epigenetics}
Ling, Charlotte and R{\"o}nn, Tina.
\newblock Epigenetics in human obesity and type 2 diabetes.
\newblock {\em Cell metabolism}. Elsevier, 2019.

\bibitem{narayanan2019asthma}
Narayanan, Ajay and Yogesh, Ahana and Mitchell, Ron B and Johnson, Romaine F.
\newblock Asthma and obesity as predictors of severe obstructive sleep apnea in an adolescent pediatric population.
\newblock {\em The Laryngoscope}. Wiley Online Library, 2019.

\bibitem{xu2019elucidation}
Xu, Shujing and Gilliland, Frank D and Conti, David V.
\newblock Elucidation of causal direction between asthma and obesity: a bi-directional Mendelian randomization study.
\newblock {\em International journal of epidemiology}, 2019.

\bibitem{uppunda2019association}
Uppunda, Deepak and Shetty, Ranjan K and Rao, Pragna and Razak, Abdul and Shetty, Kiran and Chauhan, Sheetal and Singh, Ajit and others.
\newblock Association of Metabolic Obesity and BMI Status with Severity of Angiographic Coronary Artery Disease in Stable Angina Patients..
\newblock {\em Journal of Clinical \& Diagnostic Research}, volume 13, number 4, 2019.

\bibitem{kanazawa2019overweight}
Kanazawa, Ippei and Notsu, Masakazu and Takeno, Ayumu and Tanaka, Ken-ichiro and Sugimoto, Toshitsugu.
\newblock Overweight and underweight are risk factors for vertebral fractures in patients with type 2 diabetes mellitus.
\newblock {\em Journal of bone and mineral metabolism}, volume 37, number 4, pages 703--710. Springer, 2019.

\bibitem{ruffner2018complications}
Ruffner, Melanie A and Sullivan, Kathleen E and others.
\newblock Complications associated with underweight primary immunodeficiency patients: prevalence and associations within the USIDNET Registry.
\newblock {\em Journal of clinical immunology}, volume 38, number 3, pages 283--293. Springer, 2018.

\bibitem{nakagawa2018postnatal}
Nakagawa, Yuichi and Nakanishi, Toshiki and Satake, Eiichiro and Matsushita, Rie and Saegusa, Hirokazu and Kubota, Akira and Natsume, Hiromune and Shibata, Yukinobu and Fujisawa, Yasuko.
\newblock Postnatal BMI changes in children with different birthweights: A trial study for detecting early predictive factors for pediatric obesity.
\newblock {\em Clinical Pediatric Endocrinology}, volume 27, number 1, pages 19--29. The Japanese Society for Pediatric Endocrinology, 2018.

\bibitem{cao2018openpose}
Cao, Zhe and Hidalgo, Gines and Simon, Tomas and Wei, Shih-En and Sheikh, Yaser.
\newblock OpenPose: realtime multi-person 2D pose estimation using Part Affinity Fields.
\newblock {\em arXiv preprint arXiv:1812.08008}, 2018.

\bibitem{wen2013computational}
Wen, Lingyun and Guo, Guodong.
\newblock A computational approach to body mass index prediction from face images.
\newblock {\em Image and Vision Computing}, volume 31, number 5, pages 392--400. Elsevier, 2013.

\bibitem{velardo2010weight}
Velardo, Carmelo and Dugelay, Jean-Luc.
\newblock Weight estimation from visual body appearance.
\newblock {\em 2010 Fourth IEEE International Conference on Biometrics: Theory, Applications and Systems (BTAS)}, pages 1--6. IEEE, 2010.

\bibitem{parkhi2015deep}
Parkhi, Omkar M and Vedaldi, Andrea and Zisserman, Andrew and others.
\newblock Deep face recognition..
\newblock {\em bmvc}, volume 1, number 3, pages 6, 2015.

\bibitem{simonyan2014very}
Simonyan, Karen and Zisserman, Andrew.
\newblock Very deep convolutional networks for large-scale image recognition.
\newblock {\em arXiv preprint arXiv:1409.1556}, 2014.

\bibitem{ozbulak2016transferable}
Ozbulak, Gokhan and Aytar, Yusuf and Ekenel, Hazim Kemal.
\newblock How transferable are CNN-based features for age and gender classification?.
\newblock {\em 2016 International Conference of the Biometrics Special Interest Group (BIOSIG)}, pages 1--6. IEEE, 2016.

\bibitem{nahavandi2017skeleton}
Nahavandi, D and Abobakr, A and Haggag, H and Hossny, M and Nahavandi, S and Filippidis, D.
\newblock A skeleton-free kinect system for body mass index assessment using deep neural networks.
\newblock {\em 2017 IEEE International Systems Engineering Symposium (ISSE)}, pages 1--6. IEEE, 2017.

\bibitem{NHANES1999}
\newblock National Health and Nutrition Examination Survey, Center for Disease.
\newblock {\em Control and Prevention}, 2006.

\bibitem{pfitzner2017evaluation}
Pfitzner, Christian and May, Stefan and N{\"u}chter, Andreas.
\newblock Evaluation of Features from RGB-D Data for Human Body Weight Estimation.
\newblock {\em IFAC-PapersOnLine}, volume 50, number 1, pages 10148--10153. Elsevier, 2017.

\bibitem{kocabey2018using}
Kocabey, Enes and Ofli, Ferda and Marin, Javier and Torralba, Antonio and Weber, Ingmar.
\newblock Using computer vision to study the effects of bmi on online popularity and weight-based homophily.
\newblock {\em International Conference on Social Informatics}, pages 129--138. Springer, 2018.

\bibitem{barr2018detecting}
Barr, Makenzie and Guo, Guodong and Colby, Sarah and Olfert, Melissa.
\newblock Detecting body mass index from a facial photograph in lifestyle intervention.
\newblock {\em Technologies}, volume 6, number 3, pages 83. Multidisciplinary Digital Publishing Institute, 2018.

\bibitem{drucker1997support}
Drucker, Harris and Burges, Christopher JC and Kaufman, Linda and Smola, Alex J and Vapnik, Vladimir.
\newblock Support vector regression machines.
\newblock {\em Advances in neural information processing systems}, pages 155--161, 1997.

\bibitem{viola2001robust}
Viola, Paul and Jones, Michael.
\newblock Robust real-time face detection.
\newblock {\em null}, pages 747. IEEE, 2001.

\bibitem{ricanek2006morph}
Ricanek, Karl and Tesafaye, Tamirat.
\newblock Morph: A longitudinal image database of normal adult age-progression.
\newblock {\em 7th International Conference on Automatic Face and Gesture Recognition (FGR06)}, pages 341--345. IEEE, 2006.

\bibitem{milborrow2007locating}
Milborrow, Stephen.
\newblock Locating facial features with active shape models.
\newblock {\em University of Cape Town}, 2007.

\bibitem{jahandideh2018physical}
Jahandideh, Rashidedin and Targhi, Alireza Tavakoli and Tahmasbi, Maryam.
\newblock Physical Attribute Prediction Using Deep Residual Neural Networks.
\newblock {\em arXiv preprint arXiv:1812.07857}, 2018.

\bibitem{jiang2019visual}
Jiang, Min and Shang, Yuanyuan and Guo, Guodong.
\newblock On visual BMI analysis from facial images.
\newblock {\em Image and Vision Computing}, volume 89, pages 183--196. Elsevier, 2019.

\bibitem{wu2018light}
Wu, Xiang and He, Ran and Sun, Zhenan and Tan, Tieniu.
\newblock A light cnn for deep face representation with noisy labels.
\newblock {\em IEEE Transactions on Information Forensics and Security}, volume 13, number 11, pages 2884--2896. IEEE, 2018.

\bibitem{wen2016discriminative}
Wen, Yandong and Zhang, Kaipeng and Li, Zhifeng and Qiao, Yu.
\newblock A discriminative feature learning approach for deep face recognition.
\newblock {\em European conference on computer vision}, pages 499--515. Springer, 2016.

\bibitem{deng2019arcface}
Deng, Jiankang and Guo, Jia and Xue, Niannan and Zafeiriou, Stefanos.
\newblock Arcface: Additive angular margin loss for deep face recognition.
\newblock {\em Proceedings of the IEEE Conference on Computer Vision and Pattern Recognition}, pages 4690--4699, 2019.

\bibitem{he2017mask}
He, Kaiming and Gkioxari, Georgia and Doll{\'a}r, Piotr and Girshick, Ross.
\newblock Mask r-cnn.
\newblock {\em Proceedings of the IEEE international conference on computer vision}, pages 2961--2969, 2017.

\bibitem{alhashim2018high}
Alhashim, Ibraheem and Wonka, Peter.
\newblock High quality monocular depth estimation via transfer learning.
\newblock {\em arXiv preprint arXiv:1812.11941}, 2018.

\bibitem{henderson2016perception}
Henderson, Audrey J and Holzleitner, Iris J and Talamas, Sean N and Perrett, David I.
\newblock Perception of health from facial cues.
\newblock {\em Philosophical Transactions of the Royal Society B: Biological Sciences}, volume 371, number 1693, pages 20150380. The Royal Society, 2016.

\bibitem{mayer2017bmi}
Mayer, Christine and Windhager, Sonja and Schaefer, Katrin and Mitteroecker, Philipp.
\newblock BMI and WHR are reflected in female facial shape and texture: a geometric morphometric image analysis.
\newblock {\em PloS one}, volume 12, number 1, pages e0169336. Public Library of Science San Francisco, CA USA, 2017.

\bibitem{lee2008assessment}
Lee, Seon Yeong and Gallagher, Dympna.
\newblock Assessment methods in human body composition.
\newblock {\em Current opinion in clinical nutrition and metabolic care}, volume 11, number 5, pages 566. NIH Public Access, 2008.

\bibitem{zhu2016still}
Zhu, Yu and Li, Yan and Mu, Guowang and Shan, Shiguang and Guo, Guodong.
\newblock Still-to-video face matching using multiple geodesic flows.
\newblock {\em IEEE Transactions on Information Forensics and Security}, volume 11, number 12, pages 2866--2875. IEEE, 2016.

\bibitem{gunther2017unconstrained}
G{\"u}nther, Manuel and Hu, Peiyun and Herrmann, Christian and Chan, Chi-Ho and Jiang, Min and Yang, Shufan and Dhamija, Akshay Raj and Ramanan, Deva and Beyerer, J{\"u}rgen and Kittler, Josef and others.
\newblock Unconstrained face detection and open-set face recognition challenge.
\newblock {\em 2017 IEEE International Joint Conference on Biometrics (IJCB)}, pages 697--706. IEEE, 2017.

\bibitem{wang2017learning}
Wang, Qiangchang and Guo, Guodong and Nouyed, Mohammad Iqbal.
\newblock Learning channel inter-dependencies at multiple scales on dense networks for face recognition.
\newblock {\em arXiv preprint arXiv:1711.10103}, 2017.

\bibitem{pfitzner2018body}
Pfitzner, Christian and May, Stefan and Nüchter, Andreas.
\newblock Body weight estimation for dose-finding and health monitoring of lying, standing and walking patients based on RGB-D data.
\newblock {\em Sensors}, volume 18, number 5, pages 1311. Multidisciplinary Digital Publishing Institute, 2018.

\bibitem{affuso2018method}
Affuso, Olivia and Pradhan, Ligaj and Zhang, Chengcui and Gao, Song and Wiener, Howard W and Gower, Barbara and Heymsfield, Steven B and Allison, David B.
\newblock A method for measuring human body composition using digital images.
\newblock {\em PloS one}, volume 13, number 11, pages e0206430. Public Library of Science San Francisco, CA USA, 2018.

\bibitem{andersson2015gender}
Andersson, Virginia Ortiz and Amaral, Livia S and Tonini, Aline R and Araujo, Ricardo M.
\newblock Gender and body mass index classification using a microsoft kinect sensor.
\newblock {\em The Twenty-Eighth International Flairs Conference}, 2015.

\bibitem{farina2016smartphone}
Farina, Gian Luca and Spataro, Fabrizio and De Lorenzo, Antonino and Lukaski, Henry.
\newblock A smartphone application for personal assessments of body composition and phenotyping.
\newblock {\em Sensors}, volume 16, number 12, pages 2163. Multidisciplinary Digital Publishing Institute, 2016.

\bibitem{gunel2019face}
G{\"u}nel, Semih and Rhodin, Helge and Fua, Pascal.
\newblock What Face and Body Shapes Can Tell Us About Height.
\newblock {\em 2019 IEEE/CVF International Conference on Computer Vision Workshop (ICCVW)}, pages 1819--1827. IEEE, 2019.

\bibitem{pascali2016face}
Pascali, Maria Antonietta and Giorgi, Daniela and Bastiani, Luca and Buzzigoli, E and Henr{\'\i}quez, Pedro and Matuszewski, Bogdan J and Morales, M-A and Colantonio, Sara.
\newblock Face morphology: Can it tell us something about body weight and fat?.
\newblock {\em Computers in biology and medicine}, volume 76, pages 238--249. Elsevier, 2016.

\bibitem{jiang2020visual}
Jiang, Min and Guo, Guodong and Mu, Guowang.
\newblock Visual BMI estimation from face images using a label distribution based method.
\newblock {\em Computer Vision and Image Understanding}, pages 102985. Elsevier, 2020.

\bibitem{vakli2020predicting}
Vakli, P{\'a}l and De{\'a}k-Meszl{\'e}nyi, Regina J and Auer, Tibor and Vidny{\'a}nszky, Zolt{\'a}n.
\newblock Predicting Body Mass Index From Structural MRI Brain Images Using a Deep Convolutional Neural Network.
\newblock {\em Frontiers in Neuroinformatics}, volume 14, pages 10. Frontiers, 2020.

\bibitem{ilao2019bmimatic}
Ilao, Adomar L and Cardino, Adrian Christopher and Fernandez, Clarence and Saulon, Lawrence.
\newblock BMIMatic: Body mass index derivation from captured images.
\newblock {\em Eleventh International Conference on Digital Image Processing (ICDIP 2019)}, volume 11179, pages 111790J. International Society for Optics and Photonics, 2019.

\bibitem{pantanowitz2019estimation}
Pantanowitz, Adam and Cohen, Emmanuel and Gradidge, Philippe and Crowther, Nigel and Aharonson, Vered and Rosman, Benjamin and Rubin, David M.
\newblock Estimation of Body Mass Index from Photographs using Deep Convolutional Neural Networks.
\newblock {\em arXiv preprint arXiv:1908.11694}, 2019.

\bibitem{verma2017obesity}
Verma, Shalini and Hussain, M Ejaz.
\newblock Obesity and diabetes: an update.
\newblock {\em Diabetes \& Metabolic Syndrome: Clinical Research \& Reviews}, volume 11, number 1, pages 73--79. Elsevier, 2017.

\bibitem{alpert2018obesity}
Alpert, Martin A and Karthikeyan, Kamalesh and Abdullah, Obai and Ghadban, Rugheed.
\newblock Obesity and cardiac remodeling in adults: mechanisms and clinical implications.
\newblock {\em Progress in cardiovascular diseases}, volume 61, number 2, pages 114--123. Elsevier, 2018.

\bibitem{lin2018association}
Lin, Hsien-Ho and Wu, Chieh-Yin and Wang, Chih-Hui and Fu, Han and L{\"o}nnroth, Knut and Chang, Yi-Cheng and Huang, Yen-Tsung.
\newblock Association of obesity, diabetes, and risk of tuberculosis: two population-based cohorts.
\newblock {\em Clinical Infectious Diseases}, volume 66, number 5, pages 699--705. Oxford University Press US, 2018.

\bibitem{iandola2014densenet}
Iandola, Forrest and Moskewicz, Matt and Karayev, Sergey and Girshick, Ross and Darrell, Trevor and Keutzer, Kurt.
\newblock Densenet: Implementing efficient convnet descriptor pyramids. arXiv 2014.
\newblock {\em arXiv preprint arXiv:1404.1869}.

\bibitem{wang2004image}
Wang, Zhou and Bovik, Alan C and Sheikh, Hamid R and Simoncelli, Eero P.
\newblock Image quality assessment: from error visibility to structural similarity.
\newblock {\em IEEE transactions on image processing}, volume 13, number 4, pages 600--612. IEEE, 2004.

\bibitem{lin2014microsoft}
Lin, Tsung-Yi and Maire, Michael and Belongie, Serge and Hays, James and Perona, Pietro and Ramanan, Deva and Doll{\'a}r, Piotr and Zitnick, C Lawrence.
\newblock Microsoft coco: Common objects in context.
\newblock {\em European conference on computer vision}, pages 740--755. Springer, 2014.

\bibitem{khan2020relative}
Khan, Soofia and Xanthakos, Stavra A and Hornung, Lindsey and Arce-Clachar, Catalina and Siegel, Robert and Kalkwarf, Heidi J.
\newblock Relative accuracy of bioelectrical impedance analysis for assessing body composition in children with severe obesity.
\newblock {\em Journal of pediatric gastroenterology and nutrition}, volume 70, number 6, pages e129--e135. LWW, 2020.

\bibitem{seo2018validation}
Vermeiren, Eline and Ysebaert, Marijke and Van Hoorenbeeck, Kim and Bruyndonckx, Luc and Van Dessel, Kristof and Van Helvoirt, Maria and De Guchtenaere, Ann and De Winter, Benedicte and Verhulst, Stijn and Van Eyck, Annelies.
\newblock Comparison of bioimpedance spectroscopy and dual energy X-ray absorptiometry for assessing body composition changes in obese children during weight loss.
\newblock {\em European Journal of Clinical Nutrition}, volume 75, number 1, pages 73--84. Nature Publishing Group, 2021.

\bibitem{bjorkman2020associations}
Bjorkman, Mikko P and Jyvakorpi, Satu K and Strandberg, Timo E and Pitkala, Kaisu H and Tilvis, Reijo S.
\newblock The associations of body mass index, bioimpedance spectroscopy-based calf intracellular resistance, single-frequency bioimpedance analysis and physical performance of older people.
\newblock {\em Aging clinical and experimental research}, volume 32, number 6, pages 1077--1083. Springer, 2020.

\bibitem{shannon2019comparison}
Shannon, Carley A and Brown, Justin R and Del Pozzi, Andrew T.
\newblock Comparison of Body Composition Prediction Equations with Air Displacement Plethysmography in Overweight and Obese Caucasian Males.
\newblock {\em International journal of exercise science}, volume 12, number 4, pages 1034. Western Kentucky University, 2019.

\bibitem{pellonpera2019body}
Pellonper{\"a}, Outi and Koivuniemi, Ella and Vahlberg, Tero and Mokkala, Kati and Tertti, Kristiina and R{\"o}nnemaa, Tapani and Laitinen, Kirsi.
\newblock {Body composition measurement by air displacement plethysmography in pregnancy: Comparison of predicted versus measured thoracic gas volume.}
\newblock {\em Nutrition}, volume 60, pages 227--229. Elsevier, 2019.

\bibitem{ashby2020high}
Ashby-Thompson, Maxine and Ji, Ying and Wang, Jack and Yu, Wen and Thornton, John C and Wolper, Carla and Weil, Richard and Chambers, Earle C and Laferr{\`e}re, Blandine and Pi-Sunyer, F Xavier and others.
\newblock High-Resolution Three-Dimensional Photonic Scan-Derived Equations Improve Body Surface Area Prediction in Diverse Populations.
\newblock {\em Obesity}, volume 28, number 4, pages 706--717. Wiley Online Library, 2020.

\bibitem{wells2019three}
Wells, Jonathan CK.
\newblock Three-dimensional optical scanning for clinical body shape assessment comes of age.
\newblock {\em The American journal of clinical nutrition}, volume 110, number 6, pages 1272--1274. Oxford University Press, 2019.

\bibitem{pasanta2018body}
Pasanta, Duanghathai and Tungjai, Montree and Chancharunee, Sirirat and Sajomsang, Warayuth and Kothan, Suchart.
\newblock Body mass index and its effects on liver fat content in overweight and obese young adults by proton magnetic resonance spectroscopy technique.
\newblock {\em World journal of hepatology}, volume 10, number 12, pages 924. Baishideng Publishing Group Inc, 2018.

\bibitem{bini2020body}
Bini, Jason and Bhatt, Shivani and Hillmer, Ansel T and Gallezot, Jean-Dominique and Nabulsi, Nabeel and Pracitto, Richard and Labaree, David and Kapinos, Michael and Ropchan, Jim and Matuskey, David and others.
\newblock Body Mass Index and Age Effects on Brain 11$\beta$-Hydroxysteroid Dehydrogenase Type 1: a Positron Emission Tomography Study.
\newblock {\em Molecular imaging and biology}, volume 22, number 4, pages 1124--1131. Springer, 2020.

\bibitem{vaswani2017attention}
Vaswani, Ashish and Shazeer, Noam and Parmar, Niki and Uszkoreit, Jakob and Jones, Llion and Gomez, Aidan N and Kaiser, {\L}ukasz and Polosukhin, Illia.
\newblock {Attention is all you need.}
\newblock {\em Advances in neural information processing systems}, volume 30, 2017.

\bibitem{dosovitskiy2020image}
Dosovitskiy, Alexey and Beyer, Lucas and Kolesnikov, Alexander and Weissenborn, Dirk and Zhai, Xiaohua and Unterthiner, Thomas and Dehghani, Mostafa and Minderer, Matthias and Heigold, Georg and Gelly, Sylvain and others.
\newblock {An image is worth 16x16 words: Transformers for image recognition at scale.}
\newblock {\em arXiv preprint arXiv:2010.11929}, 2020.

\bibitem{vermeiren2021comparison}
Eline Vermeiren, Marijke Ysebaert, Kim Van Hoorenbeeck, Luc Bruyndonckx, Kristof Van Dessel, Maria Van Helvoirt, Ann De Guchtenaere, Benedicte De Winter, Stijn Verhulst, and Annelies Van Eyck.
\newblock {Comparison of bioimpedance spectroscopy and dual energy X-ray absorptiometry for assessing body composition changes in obese children during weight loss.}
\newblock {\em European Journal of Clinical Nutrition}, volume 75, number 1, pages 73--84, 2021.

\end{thebibliography}
\end{document}